\documentclass[twoside,11pt]{article}
\usepackage{subcaption}
\usepackage{graphicx}

\usepackage{blindtext}

 \usepackage[abbrvbib]{jmlr2e}

\usepackage{wrapfig} %
\usepackage{adjustbox} %
    \usepackage[column=0]{cellspace} %
    \usepackage{multirow}
\usepackage{enumitem}
\usepackage{float}
\usepackage[colorinlistoftodos]{todonotes}
\usepackage{amsmath,amsfonts,bm,amssymb,mathtools}

\usepackage{natbib}       %
\usepackage[capitalise]{cleveref}
\usepackage[T1]{fontenc}    %
\usepackage{booktabs}       %
\usepackage{amsfonts}       %
\usepackage{nicefrac}       %
\usepackage{microtype}      %
\usepackage{bbm}            %

\usepackage{hyperref} 
\makeatletter
\def\@footnotecolor{red}
\define@key{Hyp}{footnotecolor}{%
 \HyColor@HyperrefColor{#1}\@footnotecolor%
}
\def\@footnotemark{%
    \leavevmode
    \ifhmode\edef\@x@sf{\the\spacefactor}\nobreak\fi
    \stepcounter{Hfootnote}%
    \global\let\Hy@saved@currentHref\@currentHref
    \hyper@makecurrent{Hfootnote}%
    \global\let\Hy@footnote@currentHref\@currentHref
    \global\let\@currentHref\Hy@saved@currentHref
    \hyper@linkstart{footnote}{\Hy@footnote@currentHref}%
    \@makefnmark
    \hyper@linkend
    \ifhmode\spacefactor\@x@sf\fi
    \relax
  }%
\makeatother
    \hypersetup{
    	colorlinks = true,
    	linkcolor = blue,
    	anchorcolor = blue,
    	citecolor = blue,
    	filecolor = blue,
    	urlcolor = blue,
    	footnotecolor=black
    }

\usepackage{xcolor} %

\usepackage{xparse}

\def\1{\bm{1}}

\def\bP{{\bm{P}}}

\def\calB{{\mathcal{B}}}
\def\calC{{\mathcal{C}}}
\def\calD{{\mathcal{D}}}
\def\calE{{\mathcal{E}}}
\def\calF{{\mathcal{F}}}

\def\calH{{\mathcal{H}}}

\def\calN{{\mathcal{N}}}
\def\calO{{\mathcal{O}}}
\def\calP{{\mathcal{P}}}

\def\calR{{\mathcal{R}}}

\def\calW{{\mathcal{W}}}
\def\calX{{\mathcal{X}}}
\def\calY{{\mathcal{Y}}}
\def\calZ{{\mathcal{Z}}}

\def\sE{{\mathbb{E}}}

\def\sN{{\mathbb{N}}}

\def\sP{{\mathbb{P}}}

\def\sR{{\mathbb{R}}}

\newcommand{\WD}{\mathcal{W}_1}

\newcommand{\E}{\mathbb{E}}
\newcommand{\Ls}{\ell}
\newcommand{\R}{\mathbb{R}}

\newcommand{\parterror}{\mathrm{cost}_{\mathrm{partition}}}
\newcommand{\transerror}{\mathrm{err}_{\mathrm{transport}}}
\newcommand{\transcost}{\mathrm{cost}_{\mathrm{transport}}}
\newcommand{\totalerror}{\mathrm{err}}

\DeclareMathOperator*{\argmin}{arg\,min}

\newcommand{\muadv}{\mu^{\mathrm{adv}}}

\newcommand{\diam}{\mathrm{diam}}
\newcommand{\Lip}{\mathrm{Lip}}

\newcommand{\fhat}{{\hat{f}}}
\newcommand{\Rhat}{{\hat{\mathfrak{R}}}}

\NewDocumentCommand{\nn}{}{\mathbb{N}}
    \NewDocumentCommand{\rr}{o}{\mathbb{R}^{\IfValueT{#1}{#1}}}

\definecolor{darkcerulean}{rgb}{0.03, 0.27, 0.49}
\definecolor{darkmidnightblue}{rgb}{0.0, 0.2, 0.4}
\definecolor{darkcyan}{rgb}{0.0, 0.55, 0.55}
\definecolor{LandscapeBlue}{RGB}{10,38,255}
\definecolor{darkgreen}{rgb}{0.0, 0.2, 0.13}
\definecolor{deepjunglegreen}{rgb}{0.0, 0.29, 0.29}
\definecolor{applegreen}{rgb}{0.55, 0.71, 0.0}
\definecolor{LandscapeGreen}{RGB}{18,125,9}
\definecolor{darkcandyapplered}{rgb}{0.64, 0.0, 0.0}
\definecolor{darkred}{rgb}{0.55, 0.0, 0.0}
\definecolor{darkscarlet}{rgb}{0.34, 0.01, 0.1}
\definecolor{jasper}{rgb}{0.84, 0.23, 0.24}
\definecolor{darkjazzberryjam}{rgb}{0.45, 0.04, 0.37}
\definecolor{LandscapePurple}{RGB}{135,0,170}
\definecolor{MidnightBlue}{RGB}{25,25,112}
\definecolor{MidnightBlueComplementingGreen}{RGB}{25,112,25}
\definecolor{MidnightBlueComplementingPurple}{RGB}{112,25,112}
\definecolor{MidnightBlueComplementingRed}{RGB}{112,25,69}

\newcommand{\eqdef}{\ensuremath{\stackrel{\mbox{\upshape\tiny def.}}{=}}}

\makeatletter
\renewcommand*{\p@equation}{}
\makeatother

\newtheorem{assumption}[theorem]{Assumption}

\DeclarePairedDelimiter\abs{\lvert}{\rvert}%
\DeclarePairedDelimiter\norm{\lVert}{\rVert}%

\makeatletter
\let\oldabs\abs
\def\abs{\@ifstar{\oldabs}{\oldabs*}}
\let\oldnorm\norm
\def\norm{\@ifstar{\oldnorm}{\oldnorm*}}
\makeatother

\usepackage{booktabs}
\newcommand{\raa}[1]{\renewcommand{\arraystretch}{#1}}
\renewcommand{\arraystretch}{1.25}
\usepackage{lscape}

\usepackage{comment}

\usepackage{marginnote}
\definecolor{MidnightBlue}{RGB}{25,25,112}
\definecolor{MidnightBlueComplementingGreen}{RGB}{25,112,25}
\definecolor{MidnightBlueComplementingPurple}{RGB}{112,25,112}
\definecolor{MidnightBlueComplementingRed}{RGB}{112,25,69}
\definecolor{deepjunglegreen}{rgb}{0.0, 0.29, 0.29}
\definecolor{applegreen}{rgb}{0.55, 0.71, 0.0}
\definecolor{WowColor}{rgb}{.75,0,.75}
\definecolor{MildlyAlarming}{rgb}{0.85,0.25,0.1}
\definecolor{SubtleColor}{rgb}{0,0,.50}
\definecolor{SubtleColor2}{rgb}{0.6,0.21,.50}

\newcounter{margincounter}

\usepackage{ifthen}

\usepackage{etoolbox}

\newboolean{showRevision}
\setboolean{showRevision}{false} %

\definecolor{darkspringgreen}{rgb}{0.09, 0.45, 0.27}
\newcommand{\added}[1]{
    \ifthenelse{\boolean{showRevision}}{\textcolor{darkspringgreen}{#1}}{#1}
}

\definecolor{dollarbill}{rgb}{0.52, 0.73, 0.4}
\newcommand{\edit}[2]{
    \ifthenelse{\boolean{showRevision}}{\textcolor{darkspringgreen}{\sout{#1} #2}}{#2}}

\definecolor{darkpastelred}{rgb}{0.76, 0.23, 0.13}
\newcommand{\deleted}[1]{\ifthenelse{\boolean{showRevision}}{\textcolor{darkspringgreen}{\sout{#1}}}{}}

\NewDocumentCommand{\Anastasis}{mo}{
    \IfValueF{#2}{
                        {{\scriptsize
                            \textcolor{deepjunglegreen}{ 
                            \textbf{A:}
                            \textit{{#1}}
                            }
                        }}
        }
    \IfValueT{#2}{
                        \marginnote{{\scriptsize
                            \textcolor{deepjunglegreen}{ 
                            \textbf{A:}
                            \textit{{#1}}
                            }
                        }}
        }
                    }
\NewDocumentCommand{\Jonas}{mo}{
    \IfValueF{#2}{
                        {{\scriptsize
                            \textcolor{purple}{ 
                            \textbf{J:}
                            \textit{{#1}}
                            }
                        }}
        }
    \IfValueT{#2}{
                        \marginnote{{\scriptsize
                            \textcolor{purple}{ 
                            \textbf{J:}
                            \textit{{#1}}
                            }
                        }}
        }
                    }
\definecolor{airforceblue}{rgb}{0.36, 0.54, 0.66}                    
\NewDocumentCommand{\Parnian}{mo}{
    \IfValueF{#2}{
                        {{\scriptsize
                            \textcolor{airforceblue}{ 
                            \textbf{P:}
                            {#1}
                            }
                        }}
        }
    \IfValueT{#2}{
                        \marginnote{{\scriptsize
                            \textcolor{airforceblue}{ 
                            \textbf{P:}
                            {#1}
                            }
                        }}
        }
                    }

\NewDocumentCommand{\Songyan}{mo}{
    \IfValueF{#2}{
                        {{\scriptsize
                            \textcolor{applegreen}{ 
                            \textbf{S:}
                            \textit{{#1}}
                            }
                        }}
        }
    \IfValueT{#2}{
                        \marginnote{{\scriptsize
                            \textcolor{applegreen}{ 
                            \textbf{S:}
                            \textit{{#1}}
                            }
                        }}
        }
                    }

\NewDocumentCommand{\Andreas}{mo}{
    \IfValueF{#2}{
                        {{\scriptsize
                            \textcolor{WowColor}{ 
                            \textbf{A1:}
                            \textit{{#1}}
                            }
                        }}
        }
    \IfValueT{#2}{
                        \marginnote{{\scriptsize
                            \textcolor{WowColor}{ 
                            \textbf{A1:}
                            \textit{{#1}}
                            }
                        }}
        }
                    }

\usepackage{lastpage}
\jmlrheading{24}{2023}{1-\pageref{LastPage}}{11/22; Revised
7/23}{11/23}{22-1293}{Songyan Hou, Parnian Kassraie, Anastasis Kratsios, Andreas Krause and Jonas Rothfuss}
\ShortHeadings{Instance-Dependent Generalization Bounds via OT}{Hou, Kassraie, Kratsios, Krause and Rothfuss}
\firstpageno{1}

\begin{document}

\title{Instance-Dependent Generalization Bounds via Optimal Transport}

\author{\name Songyan Hou\thanks{Equal contribution, all authors are listed in alphabetic order. } \email songyan.hou@ethz.ch \\
        \addr Department of Mathematics,  ETH Zurich 
        \AND
        \name Parnian Kassraie\footnotemark[1] \thanks{Corresponding author.} \email pkassraie@ethz.ch \\
        \addr Department of Computer Science, ETH Zurich
        \AND
        \name Anastasis Kratsios\footnotemark[1]\email kratsioa@mcmaster.ca \\
        \addr Department of Mathematics, McMaster University and the Vector Institute
        \AND    
        \name Andreas Krause \email krausea@ethz.ch \\
        \addr Department of Computer Science, ETH Zurich
        \AND
        \name Jonas Rothfuss\footnotemark[1]\email jonas.rothfuss@inf.ethz.ch \\
        \addr Department of Computer Science, ETH Zurich}

\editor{Gabor Lugosi}

\maketitle

\begin{abstract}
\looseness -1 Existing generalization bounds fail to explain crucial factors that drive the generalization of modern neural networks.
Since such bounds often hold uniformly over all parameters, they suffer from over-parametrization and fail to account for the strong inductive bias of initialization and stochastic gradient descent.
As an alternative, we propose a novel {\em optimal transport} interpretation of the generalization problem. This allows us to derive {\em instance-dependent} generalization bounds that depend on the {\em local Lipschitz regularity} of the {\em learned prediction function} in the data space. Therefore, our bounds are agnostic to the parametrization of the model and work well when the number of training samples is much smaller than the number of parameters.
With small modifications, our approach yields accelerated rates for data on {\em low-dimensional manifolds} and guarantees under {\em distribution shifts}. We empirically analyze our generalization bounds for neural networks, showing that the bound values are meaningful and capture the effect of popular regularization methods during training.
\end{abstract}


\section{Introduction}
A core challenge in machine learning is to generalize well beyond the training data. We want to choose a hypothesis $f \in \calF$, from a hypothesis class $\calF$, that not only gives a small training error but also yields good predictions for previously unseen data points. Accordingly, statistical learning theory aims to provide generalization guarantees and understand the factors that drive it.
Generalization is typically described through the discrepancy between two key quantities: The empirical risk $ \hat{\mathfrak{R}}(f)$, i.e., the prediction error of $f$ on the training data and the expected risk $\mathfrak{R}(f)$, i.e., the expected error under the unknown data-distribution.
A common type of guarantees are \emph{uniform bounds} which control the generalization gap $\mathfrak{R}(f) - \hat{\mathfrak{R}}(f)$ with high probability, simultaneously for \emph{all hypotheses} $f \in \calF$ \citep[e.g.,][]{vapnik1971vc_dim, JMLR:bartlett2002rademacher}. Such bounds include terms that quantify the complexity of the hypothesis $f$ or hypothesis space $\calF$.
For neural networks (NNs), this complexity term grows rapidly with the number of parameters \citep[e.g.,][]{bartlett2017spectrally, neyshabur2015norm, harvey2017vcdim}. 
While the parameter space of NNs is vast, regular networks that are used in practice only seem to populate a small subset of the parameter space. This subset seemingly generalizes well and depends on model structure, initialization scheme, and optimization method in a complex manner.
In addition, there are many NN parameter configurations that correspond to the same neural network mapping, artificially inflating the complexity of the parametric hypothesis space.
Thus, such uniform bounds in the parameter space fail to explain the empirical generalization behavior of neural networks in the over-parameterized setting where the number of training examples is much smaller than the number of parameters \citep{belkin2019reconciling}.

Addressing this issue, we base our analysis on the geometric properties of the learned prediction function (i.e., hypothesis $f$) in the data domain. 
In particular, we partition the input domain into smaller neighborhoods and locally characterize $f$ via its \emph{local Lipschitz constant} when the domain is restricted to each neighborhood. Using principles from {\em optimal transport}, we obtain a bound that depends on the instance $f$ through its local Lipschitz constants.\edit{and is built on the following two key ideas. First,}{In particular,} we view the generalization gap as the worst-case impact on the loss when probability mass is transported from the empirical measure to the true data distribution. The magnitude of this impact depends on the local regularity of $f$ multiplied by the local transport cost, which decreases w.h.p. with the number of samples. \\
\edit{Second, unlike uniform bounds that hold with high probability simultaneously for all $f \in \calF$, our analysis focuses on one instance $f \in \calF$.
This approach is an alternative to the classical uniform bound and allows us to forego arguments about the complexity of the hypothesis space, which typically leads to vacuous bounds.}{Classical uniform bounds depend on the complexity of the hypothesis space or the global regularity of $f$ which is typically determined by the single most irregular part of $f$ in the domain. However, neural network functions often vary significantly in their (local) regularity across the domain. This typically leads to extremely loose or vacuous bounds. In contrast, our bounds can adapt to the local regularity properties of $f$ and, thus, minimize the negative impact of irregular parts of the learned function on the tightness of the bound.}

Overall, the presented generalization bound (\cref{thm:generalization}) has the following properties:
1) It is instance-dependent,\added{i.e., it adapts to the trained function $f$}and thus can capture the combined effect of initialization, training method, and model structure.
2) It characterizes $f$ geometrically via its local Lipschitz regularity; therefore, in contrast to parametric bounds, it does not suffer from over-parametrization. 
3) It is tighter than bounds based on the global Lipschitz properties of $f$ due to the fine-grained local analysis, which takes into account changes in the regularity of $f$ throughout the domain.
While our bounds generically hold for any machine learning model, we focus our exposition on neural network generalization and empirically verify the mentioned properties through experiments.
When applied to fully-connected ReLU networks, trained on simple regression and classification tasks, we observe that our result provides meaningful bound values in the same order of magnitude as the empirical risk, even for small sample sizes.
We empirically show that, unlike the majority of prior works, the bound does not explode as the number of network parameters increases.
Moreover, the value of the bound reflects the effect of regularization techniques applied \emph{during} training, e.g., weight-decay, early-stopping, and adversarial training.

Due to its transport-based derivation, our framework can be seamlessly adapted to obtain generalization certificates under distribution shifts or adversarial perturbations. 
The results mentioned above are corollaries of our core theorem, which is an optimal-transport-based concentration inequality for data-dependent locally regular functions. This theorem may be of independent interest and considers a spectrum of functions with different degrees of regularity, from non-smooth $\alpha$-H\"{o}lder functions to smooth and $s$-time differentiable instances. 

\paragraph{Outline} The paper is structured as follows. 
\begin{itemize}
    \item \cref{sec:gen_bounds} formalizes the problem setting and presents our main generalization bound (\cref{thm:generalization}) together with an extension to when the data is known to be concentrated on a low-dimensional manifold (Proposition~\ref{prop:manifold_generalization}).
    \item \cref{sec:props} discusses the key properties of our generalization bound which is data-dependent (\cref{sec:instance_dep}), non-parametric (\cref{sec:non-param}), and localized (\cref{sec:localized}). Every section also presents corresponding experiments on neural networks.\looseness -1
    \item \cref{sec:adversarial} considers\deleted{instance-dependent}generalization under distribution shifts, which is a natural corollary of our approach (Corollary~\ref{cor:adv_robust}). 
    \item \cref{sec:general_bounds} focuses on our core result (\cref{thm:concentration_main}). \cref{sec:Discussion} highlights the key technical tools used for this theorem, and \cref{sec:proof_sketch} outlines the proof methodology.
\end{itemize}

\section{Related Work}
Our work provides generalization bounds for learned prediction functions, contributing to the rich literature on generalization. %
A classic approach to explaining generalization are uniform bounds, which provide {\em uniform} guarantees over a class of estimators, also referred to as the hypothesis space.
Uniform bounds often depend on the combinatorial complexity of the hypothesis space, e.g., expressed in\added{the}form of the VC-dimension \citep{vapnik1971vc_dim} or the Rademacher complexity \citep{koltchinskii2001rademacher, JMLR:bartlett2002rademacher}. 
For neural networks, however, the hypothesis space is large and combinatorially explodes in size with the neural network width and depth, making the corresponding bounds loose \citep[cf.][]{bartlett1998almost, harvey2017vcdim, bartlett2019vc, sun2016depth}. Uniform bounds that utilize the parametric characterization of the network rapidly with the size of the neural network \citep[e.g.,][]{ neyshabur2015norm}. Overall, these approaches hardly explain the empirical generalization behavior of neural networks in the over-parameterized setting, where the number of samples is much smaller than the number of parameters \citep{belkin2019reconciling}. 
In fact, measures of neural network complexity based on the VC-dimension or parameter norm were found to be negatively correlated with the expected risk of convolutional neural networks \citep{jiang2019fantastic, Kuhn2021RobustnessTP}.
In contrast, we present results that use the geometric properties, i.e., local regularity, of the learned prediction function $f$. This allows us to avoid the dependence on the combinatorial complexity of function classes as well as direct dependency on the parametrization of $f$.

An alternative to guarantees that hold uniformly over a hypothesis space are instance-dependent bounds, where the value of the bound changes based on properties of the learned hypothesis, which generally depends on the training data. 
In this spirit, PAC-Bayesian learning theory provides generalization bounds which depend on the chosen posterior distribution, e.g., an instance of the random (Gibbs) learner \citep{McAllester98pacbayes, shawetaylor1997pacbayes, catoni2007pac, Alquier2016a, mhammedi2019pac}. 
Since PAC-Bayesian bounds do not trivially explode with the number of parameters of the model, they have gained increasing popularity in the context of neural networks \citep{langford2011pacbayes, dziugaite2017computing, Dziugaite2018EntropySGDOT,neyshabur2018pac, zhou2019non, golowich2020size}. For instance, they have been related to the sharpness of minima, i.e., the robustness to perturbations in the weight space \citep{keskar2016large, Neyshabur2017, dziugaite2017computing}, or the compressibility of a neural network \citep{zhou2019non, arora2018stronger, Kuhn2021RobustnessTP}.
Alternatively, in the case of neural networks, there also exist instance-dependent bounds that directly depend on the norm of the weights \citep{bartlett2017spectrally, golowich2020size}.
Nonetheless, due to their inherent focus on a model's parameters, the above results all suffer from the standard pitfalls of the over-parameterized setting so that the bounds become very loose once employed for larger networks.
We argue that the generalization capability of a learner is directly influenced by the geometrical properties of the learned model in the data domain rather than the number or values of its constructing parameters.
Following this idea, a body of work uses the properties of the classification margin \citep{antos2002data, sokolic2017, jiang2019margindistribution,soudry2018implicit,gunasekar2018implicit} to quantify generalization. A common theme in such works is that the generalization ability of neural networks relies crucially on the optimization procedure and can not be solely described by the hypothesis class. 
Following this logic, \citet{dziugaite2017computing, Dziugaite2018EntropySGDOT} adjust the training procedure so that it minimizes the bounds and, thereby, attain non-vacuous PAC-Bayesian bounds.
While the bound of \citet{Dziugaite2018EntropySGDOT} depends on a data-dependent prior, it considers generalization error with respect to a posterior distribution over neural network parameters. 
In contrast, we focus on the generalization properties of a single learning hypothesis (e.g., a single neural network) which is the result of training.

Our work also relates to approaches that quantify the local regularity of the learned prediction function. Examples of this are counting the number of linear regions of trained neural networks \citep{montufar2014number}, calculating the local Lipschitz constant of neural networks \citep{jordan2020exactly,herrera2020estimating} or the local Rademacher complexity \citep{BartlettBousquetMendelson_2005_AnnStat_LocalRademacherComplexities}. 
 
\looseness-1 We also contribute to the literature of distributionally robust optimization, \citep[cf.][]{sinha2017certifying}, since\deleted{,with little effort,} our bounds can be\added{easily}extended into a distributional robustness certificate (see Section \ref{sec:adversarial}). Our bound suggests that local Lipschitz estimators are more robust to distribution shifts, 
confirming recent results which control the global or local Lipschitz constants in order to achieve adversarially robust neural networks \citep{cisse2017parseval, salman2019provably, cohen2019certified, gouk2021regularisation, anil2019sorting,Muthukumar2022AdversarialRO}. 
Similar to recent work on distributional robustness \citep{gao2022distributionally, kuhn2019wasserstein, cranko2021generalised}, we rely on a transport-based change of measure inequality. The majority of prior work only bounds the difference between the empirical validation error and the expected risk under distribution shift. 
In contrast, we present a \added{much} stronger result, which bounds the gap to the training error. 
\citet{mohajerin2018data, staib2019distributionally} also consider the gap to the training error, but only for a particular minimax estimator and, in the latter case, only under much more restrictive smoothness assumptions.
\deleted{Contrary to our analysis, which}
\deleted{holds for arbitrary Lipschitz estimators.}
We provide a more in-depth comparison in
 \cref{sec:adversarial}, once the notation is formally set.

\section{A Localized Bound on the Generalization Error}
\label{sec:gen_bounds}

We consider datapoints $(x, y)$ where $x \in \calX $ are observed input features and $y \in \calY$ are target values/labels. 
To formulate the learning problem, we assume that the data is generated via an \emph{unknown} probability measure $\mu \in \calP(\calX \times \calY)$\added{, where $\calP$ denotes the space of probability measures defined over $\calX \times \calY$.}
Given a dataset $\calD^N = \{ (x_i,y_i) \}_{i=1}^N$ of i.i.d. draws from $\mu$, the goal of supervised learning is to
find a function $\hat{f}^N$ which can accurately predict the targets.
The quality of an estimator $\hat{f}^N$ is measured through a loss function $\Ls: \calY  \times \calY \rightarrow \R$.
Accordingly, we seek to attain a small {\em expected risk}, i.e., the expected loss under the data generating distribution 
\begin{equation*}
    \mathfrak{R}(\fhat^N; \mu) \coloneqq \sE_{(x,y) \sim \mu} \left[ \Ls(\hat{f}^N(x), y)) \right].
\end{equation*} 
Since $\mu$ is unknown, it is not possible to directly evaluate $\mathfrak{R}(\fhat^N; \mu)$ given the training data $\calD^N$. However, based on the dataset, we can compute the {\em empirical risk}
\begin{equation*}
\Rhat(\fhat^N) \coloneqq  \mathfrak{R}(\fhat^N; \mu^N)   = \frac{1}{N}\sum_{n=1}^N \Ls(\hat{f}^N(x_i), y_i).
\end{equation*}
which corresponds to the expected loss under an empirical measure $\mu^N := \frac{1}{N} \sum_{i=1}^N \delta_{(x_i,y_i)}$ for $\calD^N$. Often $\Rhat(\fhat^N)$ is also referred to as training error.
In this work, we aim to bound the {\em generalization gap} $\mathfrak{R}(\hat f^N; \mu) - \Rhat(\fhat^N)$. 
Importantly, as $\hat{f}^N$ already depends on the data $\calD^N$, $\Rhat(\fhat^N)$ is a biased estimator of the expected risk $\mathfrak{R}(\fhat^N; \mu)$. Thus, standard results for the concentration of averages do not apply. Instead, to bound the generalization gap, we also need to take into consideration the learning hypothesis $\hat{f}^N$ and quantify how well it generalizes from the training data $\calD^N$ to the general data distribution $\mu$.

In the following, we introduce the basic assumptions and tools which form the foundation of our generalization bounds:
\begin{assumption}\label{assum:domain}
The domain $\calX$ is a compact subset of $\sR^d$, the $d$-dimensional Euclidean space, and $\calY$ is a compact subset of $\sR$.
\end{assumption}
The assumption that $\calX$ and $\calY$ are compact is very common in statistical learning theory and implies that the bounded is the target values $y$ are observed with a bounded noise.
For instance, they are commonly used for uniform generalization bounds \citep[e.g.,][]{alon1997scale,JMLR:bartlett2002rademacher}, PAC-Bayesian Bounds \citep[e.g.,][]{McAllester98pacbayes, catoni2007pac}, and the more recent instance-dependent generalization bounds \citep[e.g., ][]{dziugaite2017computing,neyshabur2018pac,golowich2020size}. 

In addition, we require geometric regularity assumptions on both the estimator and the loss function. For this purpose,
we define the {\em local} Lipschitz constant of a function $g:\calX \rightarrow \calY$ when restricted to $P \subset \calX$ as,
\[
\Lip(g\vert P) \coloneqq \sup_{\substack{x_1,x_2\in P\\ x_1\neq x_2}}\, \frac{|g(x_1)-g(x_2)|}{\|x_1-x_2\|_2} \;.
\]
For $0 \leq L< \infty$, we say that a function $g$ is $L$-Lipschitz if the global Lipschitz constant $\Lip(g) \coloneqq \Lip(g\vert \calX)$ is bounded by $L$. We assume that the learned function $\fhat$ is Lipschitz continuous with Lipschitz constant $L_\fhat$:
\begin{assumption}\label{assum:estimator}
There exists a constant $L_\fhat\geq 0$ such that the estimator $\fhat^N$ almost surely satisfies $\Lip(\fhat^N) \leq L_\fhat$. 
\end{assumption}
Lipschitz estimators are perhaps the most common class of estimators and include, Gaussian processes with non-smooth kernels and neural networks with popular activation functions such as ReLUs, ELUs and tanh functions. In \cref{sec:general_bounds}, we extend our result to $\alpha$-H{\"o}lder and smooth estimators. We also require a Lipschitz loss function:
\begin{assumption}\label{assum:loss}
The loss function $\Ls:\calY \times \calY \rightarrow \sR$ is $L_\Ls$-Lipschitz.
\end{assumption}
Examples of Lipschitz continuous loss functions for classification are logit, hinge, or ramp loss \citep{hajek2019ece}. A Lipschitz loss for regression is the Huber loss, which satisfies $L_\Ls = 1$.
This loss is commonly used for training neural networks \citep[e.g.,][]{morales2020grokking, meyer2021alternative}, since compared to the squared error loss, it is more robust to outliers and large gradients that destabilize training.  

We take a localized approach, and instead of bounding the generalization error directly on the entire $\calX \times \calY$ space, we first partition the space and then compare the empirical and expected risk separately on each element of this partitioning.
A partitioning $\bP$ of size $k$ is a collection $\{P_1, \dots, P_i, \dots, P_k\}$ subsets of $\calX\times \calY$,
where $P_i \cap P_j = \emptyset$ and $\cup_{i=1}^k P_i = \calX\times \calY$, for every $1 \leq i < j \leq k$.
Consequently, our analysis relies on two key localized notions: $\Lip(\fhat^N \vert P)$, the local Lipschitz constant of the estimator restricted to a part $P \in \bP$, and $\mu\vert_P$, the data generating distribution restricted to $P$, defined via
$
    \mu|_P(\cdot)\coloneqq\mu(\cdot \cap P)/\mu(P).
$ 
The localized empirical distribution can be similarly defined as
$
    \mu^N|_P(\cdot)\coloneqq \mu^N(\cdot \cap P) \mu^N(P).
$
We note that $\mu^N(P) = N_P/N$ where $N_P \coloneqq \vert \{ \calD^N \cap P \}\vert$ counts the number of samples which fall into the set $P$. 
We are now ready to present our instance-dependent bound on the generalization error. 
This theorem is a corollary of our main result of \cref{thm:concentration_main}, and \cref{app:proof_thm_generalization} presents its proof.

\begin{theorem}[Generalization error of Lipschitz estimators]\label{thm:generalization}
Let $\fhat^N$ be a learned function which may depend on the dataset $\calD^N$. Suppose Assumptions~\ref{assum:domain},~\ref{assum:estimator},~and~\ref{assum:loss} hold with some $L_\Ls, L_\fhat>0$. 
For any $0< \delta\le 1$ and any data-independent partitioning $\bP$ of $\calX \times \calY$, we have
\begin{equation*}
\begin{split}
  \mathfrak{R}(\fhat^N; \mu) -  \hat{\mathfrak{R}}(\fhat^N)  \leq \, \transcost(\bP) + \transerror(\bP) + \parterror(\bP)
\end{split}
\end{equation*}
with probability greater than $1-\delta$, where
\begin{align*}
\transcost(\bP) & \coloneqq \frac{C_{d+1,1}L_\Ls}{N}\sum_{P \in \bP} N_P^{\tfrac{d}{d+1}} \max\left\{1,  \Lip(\fhat^N \vert P_\calX) \right\}\diam(P),\\
\transerror(\bP) &\coloneqq \sqrt{\frac{ \ln(4/\delta)}{N}} L_\ell \max\{1, L_\fhat \} 
\max_{P \in \bP} \diam(P),\\
    \parterror(\bP) & \coloneqq  \begin{cases}
     \norm{\Ls}_\infty \max\left\{\sqrt{\frac{2\ln(4/\delta)}{N}},  \sqrt{\frac{\vert \bP \vert}{N}}\right\}\quad & \vert \bP \vert>1\\
    0 & \vert \bP \vert=1 \;
    \end{cases}
\end{align*}
Here $P_\calX$ denotes the projection of $P$ onto $\calX$, and the constant $C_{d+1,1}$ is recorded in \cref{tab:concentration_main_table_version__FULL}.
\end{theorem}
\looseness -1 The generalization gap is the discrepancy in calculating the expectation of the loss calculated with respect to the two distributions $\mu$ and $\mu^N$. Intuitively, our bound is based on the cost of transporting probability mass from $\mu^N$ to $\mu$. In particular, this cost accounts for how far we have to transport probability mass on average and how much the loss can change in the process.
We perform this transport-based analysis locally by partitioning $\calX$ and bounding the cost of changing the measure from $\mu\vert_P$ to $\mu^N\vert_P$ for every $P \in \bP$. 
Since the dataset $\calD^N$ is drawn at random, this cost is a random variable. The term $\transcost$ upper bounds the expected value of this cost, and the term $\transerror$ controls the deviation from the expected cost.
The last term, $\parterror$, denotes the cost we pay for partitioning, and it is equal to zero if $\bP = \{\calX \times \calY\}$.
The previous two terms account for transporting probability mass within parts of the domain. However, if  $\mu(P) \neq \mu^N(P)$, mass also needs to be transported across parts. $\parterror$ upper bounds the potential change in the risk due to this global transport of mass. Naturally, the more parts we have in our partitioning, the higher the $\parterror$.

\looseness -1 The error bound of \cref{thm:generalization} converges with $\calO(N^{-1/(d+1)})$ which, for higher dimensional domains, implies relatively slow convergence. However, this rate is already an improvement upon Rademacher generalization bounds for Lipschitz estimators (see \cref{sec:instance_dep}).
We do not impose any constraints on $\mu$ other than having compact support.
Thus, our bound holds for any $\mu \in \calP(\calX \times \calY)$, also unfavorable edge-cases such as a uniform distribution over the domain. Without further assumptions, the optimal transport cost (i.e., Wasserstein distance) of $\mu^N$ to $\mu$ inherently has an exponential dependence on the dimensionality of the domain. For sufficiently smooth functions, we can remove the exponential dependence on $d$ and obtain $\calO(N^{-1/2})$ convergence rates (see Theorem \ref{thm:concentration_main} and Table \ref{tab:concentration_main_table_version__FULL}).

In many applications with high-dimensional data domains, it has been postulated that the data lies on some low-dimensional manifold \citep{narayanan2010sample, fefferman2016testing}. For instance, \citet{pope2021the} empirically demonstrate the validity of this assumption on popular image datasets. Under the assumption that the data lies on a $\tilde d$-dimensional manifold where $\tilde d \ll d$, we can improve the convergence rate. Proposition~\ref{prop:manifold_generalization} shows that the generalization error would then {\em only} depend on the intrinsic dimension $\tilde d$. The proof is presented in \cref{app:proof_manifold}.

\begin{proposition}[Fast rates for structured data]\label{prop:manifold_generalization}
Consider the setting and assumptions of \cref{thm:generalization}. In addition, suppose $\mu$ is such that the data lies almost surely on a $\tilde d$-dimensional $C^1$-Riemannian manifold.
Then for any $0<\delta\leq 1$ and any data-independent partitioning $\bP$ on $\calX \times \calY$, there exists $C(\tilde d)$ for which
\begin{equation*}
\begin{split}
  \mathfrak{R}(\fhat^N; \mu) -  \hat{\mathfrak{R}}(\fhat^N)  \leq \,&
  \frac{C(\tilde d) L_\Ls}{N}  \sum_{P \in \bP} N_P^{1-1/\tilde d} \max\left\{1,  \Lip(\fhat^N \vert P_\calX) \right\}\diam(P)\\
  &  \,\,+ \transerror(\bP) + \parterror(\bP)
\end{split}
\end{equation*}
with probability $1-\delta$ where the terms
$\transerror$ and $\parterror$ are as defined in \cref{thm:generalization}.
\end{proposition}

\looseness-1 Bounds of Theorem~\ref{thm:generalization} and Proposition~\ref{prop:manifold_generalization} can be made tighter by directly considering $\Lip(\Ls \circ \fhat^N | P)$ the local Lipschitz regularity of the {\em composition} of the loss and the prediction function. 
In fact, a direct instantiation of \cref{thm:concentration_main} would result in a bound that depends on this quantity. However, for a clearer exposition, we split the Lipschitz constants of $\Ls \circ \fhat^N$ using $\Lip(\Ls \circ \fhat^N | P) \leq L_\ell \cdot \text{Lip}(\fhat^N | P)$, and present the bounds in terms of the global Lipschitz constant of the loss. In our numerical experiments, we use the tighter variant based on $\text{Lip}(\Ls \circ \fhat^N | P)$.

\section{Key Properties of the Generalization Bound: NN Perspective}\label{sec:props}

\cref{thm:generalization} presents a generalization bound that 
captures the post-training properties of the learned prediction function $\fhat^N$. In this section, we elaborate on these properties with a focus on neural networks. 
As an empirical running example, we consider two simple supervised learning tasks. 
We generate synthetic random datasets for $1$D regression and $2$D binary classification (\cref{fig:reg_class_problem}) and train over-parametrized fully-connected ReLU networks on them with stochastic gradient descent (SGD). We then evaluate the bound of \cref{thm:generalization} for the resulting estimator.\footnote{More precisely, we visualize the tighter variant of Theorem~\ref{thm:generalization} which directly depends on $\Lip(\Ls \circ \fhat^N | P)$, since splitting the constant as $L_\Ls \cdot \Lip( \fhat^N | P)$ may loosen the bound.}
Details of the experiments are reserved for \cref{app:experiments}.
To calculate the local Lipschitz constant $\Lip(\fhat\vert P)$, we simply consider a fine grid of the domain and evaluate the gradient of the network over this mesh. This only requires light computations since our toy examples are two-dimensional at most. For higher dimensional domains, 
\citet{jordan2020exactly} and \citet{fazlyab2019efficient} propose scalable algorithms that approximate the local Lipschitz constant of a neural network.
For the regression task, we use the Huber loss (Equation~\ref{eq:huber_loss}). Since the Huber loss has a Lipschitz constant of $L_\ell = 1$, \cref{thm:generalization} applies directly to the regression case.
For the binary classification, we use the labels $\calY = \{-1, 1\}$ and aim to bound the expected classification error $\sP(\fhat^N(X) \neq Y)$. The 0-1 classification error $\mathbf{1}(\fhat^N(x) \cdot y < 0)$ is not Lipschitz. However, following \citet{hajek2019ece}, we use the ramp loss
\begin{equation} \label{ramp_loss}
    \Ls_\gamma(y_1, y_2) \coloneqq \min\left\{1, \left(1- \tfrac{y_1y_2}{\gamma}\right)_+ \right\} \;, ~ \text{with} ~ \gamma > 0~, 
\end{equation} 
as a Lipschitz proxy and upper bound of the 0-1 loss. This allows us to obtain a corollary of \cref{thm:generalization} which upper bounds the classification error:

\begin{corollary}[Classification error bound]\label{cor:classification_bound}
Consider a compact input domain and labels in $\calY = \{ -1, 1\}$. Assume that the observation noise is i.i.d.~and may only flip the label.
Let  $\gamma >0$, $\bP$ be any partitioning of size $k$ on $\calX$, independent of the data $\calD^N$. Then under Assumption~\ref{assum:loss}, with probability greater than $1-\delta$,
 \begin{equation*}
\begin{split}
    \sP(\fhat^N(X) \neq Y)\leq &  \frac{1}{N}\sum_{i=1}^N \Ls_\gamma(\fhat^N(X_i), Y_i) +  \frac{2^{1/d} C_{d,1}}{\gamma} \sum_{P \in \bP} \frac{N_P^{1-1/d}}{N} \Lip(\fhat^N \vert P) \diam(P)\\
    &+   \sqrt{\frac{\ln(4/\delta)}{N}} \frac{L_\fhat}{\gamma}  \max_{P \in \bP} \diam(P) + \sqrt{\frac{2 }{N}} \max\left\{\sqrt{\ln(4/\delta)}, \sqrt{k}\right\}
\end{split}
\end{equation*}
\looseness-1 here \smash{$N_P = \left\vert \left\{(X, Y)\in \calD^N \, \text{ s.t. }X \in P\right\}\right\vert$} counts the number of samples that lie in partition $P$.
\end{corollary}
The proof of Corollary~\ref{cor:classification_bound} is given in \cref{app:proof_cor_classification_bound}.
Since for classification $\calY$ is only a finite set, we can marginally reduce the dimension dependence of the generic bound from $\calO(N^{1-1/(d+1)})$ to $\calO(N^{1-1/d})$.

\subsection{Instance-Dependent vs. Fixed Hypothesis Classes}\label{sec:instance_dep}
Understanding\added{the}generalization of over-parametrized neural networks requires analyzing the \emph{combination} of model architecture, initialization method, and training procedure.  
A trained network $\fhat^N$ inherits the joint effect of the three elements. 
Therefore, instance-dependent bounds, which are calculated for $\fhat^N$ post-hoc are more informative in describing the generalization behavior of over-parametrized networks.\added{In particular, Theorem \ref{thm:generalization} can be understood as adapting the sub-hypothesis space in each partition $P$ to the local Lipschitz regularity $L(\fhat^N|P)$.
The adaptive hypothesis space that arises from the intersection of the sub-hypothesis spaces across partitions is tailored to the trained function $\fhat^N$ and is typically much smaller than the hypothesis space considered a-priori.}\deleted{Such bounds also allow the practitioner to predict the test error of the model once it is trained and can be used as a certificate for model selection between a finite number of estimators.}\added{On the other hand, bounds that}only reflect the properties of the\added{function/hypothesis class prior to the training}are known to be vacuous for large hypothesis classes such as neural networks \citep{bartlett2021failures, golowich2020size}.

We empirically evaluate our generalization bound when applied to neural networks trained on regression and classification tasks. Figures~\ref{fig:bounds_num_nn_params}~and~\ref{fig:num_parts} show that contrary to the majority of prior works, which yield vacuous bounds for over-parametrized neural networks, our bound assumes values in the same order of magnitude as the expected error and becomes non-vacuous for around $N > 10000$ classification examples.

\begin{figure}[ht]
     \centering
     \begin{subfigure}[b]{0.32\textwidth}
         \centering
         \includegraphics[width=\textwidth]{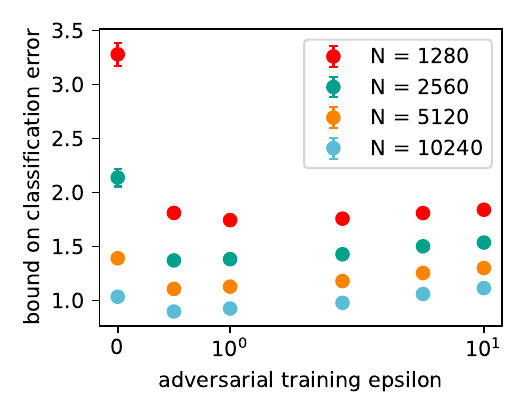}
         \caption{\textsc{AT} for Classification}
         \label{fig:class_avd_training}
     \end{subfigure}
     \hfill
     \begin{subfigure}[b]{0.32\textwidth}
         \centering
         \includegraphics[width=\textwidth]{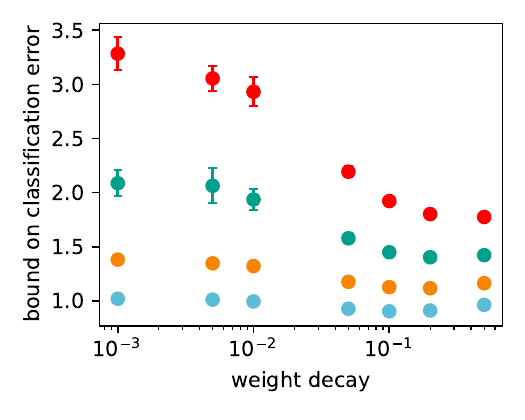}
         \caption{\textsc{WD} for Classification}
         \label{fig:class_weight_decay}
     \end{subfigure}
          \hfill
       \begin{subfigure}[b]{0.32\textwidth}
         \centering
         \includegraphics[width=\textwidth]{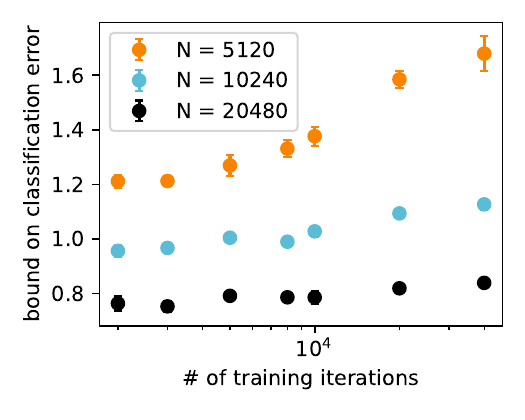}
         \caption{\textsc{ES} for Classification}
         \label{fig:class_early_stopping}
     \end{subfigure}
     \begin{subfigure}[b]{0.32\textwidth}
         \centering
         \includegraphics[width=\textwidth]{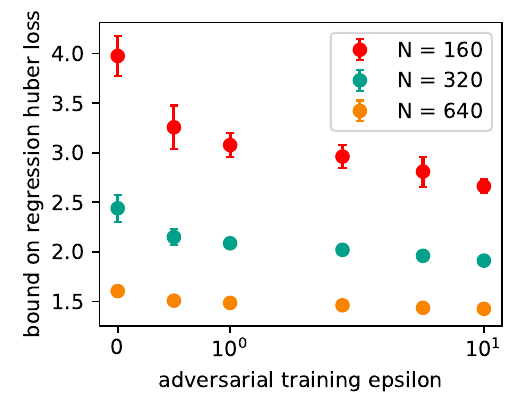}
        \caption{\textsc{AT} for Regression}
         \label{fig:reg_avd_training}
     \end{subfigure}
     \hfill
     \begin{subfigure}[b]{0.32\textwidth}
         \centering
         \includegraphics[width=\textwidth]{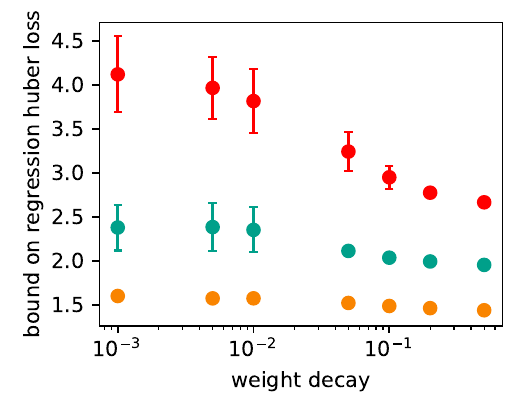}
         \caption{\textsc{WD} for Classification}
         \label{fig:reg_weight_decay}
     \end{subfigure}
          \hfill
       \begin{subfigure}[b]{0.32\textwidth}
         \centering
         \includegraphics[width=\textwidth]{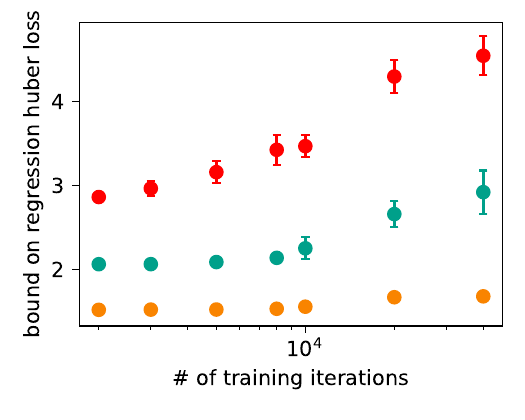}
         \caption{\textsc{ES} for Regression}
         \label{fig:reg_early_stopping}
     \end{subfigure}
        \caption{Effect of adversarial training (\textsc{AT}), weight decay (\textsc{WD}) and early stopping (\textsc{ES}). Generalization bounds of \cref{thm:generalization} and Corollary~\ref{cor:classification_bound} suggest that these training techniques result in networks that generalize better. See \cref{app:exp_gen} for details of the plots.}
        \label{fig:regularization}
\end{figure}

Furthermore, our instance-dependent bound reflects the positive effect of common regularization techniques on the generalization performance.
It is known that certain training techniques, such as adversarial training, weight decay, and early stopping, can lead stochastic gradient descent to solutions that generalize better. \cref{fig:regularization} illustrates the effect of these methods on our bound when applied to neural networks. 
As we can observe in \cref{fig:regularization}, the bound improves once the aforementioned regularization techniques are employed during training. In particular, for smaller sample sizes, the change in the value of the bound suggests that all three methods of adversarial training, weight decay, and early stopping produce networks that tend to generalize better. 
This observation matches the prior works of \citet{xing2021generalization}, \citet{krogh1991simple} and \citet{earlystopping2020} respectively. This empirically supports our core idea that, since \cref{thm:generalization} directly depends on the learned neural network instance $\fhat^N$, it is able to capture the joint effect of model structure, initialization, and training. \looseness -1

\subsection{Geometric vs. Parametric Characterization of the Estimator}\label{sec:non-param}
We characterize the estimator via its local Lipschitz regularity.
A key idea in our work is that this local geometry has an immediate effect on the generalization ability of the network compared to the network size or architecture.
An alternative approach are parametric bounds which consider the network structure. 
Such bounds are data-dependent and are often a function of the Frobenius norm of the network's weights, i.e., $\norm{\bm{W}_j}_F$ where $1\leq j\leq l$ indexes the layer number. 
Examples are the Rademacher-type bound of \citet{neyshabur2015norm}, \citet{bartlett2017spectrally}, which follows a covering number argument, and \citet{neyshabur2018pac}, which takes a PAC-Bayesian approach. 
These norm-based bounds roughly grow with $\calO(\diam(\calX)\mathrm{Poly}(d, h,l)\prod_{j=1}^l \norm{\bm{W}_j}_F\sqrt{1/N})$, where $h$ denotes the width of the network.\footnote{Not all the bounds have this polynomial dependency, e.g. the bound of \citet{neyshabur2015norm} depends on the network depth exponentially.}
\citet{golowich2020size} improve prior results and present a bounds of the rate $\calO(\diam(\calX)\prod_{j=1}^l \norm{\bm{W}_j}_F\sqrt{l/N})$. 
While these bounds have a polynomial dependency on the input dimension $d$, they quickly become vacuous for larger networks due to their polynomial dependence on network size. 
Therefore such bounds fail to capture or explain the benefit of over-parametrization \citep{bartlett2021failures}, in contrary to the observation that large neural networks tend to generalize better in practice \citep{zhang2017understanding}.
A key issue with parametric analysis is that there are combinatorially many parameter configurations, or in cases even entire sub-spaces that correspond to the same neural network mapping. This artificially inflates the hypothesis space compared to the set of neural network functions that are actually considered by the learning algorithm.

\begin{figure}
     \centering
     \begin{subfigure}[b]{0.48\textwidth}
         \centering
         \includegraphics[width=\textwidth]{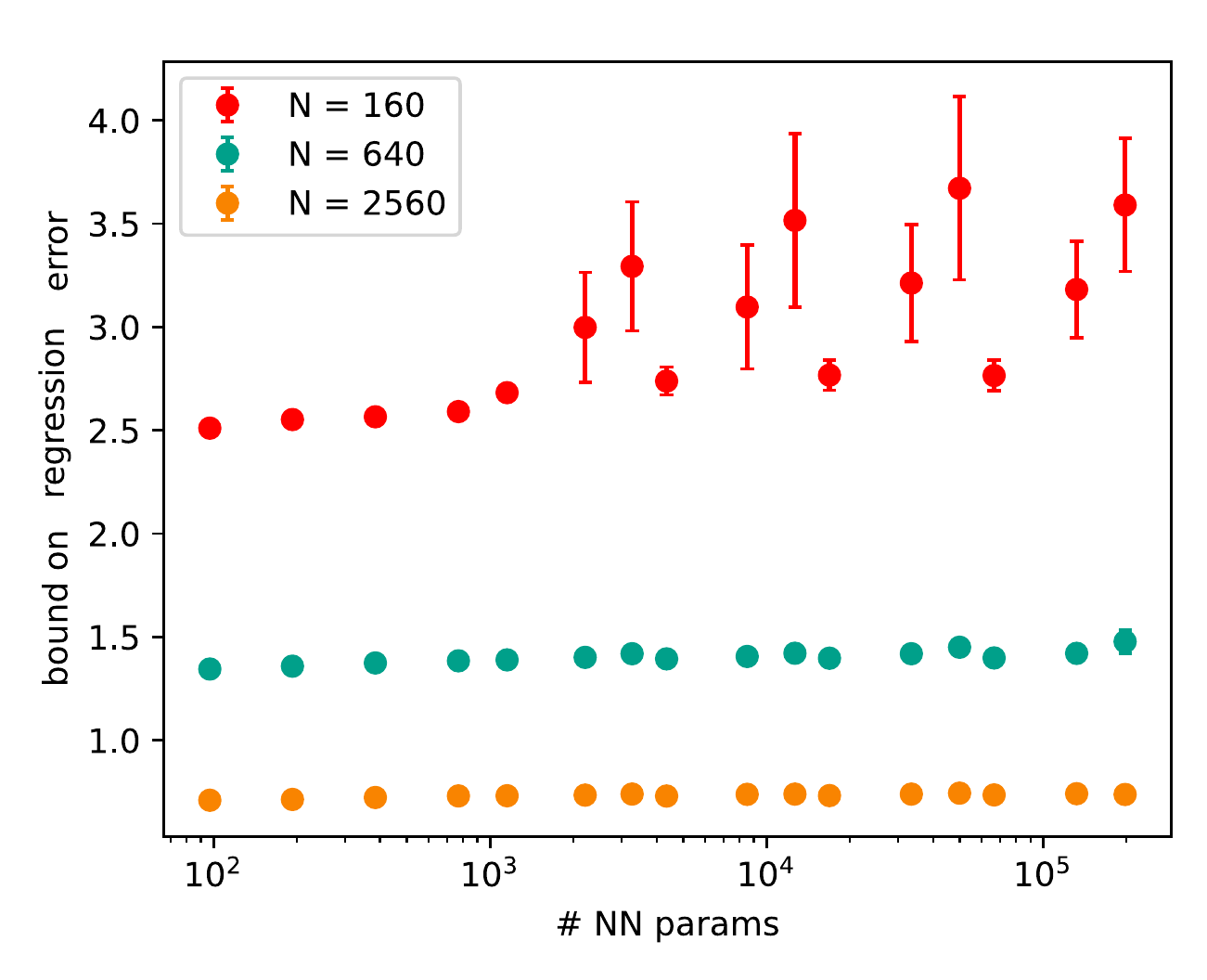}
         \caption{Regression}
         \label{fig:reg_nn_size}
     \end{subfigure}
     \hfill
     \begin{subfigure}[b]{0.48\textwidth}
         \centering
         \includegraphics[width=\textwidth]{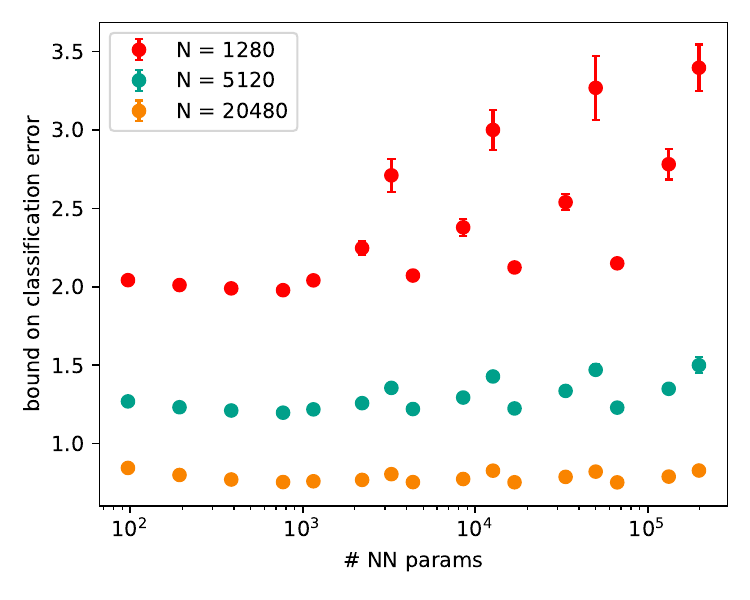}
         \caption{Classification}
         \label{fig:class_nn_size}
     \end{subfigure}
        \caption{Generalization error bound vs. number of neural network parameters. The neural network size has only a minor effect on the bound values. See \cref{app:exp_nnsize} for details.}
        \label{fig:bounds_num_nn_params}
\end{figure}

Our bound is based on the geometry of the learned function rather than its parameters, and it does not suffer from the described issue.
We observe that increasing the network size has negligible effects on the local regularity of the learned prediction function.  Thus, the generalization bound of \cref{thm:generalization} hardly grows with the neural network size.
To demonstrate this empirically, we train networks of increasingly larger width and depth and calculate the corresponding bounds. \cref{fig:bounds_num_nn_params} shows that even increasing the number of neural network parameters by factor 1000 has only a minor effect on the value of the bound, in particular when the dataset size is large. Our geometry-based approach seems to capture the empirically observed generalization behavior of over-parametrized neural networks better than previous bounds.

\subsection{Localized vs.\ Global Analysis}\label{sec:localized}

\begin{figure}
     \centering
     \begin{subfigure}[b]{0.48\textwidth}
         \centering
         \includegraphics[width=\textwidth]{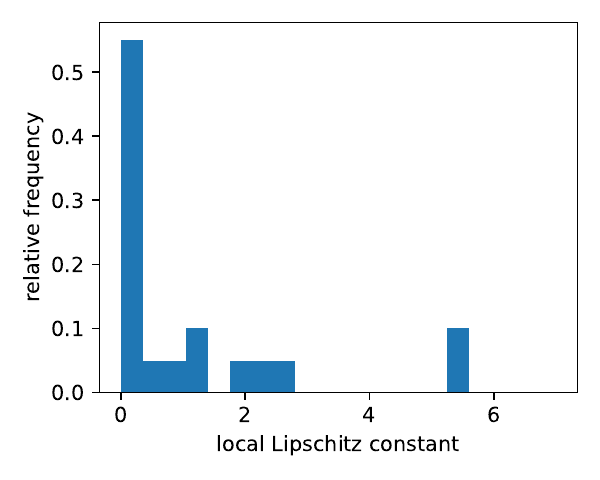}
         \caption{Regression}
     \end{subfigure}
     \hfill
     \begin{subfigure}[b]{0.48\textwidth}
         \centering
         \includegraphics[width=\textwidth]{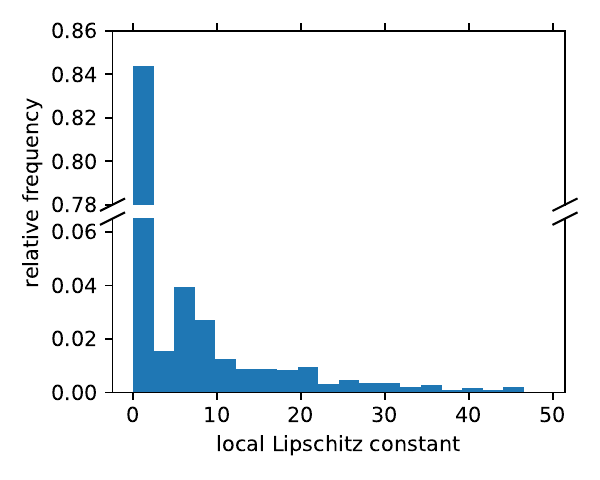}
         \caption{Classification}
     \end{subfigure}
        \caption{Frequency of local Lipschitz constant values per part $P \in \bP$ for fully-connected ReLU nets trained with SGD. The local Lipschitz constant is small in the majority of parts $P$, contrary to the large global constant. For training details, see \cref{app:exp_training}.}
        \label{fig:lipschitz_histograms}
\end{figure}
\begin{figure}
    \centering 
    \includegraphics[width=\textwidth]{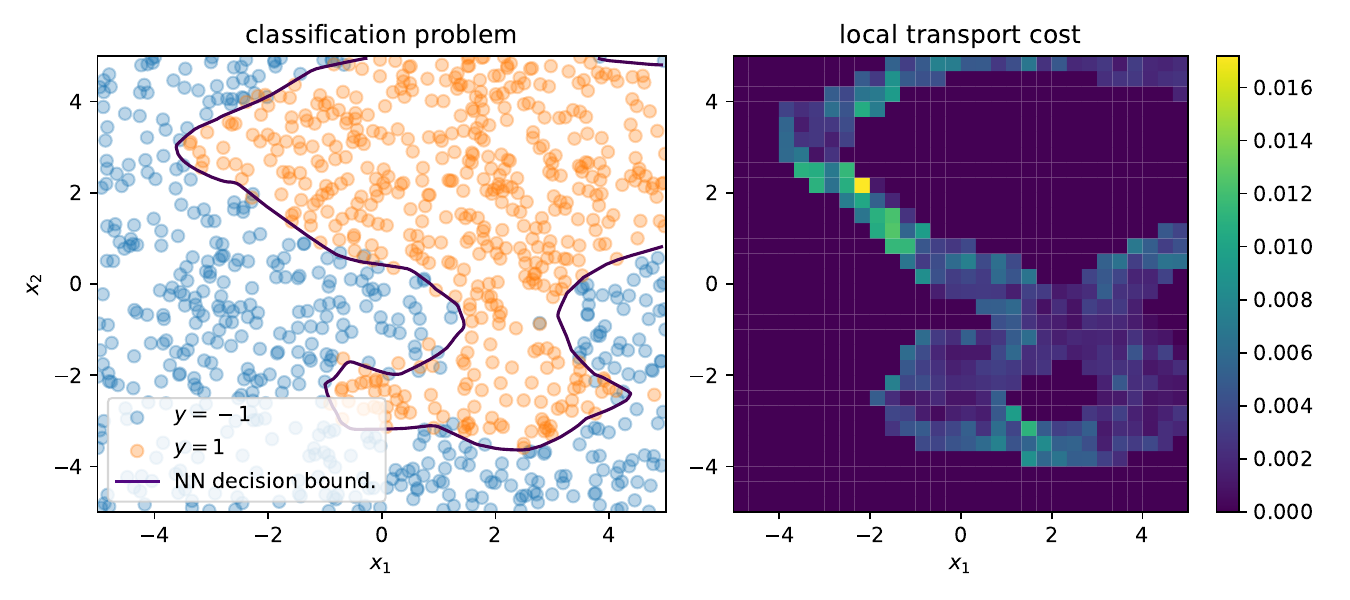}
    \caption{Local break down of $\transcost$. Areas closer to the decision boundary of $\fhat^N$ contribute more to the bound. Each cell $P$ in the heatmap shows the local transport cost $\Lip(\Ls \circ \fhat^N \vert P)\diam(P)\sqrt{N_P}/N$.}
    \label{fig:local_transport_cost}
\end{figure}

In \cref{thm:generalization}, we partition the domain into many subsets and locally bound the generalization risk when the domain is restricted to each subset. As a result, $\transcost$, which is often the largest term of the bound and shrinks with $N$ the slowest, is independent of the global Lipschitz constant and depends on $\fhat^N$ only through $\Lip(\fhat^N\vert P)$. 
When applied to neural networks, \cref{thm:generalization} tends to benefit from this localized analysis, as the local regularity of trained networks strongly varies across the domain.
Figure \ref{fig:lipschitz_histograms} displays the distribution of local Lipschitz constants across parts $P \in \bP$.
We observe that $\Lip(\fhat^N|P)$ is fairly low in the majority of parts, while there exist a few parts with much higher local Lipschitz constant, which contributes to a large overall global constant.
Therefore, we expect our localized analysis to be more informative than a global argument which treats all parts uniformly and in turn, produces a bound depending on $\diam(\calX)\Lip(\fhat^N)$. 
\cref{fig:local_transport_cost} empirically supports this claim by visualizing the local generalization error restricted to each part $P$, which changes strongly across the neighborhoods. In particular, we observe that the local generalization error is small in areas away from the decision boundary of $\fhat^N$, and that it achieves its maximum in an area where the estimator misclassifies the training data.

\begin{figure}
     \centering
     \begin{subfigure}[b]{0.48\textwidth}
         \centering
         \includegraphics[width=\textwidth]{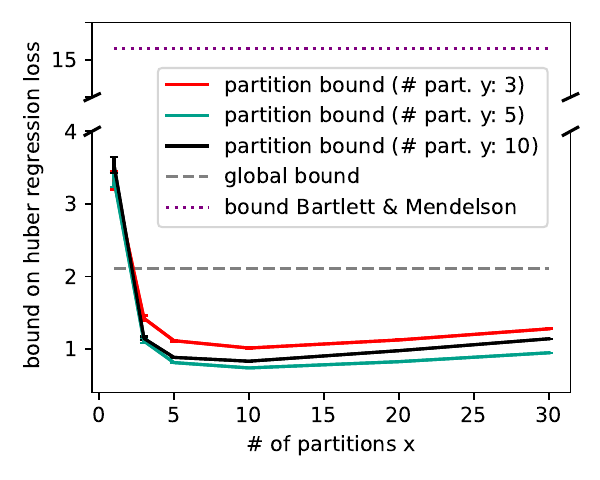}
         \caption{Regression. All curves correspond to $N = 2560$ samples.}
         \label{fig:num_parts_reg}
     \end{subfigure}
     \hfill
     \begin{subfigure}[b]{0.48\textwidth}
         \centering
         \includegraphics[width=\textwidth]{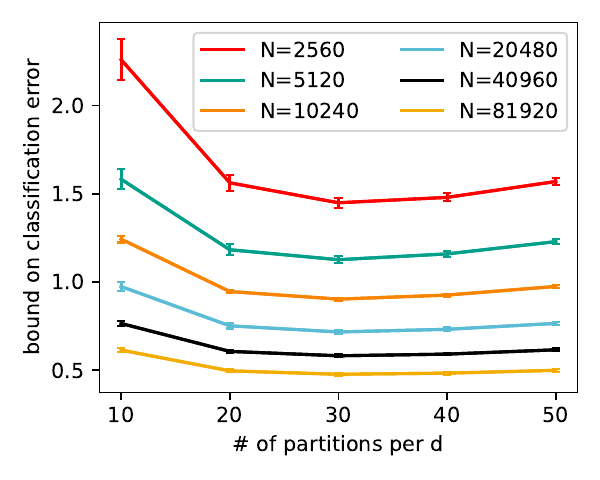}
         \caption{Classification. Depending on $N$, the global bound is 8-20 times larger (see Table \ref{tab:classification_bound_comparison}).}
         \label{fig:num_parts_class}
     \end{subfigure}
        \caption{Generalization error bound values for a varying number of partitions. Left: Bound on the Huber regression loss for differing numbers of partitions in $\mathcal{X}$ and $\mathcal{Y}$, together with the global Lipschitz bound in (\ref{eq:global_bound}) the bound of \citet{JMLR:bartlett2002rademacher}. Right: Bound on the classification error for differing numbers of partitions per dimension of $\calX$. For more details see \cref{app:exp_part}.}
        \label{fig:num_parts}
\end{figure}

As we partition the data domain into ever finer parts, two forces are at play:
The diameter of each part $P$ shrinks, and the local Lipschitz constant may get smaller, making $\transcost$ and $\transerror$ in \cref{thm:generalization} shrink. At the same time, as parts become smaller, mismatches in probability mass of $\mu$ and $\mu^N$ become more pronounced so that we have to account for more transport of mass across the partitions, increasing the $\parterror$ term. Hence, by making partitioning finer, we trade off $\transcost$ and $\transerror$ against $\parterror$. 
Figure \ref{fig:num_parts} displays our bounds in response to anincreasing partition size. We can empirically observe the the trade-off as the bound values initially decrease and, as the partitioning becomes much fines, increase again. Since the marginal gains from a finer partitioning decrease, there is typically a sweet spot, i.e., a degree partitioning that leads to the tightest bounds.

\begin{table}[ht!]
    \centering
{
\raa{1.3}
\begin{adjustbox}{width=\columnwidth,center}
\begin{tabular}{@{}rlll@{}}
\toprule
 \# \textbf{Train Data} & \textbf{Partition Bound (ours)} &  \textbf{Global Bound (in \ref{eq:global_bound})} & \textbf{Bound of B\&M (2002)}\\
\midrule
  2560 &    1.447 $\pm$ 0.028 &  18.407 $\pm$ 3.609 &         39.545 $\pm$ 2.951 \\
  5120 &    1.126 $\pm$ 0.019 &  10.234 $\pm$ 0.933 &         30.622 $\pm$ 1.096 \\
   10240 &    0.903 $\pm$ 0.009 &   7.023 $\pm$ 0.398 &         25.743 $\pm$ 0.606 \\
   20480 &    0.717 $\pm$ 0.012 &   5.232 $\pm$ 0.378 &         22.365 $\pm$ 0.665 \\
   40960 &    0.582 $\pm$ 0.008 &   3.813 $\pm$ 0.282 &         19.246 $\pm$ 0.600 \\
     81920 &    0.478 $\pm$ 0.009 &   2.856 $\pm$ 0.265 &         16.738 $\pm$ 0.658 \\
\bottomrule
\end{tabular}
\end{adjustbox}
}
    \caption{
    Comparison of bounds on the classification error. For the partition bound, we report to lowest bound values across a varying number of partitions. Our partition-based bounds are $5$ to $10$ times smaller than the global bound in (\ref{eq:global_bound}) and at-least $20$ times smaller than the uniform bound of \citet{JMLR:bartlett2002rademacher}.}
    \label{tab:classification_bound_comparison}
\end{table}

Without partitioning, or equivalently by considering only one \emph{global} partition  $\bP_\mathrm{global}  \coloneqq \{ \calX\times \calY\}$, we obtain the \emph{global} counterpart of \cref{thm:generalization}:
\begin{equation}
\label{eq:global_bound}
  \mathfrak{R}(\fhat^N; \mu) - \hat{\mathfrak{R}}(\fhat^N) \leq \transcost(\bP_\mathrm{global}) + \transerror(\bP_\mathrm{global}).
\end{equation}
We can also compare \cref{thm:generalization} with a classic uniform law on the generalization of Lipschitz estimators since our only assumption about the estimator is almost sure Lipschitz continuity. 
Let $\calF_L$ denote the class of $L$-Lipschitz functions mapping $\calX$ to $\calY$, and recall that $\calD^N$ is an i.i.d. random sample of size $N$ drawn according to the probability distribution $\mu$. 
Under Assumption~\ref{assum:domain}~and~\ref{assum:loss}, the Rademacher generalization bound \citep[Theorem 9,][]{JMLR:bartlett2002rademacher} implies that, with probability $1-\delta$ there exists $C>0$ for which \emph{every}
 $f \in \calF_L$ satisfies
\begin{equation}\label{eq:rademacher_bound}
    \mathfrak{R}(f ; \mu) - \hat{\mathfrak{R}}(f) \leq C L_\Ls \left(\frac{\left(\diam(\calX)L\right)^dd^2D^2}{N}\right)^{1/(d+3)} + \norm{\Ls}_\infty \sqrt{\frac{8\log 2/\delta}{N}}
\end{equation}
where $D \coloneqq \sup_{f \in \calF_L} \norm{f}_\infty$. In \cref{app:rademacher}, we formalize this statement, calculate $C$, and provide a proof for completeness. The first term on the right-hand-side of \eqref{eq:rademacher_bound}, which corresponds to the Rademacher complexity of $\calF_L$, dominates this bound. It rapidly grows for large high-dimensional domains or for a large Lipschitz constant and converges at a $O(N^{-1/(d+3)})$ rate.
\cref{thm:generalization} only marginally improves upon this rate since it converges with $O(N^{-1/(d+1)})$. 
However, the value of the constants is significantly smaller. 
The term $\transcost$ has the slowest decay with $N$, and its constant is proportional to $\Lip(\fhat^N \vert P)\diam(P)$.
Consequently, for a typical estimator $\fhat^N$, \cref{thm:generalization} yields a tighter bound that that of  
\eqref{eq:global_bound} and \eqref{eq:rademacher_bound}.
\added{This is further demonstrated in \cref{fig:num_parts}, which visualizes the bound values corresponding to \cref{thm:generalization} for a varying number of partitions and compares it to the global bound in \eqref{eq:global_bound}, as well as to the global Rademacher bound of \cref{eq:rademacher_bound}. 
For classification, we compare the bound values in \cref{tab:classification_bound_comparison}.  In particular, for the classification task, the global bound (8-20 times larger), as well as the Rademacher bound (> 20 times larger), are vacuous, while partitioning allows us to obtain non-vacuous guarantees.}

Crucially, there exist estimators for which the generalization error of \cref{thm:generalization} is bounded and vanishes as $N \rightarrow \infty$, while the global inequality \eqref{eq:global_bound} diverges.  
In Proposition~\ref{prop_existence_blowup_ReLUs}, we construct such an estimator. Perhaps surprisingly, we prove that the global bound diverges already for a shallow ReLU network with exactly one neuron defined over $\calX=[0,1]$, while a localized analysis with a partition with a size of order $\vert\bP\vert = \calO(N^{0.6})$ gives a vanishing error bound.
The proof is presented in \cref{app:proof_Corollary_Paritions_imporant}.

\begin{proposition}[Partitioning can help]
\label{prop_existence_blowup_ReLUs}
Let $\calX = [0,1]$ and $\calY = [0,1]$. 
For any probability measure $\mu\in\calP(\calX\times\calY)$, there is an increasing sequence of Lipschitz constants $\{L_{f,N}\}_{N=1}^{\infty}$, a sequence of partitions $\{\boldsymbol{P}_{N}\}_{N=1}^{\infty}$, and a sequence of ReLU feedforward networks with one neuron $\{f_N\}_{N=1}^{\infty}$, such that the local bound of \cref{thm:generalization} for the tuple $(L_{f, N}, \bP_N, f_N)$ converges:
\[
\lim_{N\to\infty} \transcost(\bP_N) + \transerror(\bP_N) + \parterror(\bP_N) \to 0
\]
while the global bound of Equation \eqref{eq:global_bound} diverges:
\[
 \lim_{N\to\infty} \transcost\left(\bP_\mathrm{global}\right)  + \transerror\left(\bP_\mathrm{global}\right) \to \infty.
\]
where $\bP_\mathrm{global}  = \{[0,1]\times [0,1] \}$, and the terms $\transcost, \transerror$ and $\parterror$ are defined as in \cref{thm:generalization}.
\end{proposition}

\section{Generalization under Distribution Shifts}\label{sec:adversarial}
A desirable characteristic of an estimator $\fhat^N$ is robustness to changes in the data generating distribution $\mu$ between training and test time. A change in $\mu$ may be due to covariate shifts, adversarial attacks, or small changes in the data-generating process over time. In safety-critical applications such as perception systems in self-driving cars or models for medical diagnosis, it is crucial that we can certify the performance of $\fhat^N$ under changes in the distribution. 
In this section, we employ our framework to obtain an instance-dependent generalization bound under distribution shift.
In particular, we bound the risk calculated with respect to a shifted distribution $\muadv$, by the {\em training} error of the estimator on a dataset of size $N$ which is sampled from data generating distribution $\mu$.

In addition to the transport of mass from $\mu^N$ to $\mu$ which is at the core of \cref{thm:generalization}, our optimal transport-based approach allows us to seamlessly consider the additional change of measure from $\mu$ to $\muadv$.
In particular, we use the Wassertstein-1 distance $\WD(\mu, \muadv)$ to quantify the amount of distribution shift. This is consistent with previous work on robustness certificates which are often given for distributions within an $\epsilon$-Wasserstein ball centered in the data generating distribution \citep{sinha2017certifying,lee2018minimax, blanchet2019quantifying,levine2020wasserstein,gao2022distributionally}. 
The Wasserstein-1 distance is defined as the minimum $\ell_1$-cost of transporting probability mass from $\nu_1$ to $\nu_2$, that  \looseness-1
\[
\WD(\nu_1, \nu_2) \coloneqq \inf_{\gamma \in \Gamma(\nu_1, \nu_2)} \int_{\calZ \times \calZ} \norm{x-x^\prime}_1 d(\gamma(x, x^\prime))
\]
where $\Gamma(\nu_1, \nu_2) \subset \calP(\calZ \times \calZ)$ denotes the set of all couplings between $\nu_1$ and $\nu_2$, in other words, the set of joint distributions whose marginals are $\nu_1$ and $\nu_2$. 

Corollary~\ref{cor:adv_robust} presents our instance-dependent generalization bound under distribution shift. For simplicity, we present this result in the global case of $\bP_\mathrm{global} = \{\calX \times \calY\}$, i.e., without partitioning.
However, the analysis can also be carried out locally with partitioning analogous to \cref{thm:generalization}. The proof is given in \cref{app:proof_cor_adv_robust}.

\begin{corollary}[Locally Lipschitz estimators are robust to distribution shift]\label{cor:adv_robust} 

Let \\$\muadv \in \calP(\calX\times \calY)$.
Under Assumptions~\ref{assum:domain},~\ref{assum:estimator},~and~\ref{assum:loss} with probability $\geq 1-\delta$, we have
\begin{equation*}
    \begin{split}
   \mathfrak{R}(\fhat^N; \muadv) -  \hat{\mathfrak{R}}(\fhat^N) \leq & ~
   \transcost\left(\bP_\mathrm{global}\right) + \transerror(\bP_\mathrm{global}) \\
   &\,+ L_\ell \max\left(1,  \Lip(\fhat^N) \right)\calW_1\left(\mu, \muadv\right)
\end{split}
\end{equation*}
where $\bP$ denotes a data-independent partition on $\calX$, and $(\transcost, \transerror)$ are identical to that of \cref{thm:generalization}.
\end{corollary}

\looseness-1 The bound of Corollary~\ref{cor:adv_robust} has the same generalization terms as the global bound in Equation \eqref{eq:global_bound}, plus an additional term which accounts for the potentially negative impact of the distribution shift by multiplying $\calW_1\left(\mu, \muadv\right)$ with the Lipschitz constants of the $\ell$ and $\hat{f}^N$.
In particular, if $\WD(\mu, \muadv) \leq \epsilon$, then the corollary implies that in the worst case, the estimator suffers from a $\epsilon L_\Ls\Lip(\fhat^N)$ increase in risk when evaluated on the perturbed distribution. 

There are many connections between Corollary~\ref{cor:adv_robust} and prior work on robust estimators. 
\citet{gao2022finite} verifies distributionally robust learnability of $\calF_L$ the class of Lipschitz functions through a uniform bound. We expect this bound to be vacuous if calculated empirically, since it has large terms depending on the complexity of the function class, e.g., via Rademacher complexity or metric entropy.
\citet{kuhn2019wasserstein} and \citet{cranko2021generalised} bound the difference between the finite-sample validation error of an estimator $f$, and $\mathfrak{R}(f; \muadv)$. Both works use this robustness certificate to develop methods for distributionally robust optimization.
Perhaps, closest to our result, is \citet{mohajerin2018data} who give a  generalization bound for a data-dependent, robust estimator defined via
\[
\fhat^N_\epsilon = \argmin_{f \in \calF_L} \max_{\WD(\mu^N, \muadv) \leq \epsilon} \mathcal{R}(f; \muadv).
\]
\looseness -1 
Taking a similar approach, \citep{staib2019distributionally} consider the data-dependent solution of an analogous minimax problem, where $\calF_L$ is replaced with a Reproducing Kernel Hilbert Space, and subsequently the Maximum Mean Discrepancy is used instead of the Wasserstein distance.
In contrast to these works, Corollary~\ref{cor:adv_robust} holds for {\em any} data-dependent Lipschitz estimator. 
In practice, we have access to the training error $\hat{\mathfrak{R}}(\fhat^N)$ and aim to verify how this performance generalizes to unseen data generated from a shifted distribution ($\mathfrak{R}(f; \muadv)$). Therefore, instance-dependent generalization bounds such as Corollary~\ref{cor:adv_robust} or \citet{mohajerin2018data} are of more practical relevance, compared to uniform \citep{gao2022finite} or deviation bounds \citep{kuhn2019wasserstein}. 
Corollary~\ref{cor:adv_robust} further suggests that (locally) Lipschitz estimators tend to be more robust towards distribution shifts, and contribute to prior results connecting distributional or adversarial robustness to Lipschitzness \citep{cisse2017parseval,finlay2018lipschitz, sinha2017certifying, anil2019sorting}.
Corollary~\ref{cor:adv_robust} does not depend on the number of model parameters. Hence, it gives a powerful guarantee when applied to over-parametrized neural networks, in particular when the data lies on a low-dimensional manifold. 
Therefore, it acts as an advocate for training methods that effectively regularize the Lipschitz constant of the network, \citep[e.g.,][]{bartlett2017spectrally, cisse2017parseval, anil2019sorting, Sagawa2020Distributionally}.

\section{Main Result}\label{sec:general_bounds}
\looseness-1 Our main result is a transport- and partitioning-basedconcentration inequality, which states that the empirical mean of a sample-dependent function concentrates around its expectation if the function satisfies some degree of regularity. 
Let $\calZ \subset \sR^{d_\calZ}$ denote the domain, and consider functions $g: \calZ \rightarrow \sR$. 
We work with two classes of regular functions, smooth and non-smooth. 
The $\calC^s$-smooth class identifies functions that admit all partial derivatives up to order $s$, for an $s\in \mathbb{N}_+$. The smoothness of a $\calC^s$-smooth function $g$, when restricted to $P \subset \calZ$, is quantified by
\begin{equation*}
    \|g\|_{s:P}
        :=
    \max_{| \beta | \leq s} \max_{z\in P} 
    \Big | 
        \frac{
            \partial^{|\beta|} g(z)
        }{
            \partial_{\beta_1}
                \dots
            \partial_{\beta_{d_\calZ}} 
        }  
    \Big |
,
\end{equation*}
where $\beta \in \nn^{d_\calZ}$ is a multi-index and $|\beta| = \beta_1+\dots+\beta_{d_\calZ}$. Further, when $P = \calZ$, we simplify this notation to $\|g\|_{s}$. 
In machine learning applications, examples of smooth estimators include Gaussian processes with smooth kernels, physics-informed neural networks \citep{RAISSI2019686}, Neural-ODE solvers \citep{chen2018neural}, invertible neural networks \cite{JMLR:v22:18-803}, and feedforward networks with smooth activation functions \citep{DERYCK2021732}.
For the non-smooth category, i.e., functions that are not differentiable, we use the $\alpha$-H\"{o}lder property as a geometric notion for quantifying regularity. More formally, for $0<\alpha\le 1$, a function $g$ is $\alpha$-H\"{o}lder if
\begin{equation*}
    \Lip_{\alpha}(g|P) \coloneqq \sup_{\substack{x_1,x_2\in P\\ x_1\neq x_2}}\, \frac{|g(x_1)-g(x_2)|}{\|x_1-x_2\|_2^{\alpha}}.
\end{equation*}
is finite.
For $\alpha = 1$ this recovers the case of Lipschitz functions, which are discussed in the previous sections. \added{Further, when $P = \calZ$, we simplify this notation to $\Lip_{\alpha}(g)$ and, when $\alpha = 1$, we simplify the notation to $\Lip(g|P)$.}Setting $\alpha \in (0,1)$\edit{produces}{allows for} rougher models which are typically used for\edit{predicting}{making predictions} from long time-series \citep{roughtneuralmodels_PatrickTerry_2021} or rough paths of Neural-SDE models \citep{risks8040101}.\added{Further, we consider two classes of integral probability metrics to quantify how rapidly an empirical measure with i.i.d.\ samples from the data-generating distribution $\mu$, concentrates around $\mu$. Each metric evaluates the dissimilarity between any two probability measures as the worst-case average dissimilarity over all functions in a given class, e.g.,\ H\"{o}lder or Smooth functions. This metric is equivalent to the Wasserstein-1 distance when defined over the class of Lipschitz (i.e., 1-Hölder) functions. We consider function classes of different regularity since, from the universal approximation standpoint, the performance of deep learning models is known to vary significantly across different smoothness classes \cite{yarotsky2020phase,guhring2021approximation}.  A discussion is provided in Section~\ref{sec:Discussion}.}

\added{
\begin{definition}
\label{defn:Walphas}
Let $\calZ \subseteq \sR^{d_{\calZ}}$. For $\alpha \in (0,1]$, we define the $\alpha$-H\"{o}lder Wasserstein distance between two probability measures $\mu, \nu \in \calP(\calZ)$ as
\begin{equation*}
    \calW_{\alpha}(\mu,\nu)
        \eqdef 
    \sup_{g:\,  \Lip_{\alpha}(g)\leq 1}\, \int g(z)\,\mu(dz) - \int g(z)\,\nu(dz).
\end{equation*}
For $s \in [1,\infty)$, we define the $s$-smooth Wasserstein distance between two probability measures $\mu, \nu \in \calP(\calZ)$ as
\begin{equation*}
    \calW_{\calC^{s}}(\mu,\nu)
        \eqdef 
    \sup_{g:\,  \Vert g \Vert_{s}\leq 1}\, \int g(z)\,\mu(dz) - \int g(z)\,\nu(dz)
    .
\end{equation*}
\end{definition}
}

\cref{thm:concentration_main} formalizes our main result. In \cref{sec:proof_sketch}, we sketch the proof for \cref{thm:concentration_main} for the non-smooth case to highlight the main techniques. The complete proof can be found in \cref{app:proof_main}. 
\begin{theorem}%
\label{thm:concentration_main}
Set $0 < \delta\leq 1$, and $N \in \sN$. 
Let $\calZ \subseteq \sR^{d_{\calZ}}$ be a compact set, $\mu\in \mathcal{P}(\calZ)$ be a probability measure, 
and $\bP$ a data-independent partitioning of any size $k \in \sN$ on $\calZ$. 
Suppose $g^{N}: \calZ \mapsto \R$ is a real-valued random function that may depend on $Z_1,\dots,Z_N$, which are samples drawn independently from $\mu$. 
For $\alpha \in (0,1]$, define
\[
\totalerror(\alpha) \coloneqq \sqrt{\frac{ \ln(4/\delta)}{N}} L\max_{P \in \bP} \diam(P)^{\alpha} + \frac{\Vert g^N \Vert_{\infty} }{\sqrt{N}} \max\{\sqrt{2\ln(4/\delta)}, \sqrt{k}\}.
\]
\begin{enumerate}
    \item[(i)] \textbf{Non-Smooth:} Set $0<\alpha\leq 1$, and let $\calF_{L, \alpha} = \{g\in \calC(\calZ,\R)\colon \Lip_{\alpha}(g) \leq L\}$. 
    Suppose $g^{N} \in \calF_{L, \alpha}$ almost surely. Then with probability greater than $1-\delta$, we have
    \begin{equation*}
    \sE\big[g^N(Z)\big] -   \frac{1}{N}\sum_{n=1}^N g^N(Z_n) \leq  C_{d_{\calZ},\alpha}\sum_{P \in \bP} \frac{N_P}{N}\mathrm{rate}_{d_{\calZ},\alpha}(N_P) \diam(P)\mathrm{Lip}_{\alpha}(g^{N}\vert P) + \totalerror(\alpha)
    \end{equation*}
    where $\mathrm{rate}_{d_{\calZ},\alpha}$ and $C_{d_{\calZ},\alpha}$  
    depend only on the H\"{o}lder coefficient $\alpha$ and on the dimension $d_{\calZ}$. The explicit expressions
    are recorded in Table~\ref{tab:concentration_main_table_version__FULL}.
    \item[(ii)] \textbf{Smooth:} Set $s \geq 1$, and let $\calF_{L, s} = \{g\in \calC^{s}(\calZ,\R)\colon \Vert g \Vert_{s} \leq L\}$. 
    Suppose $g^{N} \in \calF_{L, s}$ almost surely. Then there exists constant $C_{d_\calZ, s} >0$ which with probability greater than $1-\delta$ satisfies
    \begin{equation*}
    \sE\big[g^N(Z)\big] -   \frac{1}{N}\sum_{n=1}^N g^N(Z_n) \leq  C_{d_{\calZ},s}\sum_{P \in \bP} \frac{N_P}{N} \mathrm{rate}_{d_{\calZ},s}(N_P) \diam(P)\Vert g^N \Vert_{s:P} + \totalerror(1)
    \end{equation*}
    where $\mathrm{rate}_{d_{\calZ},s}$ depends only on $s$ and on $d_{\calZ}$ and is recorded in Table~\ref{tab:concentration_main_table_version__FULL}.
\end{enumerate}
\end{theorem}

\begin{table}[H]
    \centering
    \raa{1.3}
    \begin{adjustbox}{width=\columnwidth,center}
    \begin{tabular}{@{}llll@{}}
    \toprule
    \textbf{Regularity} & \textbf{Dimension} & \textbf{Rate} ($\boldsymbol{\mathrm{rate}_{d_{\calZ},\alpha}}$, \textbf{or} $\boldsymbol{\mathrm{rate}_{d_{\calZ},s}}$) & \textbf{Constant} ($\boldsymbol{C_{d_{\calZ},\alpha}}$ \textbf{or} $\boldsymbol{C_{d_{\calZ},s}}$ )
    \\
    \midrule
    \multirow{3}{*}{$\alpha$-H\"{o}lder} & $d_{\calZ}<2 \alpha$ & $N_P^{-1/2}$ & $C_{d_{\calZ},\alpha} = \frac{d_{\calZ}^{\alpha/2} 2^{d_{\calZ}/2 - 2\alpha}}{1- 2^{d_{\calZ}/2-\alpha}}$ \\
    & $d_{\calZ}=2 \alpha$  &  $\big(\alpha 2^{\alpha + 2} + \log_2(N_P)\big)N_P^{-1/2}$ & $C_{d_{\calZ},\alpha}=\frac{d_{\calZ}^{\alpha/2}}{\alpha 2^{\alpha + 1}}$ 
    \\
    & $d_{\calZ}>2 \alpha$  & $N_P^{-\alpha/d_{\calZ}}$ & $C_{d_{\calZ},\alpha} = 2\Big(\frac{\frac{d_{\calZ}}{2} - \alpha}{2 \alpha (1-2^{\alpha-d_{\calZ}/2})}\Big)^{2\alpha/d_{\calZ}}\Big(1 + \frac{\alpha}{2^{\alpha}(\frac{d_{\calZ}}{2} - \alpha)}\Big)\,d_{\calZ}^{\alpha/2}$ \\
    \midrule
    \multirow{3}{*}{$s$-Smooth} & $s>\frac{d_{\calZ}}{2}$ & $N_P^{-1/2}$ & $\exists\, C_{d_{\calZ},s}>0$ \\
    & $s=\frac{d_{\calZ}}{2}$  & $\big(\log(N_P)+1)N_P^{-1/2}\big)$ & $\exists\, C_{d_{\calZ},s}>0$ \\
    & $s<\frac{d_{\calZ}}{2}$  & $N_P^{-s/d_{\calZ}}$ & $\exists\, C_{d_{\calZ},s}>0$ \\
    \bottomrule
    \end{tabular}
    \end{adjustbox}
    \caption{Rates and constants for \cref{thm:concentration_main}}
    \label{tab:concentration_main_table_version__FULL}
\end{table}

The generalization bounds of Sections~\ref{sec:gen_bounds},~\ref{sec:props}~and~\ref{sec:adversarial} all follow from \cref{thm:concentration_main} in the non-smooth case with $\alpha = 1$, so that the $\alpha$-Hölder regularity coincides with Lipschitzness. Since we are concerned with the loss of a machine learning estimator, we use $g^N(Z) = g^N(X, Y) =  \ell(\hat{f}^N(X), Y)$ to obtain the risk bounds.

\subsection{Optimal Transport Interpretation}
\label{sec:Discussion}

\added{
The risk bounds derived in Theorem~\ref{thm:concentration_main} case (i) for $\alpha$-H\"{o}lder functions, where $0 < \alpha \leq 1$ are closely connected to optimal transport theory.  In particular, the Kantorovich-Rubinstein duality \citep[e.g. see][Theorem 11.8.2]{DudleyRealProb_2002Book} links the $\mathcal{W}_1$ distance from Definition~\ref{defn:Walphas}, to the optimal transport problem as follows: for any $\mu,\nu \in \calP(\calZ)$ we have
\begin{equation}
\label{eq:duality_KR}
        \mathcal{W}_1(\mu,\nu)
    =
        \inf_{\pi \in \Pi(\mu,\nu)}\,
            \mathbb{E}_{(X,Y)\sim \pi}\big[\norm{X - Y}\big]
\end{equation}
where $\Pi(\mu,\nu)$ is the set of couplings of $\mu$ and $\nu$. A coupling $\pi \in \calP(\calZ \times \calZ)$ of $\mu$ and $\nu$ is a joint-probability measure whose marginals coincide with $\mu$ and $\nu$.  The right-hand side of~\eqref{eq:duality_KR} is the standard definition of the $1$-Wasserstein distance (e.g., \citet[Definition 6.1]{villani2009optimal}).}

\added{A crucial property that allows us to similarly connect $\mathcal{W}_{\alpha}$ for $0 < \alpha \leq 1$ to optimal transport is the following: A function $f:\calZ \rightarrow \mathbb{R}$ is $\alpha$-H\"{o}lder when $\mathcal{Z}$ is equipped with the Euclidean metric $\operatorname{dist}(x,y) \eqdef  \norm{x - x}$, where $x,y\in \mathcal{Z}$, if and only if $f$ is Lipschitz with respect to the \textit{snowflaked Euclidean metric} on $\mathcal{Z}$ defined for any $x,y\in \mathcal{Z}$ by
\[
    \operatorname{dist}_\alpha(x,y) \eqdef  \norm{x,y}^\alpha
\]
\citep[see][Proposition 2.52]{WeaverLipschitzAlgebras_2ed_2018}.  Since $(\mathcal{Z},\operatorname{dist}_{\alpha})$ is a metric space, the Kantorovich-Rubinstein duality for $(\mathcal{Z},\operatorname{dist}_{\alpha})$ implies that
\begin{equation}
\label{eq:snwoflaked_W1}
            \mathcal{W}_{\alpha}(\mu,\nu) = 
        \inf_{\pi \in \Pi(\mu,\nu)}\,
            \mathbb{E}_{(X,Y)\sim \pi}\big[
               \norm{X-Y}^{\alpha}
            \big]
\end{equation}
for any $\mu,\nu \in \calP_{\alpha}(\calZ)$, where $\calP_{\alpha}(\calZ)$ is the set of Borel probability measures on $\calZ$ for which $\mathbb{E}_{X\sim \nu}[\norm{X-z}^{\alpha}]<\infty ~ \forall z\in \calZ$. 
Since $\calZ$ is compact, every Borel probability measure on $\calZ$ belongs to $\calP_{\alpha}(\calZ)$ and, thus, ~\eqref{eq:snwoflaked_W1} holds for any $\mu,\nu\in\mathcal{P}(\mathcal{Z})$. Hence, under mild regularity conditions, the $\alpha$-H\"{o}lder Wasserstein distance from Definition~\ref{defn:Walphas} directly corresponds to the optimal transport (aka.\ Wasserstein) distance with transportation cost given by $\operatorname{dist}_\alpha(X, Y)$.}

\subsection{Proof outline for Theorem~\ref{thm:concentration_main}}\label{sec:proof_sketch}
\looseness -1 Since the function $g^N$ is dependent on $\calD^N$, we can not directly invoke concentration inequalities for bounded i.i.d. random variables such as, e.g., Hoeffding's inequality. Potentially, $g^N$ is over-fitted to the dataset $\calD^N$ such that the empirical risk is much smaller than the expected risk. Thus, our analysis has to consider relevant properties of $g^N$ that capture how well we can expect $g^N$ to generalize beyond the training data $\calD^N$.

Crucially, given arbitrary but fixed measures $\nu_1=\mu$ and $\nu_2=\mu^N$,\added{by Definition~\ref{defn:Walphas}
\begin{equation}\label{eq:IPM}
    \int_{z \in \calZ} g(z)\,\nu_1(dz)
    \leq 
    \Lip_{\alpha}(g) 
    \mathcal{W}_{\alpha}(\nu_1,\nu_2)
        + 
    \int_{z \in \calZ} g(z)\,\nu_2(dz),
\end{equation}
for any $g$ s.t. $\Lip_{\alpha}(g) < \infty$, including any function that depends on $\mu^N$.} The Lipschitz constant $\text{Lip}(g^N)$ measures how regular $g^N$ is and quantifies the maximum change of $g$ when moving probability mass. Since the local regularity of $g^N$ may vary strongly throughout the domain, we use a partitioning $\bP$ of the space $\calZ$ to obtain a {\em localized} variant of the change of measure inequality. Then we separately bound the sum of local transport costs as well as the potential error from partitioning with high probability.
In the following, we elaborate on the three main steps of this proof. \looseness -1

\paragraph{Step 1: Local change of measure.}
To localize the change of measure inequality, we use a partitioning $\bP$ of the space $\calZ$. In particular, we invoke \eqref{eq:IPM} independently on each part $P \in \bP$  by restricting the domain to $P$ and using the lemma with the corresponding restricted measures $\nu_1 = \mu|_P$ and $\nu_2 = \mu^N|_P$.
Taking a sum over all $P \in \bP$ results in
\begin{equation*} \label{eq:com}
\begin{split}
     \int_{z\in \calZ}\, g^N(z)\,\mu(dz)
\leq \int_{z\in \calZ}\, g^N(z)\,\mu^{N}(dz) &+ 
    \overbrace{\sum_{P\in \bP}
            \mu^{N}(P)
        \Lip_{\alpha}(g^N|P) 
        \calW_{\alpha}(\mu|_P,\mu^{N}|_P)}^{(\mathrm{I})}\\
    &+
    \underbrace{\sum_{P\in \bP}
        \bigg(1-\frac{
            \mu^{N}(P)
        }{
            \mu(P)
        }\bigg)
        \int_{z \in P}\, g^N(z)\,\mu(dz)}_{(\mathrm{II})}
.
\end{split}
\end{equation*}
Here, term (I) bounds the change of expectation from locally moving mass within each partition. Term (II) appears due to the potential mismatch in probability mass of $\mu(P)$ and $\mu^N(P)$ in the parts of the partitioning of $\calZ$ and thus accounts for probability mass that would have to be re-distributed across parts. We bound (I) and (II) separately.
\paragraph{Step 2: Bounding (I).} To bound (I), the crucial element is to upper-bound the Wasserstein distance between $\mu$ and its empirical counterpart $\mu^N$. Since $\mu^N$ is based on samples drawn independently from $\mu$, $\mathcal{W}_{\alpha}(\mu,\mu^N)$ is bounded in expectation and concentrates as more samples are taken into account. 
More formally, we show that due to \citet[Theorem 2.1]{MR4153634}, for all $\epsilon > 0$, and $N\in\nn$ it holds that
\begin{equation*}
    \sP \Bigg(\bigg\vert \mathcal{W}_{\alpha}(\mu,\mu^N) - \sE\Big[\mathcal{W}_{\alpha}(\mu,\mu^N)\Big] \bigg\vert \geq \epsilon\Bigg)  \leq 2e^{-\frac{2N\epsilon^2}{\diam(\calZ)^{2\alpha}}},
\end{equation*}
and 
\begin{equation*}
    \sE\Big[\mathcal{W}_{\alpha}(\mu,\mu^N)\Big] \leq  C_{d_{\calZ},\alpha}\,\diam(\calZ)\, \mathrm{rate}_{d_{\calZ},\alpha}(N)
\end{equation*}
where the rates and the constants are as in \cref{tab:concentration_main_table_version__FULL}. We apply the above inequalities locally, by considering $\calW_{\alpha}(\mu|_P,\mu^{N}|_P)$ and sum over all $P \in \bP$. We show that this summation is a weighted sum of independent sub-Gaussian random variables, which we control via Lemma~\ref{lem:subgaussian2}. This treatment results in an upper bound for (I). 
\paragraph{Step 3: Bounding (II).}  We interpret this term as the penalty we face for partitioning. It is zero if the probability mass of the data generating and empirical measures match across the partitioning, i.e. $\mu(P) = \mu^N(P)$ for all $P \in \bP$, or if the analysis is carried out globally, i.e. $\bP = \{\calZ\}$. 
Considering discretized measures $\tilde \mu, \tilde \mu^N \in \calP(\bP)$, which satisfy $\tilde\mu(\{P\}) = \mu(P)$ and $\tilde\mu^N(\{P\}) = \mu^N(P)$, we use that
\begin{equation*}
    \sum_{P\in \bP}
        \bigg(1 - \frac{
            \mu^{N}(P)
        }{
            \mu(P)
        }\bigg)
        \int_{z \in P}\, g^N(z)\,\mu(dz) \leq \Vert g\Vert_\infty \mathrm{TV}(\tilde \mu, \tilde \mu^N)
\end{equation*}
where $\mathrm{TV}(\tilde \mu, \tilde \mu^N)$ denotes the total variation distance between $\tilde \mu^N$ and $\tilde \mu$. Since the $\mathrm{TV}(\tilde \mu, \tilde \mu^N)$ concentrates around zero and we can bound term II with high probability. 
Combining these three steps concludes the proof.

\section{Conclusion}

We presented novel instance-dependent generalization bounds for locally regular estimators. We empirically and theoretically demonstrated the benefits of an instance-dependent non-parametric bound and the effectiveness of a localized treatment of the risk. In particular, we showed that the instance-dependent bound remains relatively tight for over-parametrized models and captures a number of neural network generalization phenomena. In contrast, existing uniform or data-dependent parametric bounds tend to explode for large neural networks and fail to explain their good generalization behavior.

\looseness -1 Key observations made in this work could be relevant for future work that aims to improve learning algorithms.
For example, our result suggests that the \emph{local} regularity of a model plays a crucial role in its generalization ability and robustness to distribution shifts. This might be of interest for developing robust and regularized training techniques.
Finally, our optimal-transport-based approach constitutes a novel avenue towards theoretically analyzing generalization in machine learning. We introduce the necessary technical tools and proof methodology, hoping that future work can further explore this avenue and improve our results.

\acks{This research was supported by the European Research Council (ERC) under the European Union’s Horizon 2020 research and innovation program grant agreement no. 815943, by the NSERC grants no.\ RGPIN-2023-04482 and DGECR-2023-00230, and the European Research Council (ERC) Starting Grant 852821-SWING.  Anastasis Kratsios and Songyan Hou were additionally supported by the ETH Zurich foundation. Jonas Rothfuss was supported by an Apple Scholars in AI/ML fellowship. Anastasis was additionally funded by the McMaster University Startup Grant.}  
\newpage
\appendix
\numberwithin{equation}{section}

\section{Proofs}
\label{app:proofs}
\subsection{Proof of the Main Generalization Bound (Theorem~\ref{thm:generalization})}
\label{app:proof_thm_generalization}
\begin{proof}\textbf{of Theorem~\ref{thm:generalization}: }
Let $\calZ = (\calX,\calY)$, $d_{\calZ} = d+1$ and $g^N(x,y) = \ell(\fhat^{N}(x),y)$. Notice that $\nabla_{y}\,g^{N} = \nabla_{y_2}\,\ell \leq\, L_{\ell}$ and $\nabla_{x}\,g^{N} = \nabla_{y_1}\,\ell\cdot \nabla_{x}\fhat^{N} \leq L_{\ell}\cdot \Lip(\fhat^{N})$. So $\Lip(g^N|P) \leq L_{\ell}\max\{1,\Lip(\fhat^{N}|P_\calX)\} \leq L_{\ell}\max\{1,L_{f}\} $ for all $P\in\bP$. Now we can apply Theorem~\ref{thm:concentration_main} with $\alpha = 1$, $\calF_{L, 1} = \{g\in \calC(\calZ,\R)\colon \Lip(g) \leq L = L_{\ell}\max\{1,L_\fhat\}\}$. Then for all partitions $\bP$ with $|\bP| \leq k$, for all $0 < \delta \le 1$ and $N\in\nn$, there exists an explicit constant $C_{d_{\calZ},1} > 0$ s.t.  with probability greater than $1-\delta$
\begin{equation}
\begin{split}
    &\mathfrak{R}(\fhat^N; \mu) -  \hat{\mathfrak{R}}(\fhat^N) \\
    =~&\sE\big[g^N(Z)\big] -   \frac{1}{N}\sum_{n=1}^N g^N(Z_n)\\
    \leq~& C_{d_{\calZ},\alpha}\sum_{P \in \bP} \frac{N_P}{N}\mathrm{rate}_{d_{\calZ},\alpha}(N_P) \diam(P)\mathrm{Lip}(g^{N}\vert P) + \epsilon\\
    \leq~& C_{d_{\calZ},\alpha}\sum_{P \in \bP} \frac{N_P}{N}\mathrm{rate}_{d_{\calZ},\alpha}(N_P) \diam(P)L_{\ell}\max\{1,\Lip(\fhat^{N}|P_\calX)\} + \epsilon\\
    =~& \frac{C_{d+1,1}L_\Ls}{N}\sum_{P \in \bP} N_P^{\tfrac{d}{d+1}} \max\left\{1,  \Lip(\fhat^N \vert P_\calX) \right\}\diam(P) + \epsilon\\
    =~& \transcost + \epsilon,
\end{split}
\end{equation}
where
\begin{equation}
\begin{split}
    \epsilon &\coloneqq \sqrt{\frac{ \ln(4/\delta)}{N}} L_\ell \max\{1, L_\fhat \} 
\max_{P \in \bP} \diam(P) + \frac{\Vert \ell \Vert_{\infty}}{\sqrt{N}} \max\left\{\sqrt{2\ln(4/\delta)}, \sqrt{k}\right\}\\
    &= \transerror + \parterror.
\end{split}
\end{equation}
\end{proof}
\subsection{Proof of Generalization Bound on Manifold Domain (Proposition~\ref{prop:manifold_generalization})} \label{app:proof_manifold}
We start by proving the manifold extension of our main result in \cref{thm:concentration_main}. Then present the proof of \cref{prop:manifold_generalization} as a corollary of this theorem.

\begin{theorem}[Concentration of measure on a compact manifold]
    \label{thm:concentration_main_manifold}
    Set $0 < \delta\leq 1$, and $N \in \sN$. Let $\calZ$ be a $d_{\calZ}$-dimensional compact class $C^1$ Riemannian manifold. Let $\mu$ be a Borel probability measure on $\calZ$, and $\bP$ a partition of size $k$ on $\calZ$. Suppose $g^{N}$ is a real-valued random function on $\calZ$ depending on $Z_1,\dots,Z_N$. 
    Let $\calF_{L} = \{g\in \calC(\calZ,\R)\colon \Lip(g) \leq L\}$. 
        Suppose $g^{N} \in \calF_{L}$ almost surely. Then with probability greater than $1-\delta$
        \begin{equation}
            \sE\big[g^N(Z)\big] -   \frac{1}{N}\sum_{n=1}^N g^N(Z_n) \leq  C_{\calZ} \sum_{P \in \bP} \frac{N_P^{1-1/d_{\calZ}}}{N} \diam(P)\mathrm{Lip}(g^{N}\vert P) + \totalerror 
        \end{equation}
        where $C_{\calZ}>0$ is a constant depending on $d_\calZ$ and 
    \[
    \totalerror \coloneqq \sqrt{\frac{ \ln(4/\delta)}{N}} L\max_{P \in \bP} \diam(P) + \frac{\Vert g^N \Vert_{\infty} }{\sqrt{N}} \max\{\sqrt{2\ln(4/\delta)}, \sqrt{k}\}.
    \]
    \end{theorem}
    
\begin{proof}\textbf{of Theorem~\ref{thm:concentration_main_manifold}. }
All the steps are similar to the proof of \cref{thm:concentration_main}, and we only invoke a different Wasserstein concentration lemma. Deploying Lemma~\ref{lem:con_holder_wass_manifold} we find that there exists $C_{\calZ} > 0$ for which
\begin{equation*}
    \resizebox{0.95\hsize}{!}{%
        $\sum_{P\in \bP}
            \mu^{N}(P)
        \Lip(g^N|P)\,
        \E\big[\WD(\mu|_P,\mu^{N}|_P)\vert N_P\big]\leq~ C_{\calZ} \sum_{P\in \bP}
            \mu^{N}(P)
         \Lip(g^N|P) \diam(P) N_P^{-1/d_{\calZ}} $  
        }
\end{equation*}
and that
\begin{equation*}
    \sP \Big(\Big\vert\WD(\mu|_P,\mu^{N}|_P) - 
        \E\big[\WD(\mu|_P,\mu^{N}|_P)\vert N_P\big]\Big\vert \geq \epsilon\,\big\vert\, N_P\Big)  \leq 2e^{-\frac{2N_P\,\epsilon^2}{\diam(P)^2}}.
\end{equation*}
By defining $X_P$ similar to proof of \cref{thm:concentration_main} and invoking Lemma~\ref{lem:subgaussian2}, we get that there exists $C_{\calZ}>0$ such that the following holds with probability $1-\delta_1$
\begin{equation*}
\begin{split}
    \calB^N \leq  C_{\calZ}\sum_{P\in \bP} &
            \mu^{N}(P)
         \Lip(g^N|P) \diam(P) N_P^{-1/d_{\calZ}} \\
        & +  L\max_{P \in \bP} \diam(P)\left(\frac{\ln (2/\delta_1)}{N} \right)^{1/2}.
\end{split}
\end{equation*}
Terms $\calE^N$ and $\calR^N$ are identical to proof of \cref{thm:concentration_main}. Therefore, plugging in everything we get,
\begin{equation*}
\sE\big[g^N(Z)\big] -   \frac{1}{N}\sum_{n=1}^N g^N(Z_n) \leq C_{\calZ} \sum_{P \in \bP} \frac{N_P^{1-1/d_{\calZ}}}{N} \diam(P)\mathrm{Lip}(g^{N}\vert P) + \totalerror.
\end{equation*}
\end{proof}

\begin{proof}\textbf{of Proposition \ref{prop:manifold_generalization}}
The proof is nearly identical to \cref{thm:generalization}, however, here we invoke \cref{thm:concentration_main_manifold} instead of \cref{thm:concentration_main}. For completeness we repeat some of the steps.
Let $\calZ$ be the manifold which denotes the support of $\mu$, then $d_{\calZ} = \tilde d$. Let $g^N(x,y) = \ell(\fhat^{N}(x),y)$. 
We then apply \cref{thm:concentration_main_manifold} with $\calF_{L} \coloneqq \{g: \calZ\rightarrow \R \colon \Lip(g) \leq L = L_{\ell}\max\{1,L_\fhat\}\}$. Then for all partition $\bP$ with $|\bP| \leq k$, for all $0 < \delta \le 1$, $N\in\nn$, there exists an $C_{\calZ} > 0$ s.t. with probability greater than $1-\delta$
\begin{equation*}
\begin{split}
    \mathfrak{R}(\fhat^N; \mu) -  \hat{\mathfrak{R}}(\fhat^N) =~& \sE\big[g^N(Z)\big] - \frac{1}{N}\sum_{n=1}^N g^N(Z_n)\\
    \leq~&  C_{\calZ} \sum_{P \in \bP} \frac{N_P^{1-1/d_{\calZ}}}{N} \diam(P)\mathrm{Lip}(g^{N}\vert P) + \totalerror\\
    =~& \frac{C(\tilde d)L_\Ls}{N}  \sum_{P \in \bP} N_P^{1-1/\tilde d} \max\left\{1,  \Lip(\fhat^N \vert P_\calX) \right\}\diam(P)\\
    & \quad+ \transerror(\bP) + \parterror(\bP) 
\end{split}
\end{equation*}
where the last line is obtained by using the definition of $\transerror$ and $\parterror$ (as defined in \cref{thm:generalization}). In addition, to transparently show the dependency of $C_\calZ$ on $\tilde d$, we have renamed the constant. 
\end{proof}

\subsection{Proof of the Classification Error Bound (Corollary~\ref{cor:classification_bound})}
\label{app:proof_cor_classification_bound}
\begin{proof}\textbf{ of Corollary~\ref{cor:classification_bound}}
Let $g^N\big((x,y)\big) \coloneqq \Ls_\gamma(\fhat^N(x), y)$, where $y \in \{ -1, 1\}$ and $x \in \calX \subset \sR^{d}$.
Consider a partition $\bP \in \calX$ of size $k$. 
For all $ P \in  \bP$, we may define two sets
\[
 P_{(-)} \coloneqq \{(x, -1), \forall x \in  P\} \quad\quad  P_{(+)} \coloneqq \{(x, +1), \forall x \in  P\}.
\]
Note that $\diam(P_{(-)}) = \diam(P_{(+)}) = \diam(P)$. We construct the $\bP_\pm$ partition on $\calX \times \calY$ as
\[
\bP_\pm = \{ P_{(-)}\,\,\vert\,\, P \in \bP \} \cup \{  P_{(+)}  \,\,\vert\,\,P \in \bP \}.
\]
Note that $\vert \bP_\pm\vert = 2\vert \bP \vert$. For some $P_{(+)} \in \bP_\pm$, we calculate $\Lip(g \vert P_{(+)})$,
\begin{align*}
    \Lip(g \vert P_{(+)}) = \Lip\left(\Ls_\gamma\left(\fhat^N(\cdot), +1\right) \Big\vert P \right) \leq \frac{1}{\gamma} \Lip(\fhat^N \vert P )
\end{align*}
and similarly for any $P_{(-)} \in \bP_\pm$. 
Now invoking \cref{thm:concentration_main}, for $g$ and $\bP_\pm$ we get, with probability greater than $1-\delta$
\begin{align*}
 \mathfrak{R}_\gamma(\fhat^N; \mu)-  \hat{\mathfrak{R}}_\gamma(\fhat^N) & \leq  C_{d,1} \sum_{P_\pm \in \bP_\pm}
 \frac{N_{P_\pm}^{1-1/d}}{N} \diam(P_\pm)\Lip(g^N \vert P_{\pm}) + \totalerror\\
 & = C_{d,1} \sum_{P \in \bP} \Bigg[
 \frac{N_{P_{(+)}}^{1-1/d}}{N} \diam\left(P_{(+)}\right)\Lip(g^N \vert P_{(+)})\\
 & \quad\quad+ \frac{N_{P_{(-)}}^{1-1/d}}{N} \diam\left((P_{(-)}\right) \Lip(g^N \vert P_{(-)}) \Bigg] + \totalerror\\
 & \leq C_{d,1} \sum_{P \in \bP} 
 \frac{N_{P_{(+)}}^{1-1/d} + N_{P_{(-)}}^{1-1/d} }{N}\diam(P)\frac{\Lip(\fhat^N \vert P)}{\gamma} + \totalerror\\
 & \leq 2^{1/d}C_{d,1} \sum_{P \in \bP}
 \frac{N_P^{1-1/d}}{N} \diam(P)\frac{\Lip(\fhat^N \vert P)}{\gamma} + \totalerror
\end{align*}
where $N_P = N_{P_{(+)}} + N_{P_{(-)}}$ and
\begin{align*}
   \totalerror & =  \sqrt{\frac{ \ln(4/\delta)}{N}} \Lip(g^N) \max_{P_\pm \in \bP_\pm} \diam(P_\pm) + \frac{1}{\sqrt{N}} \max\{\sqrt{2\ln(4/\delta)}, \sqrt{\vert \bP_\pm\vert}\}\\
   & = \sqrt{\frac{ \ln(4/\delta)}{N}} \frac{\Lip(\fhat^N)}{\gamma} \max_{P \in \bP} \diam(P) + \frac{1}{\sqrt{N}} \max\{\sqrt{2\ln(4/\delta)}, \sqrt{2k}\}.
\end{align*}
The ramp loss acts as a Lipschitz proxy of the zero-one loss and allows us to 
analyze the classification error defined via
\[
\mathfrak{R}_{01}(\fhat^N; \mu) \coloneqq  \sE_{(X,Y)\sim \mu}\Ls_{01}\left(\fhat^N(X), Y\right)= \sP(\fhat^N(X) \neq Y).
\]
Here $\Ls_{01}(y_1, y_2) \coloneqq \bm{1}_{[y_1y_2 \leq 0]}$ is the zero-one loss which is not Lipschitz itself.
Finally, we note that $\Ls_\gamma \geq \Ls_{01}$, and therefore
\[
\mathfrak{R}_\gamma(\fhat^N; \mu) \geq \mathfrak{R}_{01}(\fhat^N; \mu) = \sP(\fhat^N(X) \neq Y)
\]
which implies,
\[
\sP(\fhat^N(X) \neq Y) \leq  \frac{1}{N} \sum_{i=1}^N \Ls_\gamma(\fhat^N(X_i), Y_i) + 2 C_{d,1} \sum_{P \in \bP}
 \frac{N_P^{1-1/d}}{N} \diam(P)\frac{\Lip(\fhat^N \vert P)}{\gamma} + \totalerror
 \]
with probability greater than $1-\delta$.
\end{proof}
\subsection{Proof of Proposition~\ref{prop_existence_blowup_ReLUs} on Partitioning}
\label{app:proof_Corollary_Paritions_imporant}

\begin{proof}\textbf{of Proposition~\ref{prop_existence_blowup_ReLUs}:}
Let $\sigma$ be ReLU activation function and, for every $N\in\sN$, consider the feedforward neural network $f_N$ with one layer and one neuron defined by
\begin{equation*}
    f_N(x) = \frac{\sqrt{N}}{\log_2(\log_2(N))} \cdot\sigma(x-1+\frac{1}{N}).
\end{equation*}
We directly compute the following ``global quantities'' associated to $f_N$:
\[
\Vert f_N \Vert_{\infty} \leq 1
    \mbox{ and }
\Lip(f_N) = \frac{\sqrt{N}}{\log_2(\log_2(N))}
.
\]
\begin{figure}[ht]
    \centering
    \includegraphics[width = 0.35\linewidth]{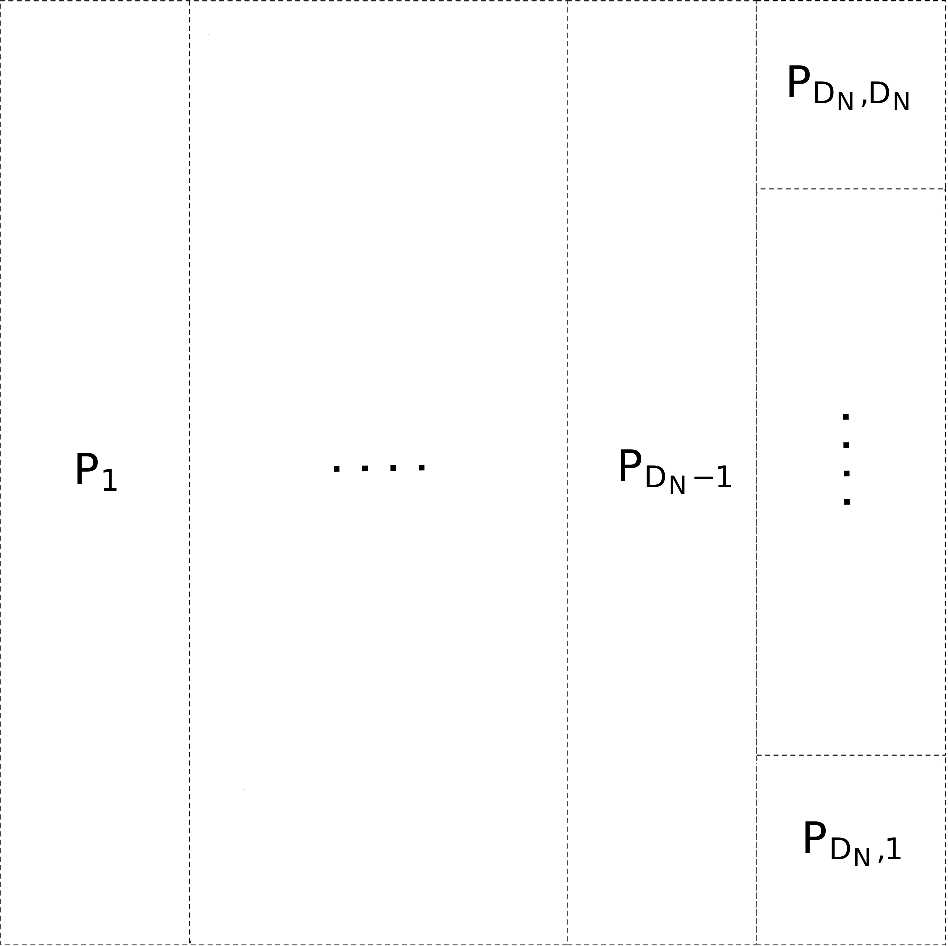}
    \caption{Partition $\bP_N$ used in proof of Proposition~\ref{prop_existence_blowup_ReLUs}}
    \label{fig:partitions}
\end{figure}
Set $L_{f,N} \eqdef \sqrt{N}$ and define each $\calF_N$ to be the set of Lipschitz functions from $[0,1]$ to itself with Lipschitz constant at-most $L_{f,N}$, as in Assumption~\ref{assum:estimator}.
We build a partition $\boldsymbol{P}_N$ of $[0,1]\times [0,1]$ as follows. Let $\Delta_N = N^{-0.6}$ and $D_N = \big\lceil\frac{1}{\Delta_N}\big\rceil = \calO(N^{0.6})$.
\begin{align*}
    \boldsymbol{P}_N &= \{B_1\times \calY, \dots, B_{D_N-1}\times\calY,B_{D_N}\times B_1,\dots,B_{D_N}\times B_{D_N}\}\\
    &= \{P_{1},\dots,P_n, \dots, P_{D_N-1},P_{D_N,1},\dots, P_{D_N, n}, \dots,P_{D_N,D_N}\}
\end{align*}
where the sets $B_1,\dots,B_{D_N}$ subdivide $[0,1]$ into $D_N$ intervals as defined by 
\begin{equation*}
B_n
:=
\begin{cases}
\big[\frac{n-1}{D_N},\frac{n}{D_N}\big)    
    : & n=1,\dots,D_N-1 \\
\big[\frac{D_N-1}{D_N},1\big]    
    : & n=D_N
\end{cases}
\end{equation*}
\cref{fig:partitions} illustrates $\bP_N$. This partition implies the following estimates on local Lipschitz constant of $f_N$
\begin{equation*}
\Lip\Big(f_N\Big\vert P_n\Big) = 0
    \mbox{ ~and~ }
        \Lip\Big(f_N\Big\vert P_{D_N,n}\Big) = \frac{\sqrt{N}}{\log_2(\log_2(N))},\quad  \forall n \in \sN_{< D_N}.
\end{equation*}
Furthermore, we compute the following partition related quantities
\begin{equation*}
    \max_{P \in \bP_N} \diam(P) \leq \sqrt{2} 
    \mbox{ and }
    \vert \bP_N\vert  = 2D_N-1 = \calO(N^{0.6}).
\end{equation*}
From the above quantities, we compute the ``localized bound'' of \cref{thm:generalization}, by calculating the terms $\parterror, \transcost, \transerror$. 
\begin{align*}
 \transcost & = \frac{C_{d+1,1}L_\Ls}{N}\sum_{P \in \bP} N_P^{\tfrac{d}{d+1}} \max\left\{1,  \Lip(f_N \vert P) \right\}\diam(P)\\
     & = C_{2,1}\sum_{P \in \bP_N} \frac{\big(8 + \log_2(N_P)\big)N_P^{1/2}}{N} \diam(P)L_{\ell}\max\{1,\Lip(f_N|P)\}\\
    & =C_{2,1}L_{\ell} \sum_{n=1}^{N}\frac{\big(8 + \log_2(N_{P_{D_N,n}})\big)N_{P_{D_N,n}}^{1/2}}{N} \frac{\sqrt{N}}{\log_2(\log_2(N))} \frac{\sqrt{2}}{D_N}\\
    & \leq \sqrt{2}C_{2,1}L_{\ell}\sum_{j=1}^{N}\frac{\big(8 + \log_2(1)\big)}{N}\frac{\sqrt{N}}{\log_2(\log_2(N))} \frac{1}{N^{0.6}}\qquad \text{(by Jensen's inequality)}\\
    & =\frac{8\sqrt{2}C_{2,1}L_{\ell}}{\log_2(\log_2(N))N^{0.1}}.
\end{align*}
For the other two terms,
\begin{align*}
    \transerror & \leq\sqrt{\frac{ \ln(4/\delta)}{N}} L_\ell \max\{1, L_\fhat \} 
\max_{P \in \bP} \sqrt{\diam(P)^2 +4B_y^2}\\
& \leq \sqrt{2}L_{\ell}\frac{\sqrt{N}}{\log_2(\log_2(N))}\frac{ \sqrt{\ln(4/\delta)}}{N^{1/2}} \\
     \parterror & = \frac{B_\Ls }{\sqrt{N}} \max\left\{\sqrt{2\ln(4/\delta)}, \sqrt{k_N}\right\}\\
    & \leq  \frac{\Vert \ell \Vert_{\infty} }{N^{1/2}} \max\{\sqrt{2\ln(4/\delta)}, \sqrt{\calO(N^{0.6})}\}   
\end{align*}
Therefore, in the $N\rightarrow \infty$ limit,
\[
\begin{aligned}
    \lim\limits_{N\rightarrow \infty}\transcost +  \transerror+ \parterror = 
    0
\end{aligned}
\]
and the ``local bound'' of Theorem~\ref{thm:generalization} converges. 
In contrast, upon inspecting the ``global bound'' of Equation \eqref{eq:global_bound}, we note that it is bounded below by the following quantity
\begin{equation}
\label{PROOF_prop_existence_blowup_ReLUs__GlobalBound}
\begin{aligned}
    \transcost =~&C_{2,1}\frac{\big(8 + \log_2(N)\big)}{N^{1/2}} \diam(\calX\times\calY)L_{\ell}\max\{1,\Lip(f_N)\} \\
    =~&\sqrt{2}C_{2,1}L_{\ell}\frac{\big(8 + \log_2(N)\big)}{N^{1/2}}\sqrt{N}\\
    =~&\sqrt{2}C_{2,1}L_{\ell}\big(8 + \log_2(N)\big).
\end{aligned}
\end{equation}
We conclude that the ``global bound'' diverges as $N$ approaches infinity, since the quantity in~\eqref{PROOF_prop_existence_blowup_ReLUs__GlobalBound} does; i.e.\
$
\lim\limits_{N\rightarrow \infty}\,
\sqrt{2}C_{2,1}L_{\ell}\big(8 + \log_2(N)\big) = \infty.
$  
\end{proof}

\subsection{Proof of Corollary~\ref{cor:adv_robust} on Robustness to Distribution Shifts}
\label{app:proof_cor_adv_robust}
\begin{proof}\textbf{of Corollary~\ref{cor:adv_robust}.} By Kantorovich Duality 
\citep[Theorem 5.10]{villani2009optimal}, we have
\begin{equation}
\label{eq:cor:adv_robust:1}
    \mathfrak{R}(\fhat^N; \muadv) \leq \mathfrak{R}(\fhat^N; \mu) + L_\ell \max\left(1,  \Lip(\fhat^N) \right)\calW\left(\mu, \muadv\right).
\end{equation}
By Theorem~\ref{thm:generalization}, we have for all $0 < \delta \le 1$ with probability greater than $1-\delta$,
\begin{equation}
\label{eq:cor:adv_robust:2}
    \begin{split}
   \mathfrak{R}(\fhat^N; \muadv) -  \hat{\mathfrak{R}}(\fhat^N) \leq &
   \transcost\left(\bP_\mathrm{global}\right) + \transerror(\bP_\mathrm{global}).
\end{split}
\end{equation}
Combining \eqref{eq:cor:adv_robust:1} and \eqref{eq:cor:adv_robust:2}, we complete the proof of Corollary~\ref{cor:adv_robust}.
\end{proof}

\subsection{Proof of Theorem~\ref{thm:concentration_main}}\label{app:proof_main}
\begin{proof}\textbf{of Theorem~\ref{thm:concentration_main}.}
We first consider the non-smooth case.  That is, we consider the case where $g^{N}$ is almost surely $\alpha$-H\"{o}lder with $0<\alpha\le 1$.\\
\textbf{Step 1 }(Change of measure)\textbf{.}  By Lemma~\ref{lem:com_holder_loc}, we deduce that
\begin{equation}
\label{PROOF__thm:concentration_main_full_version}
\begin{split}
     \int_{z\in\calZ}\, g^N(z)\,\mu(dz)
\leq \int_{z\in\calZ}\, g^N(z)\,\nu(dz) &+ 
    \sum_{P\in \bP}
            \nu(P)
        \Lip_{\alpha}(g^N|P) 
        \calW_{\alpha}(\mu|_P,\nu|_P)\\
    &+
    \sum_{P\in \bP}
        \bigg(1-\frac{
            \nu(P)
        }{
            \mu(P)
        }\bigg)
        \int_{z \in P}\, g^N(z)\,\mu(dz)
.
\end{split}
\end{equation}
Setting $\nu = \mu^{N}$, we obtain an estimation of the expectation of $g^N$ under $\mu$ since
\begin{equation*}
\begin{split}
     \int_{z\in\calZ}\, g^N(z)\,\mu(dz)
\leq \int_{z\in\calZ}\, g^N(z)\,\mu^{N}(dz) &+ 
    \sum_{P\in \bP}
            \mu^{N}(P)
        \Lip_{\alpha}(g^N|P) 
        \calW_{\alpha}(\mu|_P,\mu^{N}|_P)\\
    &+
    \sum_{P\in \bP}
        \bigg(1-\frac{
            \mu^{N}(P)
        }{
            \mu(P)
        }\bigg)
        \int_{z \in P}\, g^N(z)\,\mu(dz)
.
\end{split}
\end{equation*}
We simplify notations by defining for each $P \in \bP$, the following three abbreviations:
\begin{align*}
    \calD_P &\eqdef W_{\alpha}(\mu|_P,\mu^{N}|_P),\\
    \calB^N &\eqdef \sum_{P\in \bP}
            \mu^{N}(P)
        \Lip_{\alpha}(g^N|P) 
        \calW_{\alpha}(\mu|_P,\mu^{N}|_P), \\
    \calR^N & \eqdef \sum_{P\in \bP}
        \bigg(1-\frac{
            \mu^{N}(P)
        }{
            \mu(P)
        }\bigg)
        \int_{z \in P}\, g^N(z)\,\mu(dz)
.
\end{align*}
With these notational short-hands, we concisely  rewrite~\eqref{PROOF__thm:concentration_main_full_version} as
\begin{equation}
\label{eq:thm:concentration_main_full_version:0}
\begin{split}
     \int_{z\in\calZ}\, g^N(z)\,\mu(dz)
\leq \int_{z\in\calZ}\, g^N(z)\,\mu^{N}(dz) + \calB^N + \calR^N.
\end{split}
\end{equation}
In order to control our upper-bound in~\eqref{eq:thm:concentration_main_full_version:0}, we must control the terms $\calB^N$ and $\calR^N$; which we now do. \\ 
\textbf{Step 2 }(Integral probability metric concentration)\textbf{.} 
First we control the term $\calB^N$.  Notice that 
\begin{align*}
    \calB^N &= \sum_{P\in \bP}
            \mu^{N}(P)
        \Lip_{\alpha}(g^N|P) 
        \calW_{\alpha}(\mu|_P,\mu^{N}|_P)\\
        &= \sum_{P\in \bP}
            \mu^{N}(P)
        \Lip_{\alpha}(g^N|P)\,
        \E\big[\calW_{\alpha}(\mu|_P,\mu^{N}|_P)\vert N_P\big]\\
        &~ + \sum_{P\in \bP}
            \mu^{N}(P)
        \Lip_{\alpha}(g^N|P)\,\Big(\calW_{\alpha}(\mu|_P,\mu^{N}|_P) - 
        \E\big[\calW_{\alpha}(\mu|_P,\mu^{N}|_P)\vert N_P\big]\Big)\\
        &\leq \sum_{P\in \bP}
            \mu^{N}(P)
        \Lip_{\alpha}(g^N|P)\,
        \E\big[\calW_{\alpha}(\mu|_P,\mu^{N}|_P)\vert N_P\big]\\
        &~ + L\sum_{P\in \bP}
            \mu^{N}(P)
        \Big\vert\calW_{\alpha}(\mu|_P,\mu^{N}|_P) - 
        \E\big[\calW_{\alpha}(\mu|_P,\mu^{N}|_P)\vert N_P\big]\Big\vert.
\end{align*}
Deploying Lemma~\ref{lem:con_holder_wass}, we find that
\begin{equation}\label{eq:thm:concentration_main_full_version:1}
\begin{split}
        &\sum_{P\in \bP}
            \mu^{N}(P)
        \Lip_{\alpha}(g^N|P)\,
        \E\big[\calW_{\alpha}(\mu|_P,\mu^{N}|_P)\vert N_P\big]\\
        \leq~& \sum_{P\in \bP}
            \mu^{N}(P)
        \Lip_{\alpha}(g^N|P)\,
        C_{d_{\calZ},\alpha}\,\diam(P)\, \mathrm{rate}_{d_{\calZ},\alpha}(N_P).
\end{split}
\end{equation}
and that
\begin{equation*}
    \sP \Big(\Big\vert\calW_{\alpha}(\mu|_P,\mu^{N}|_P) - 
        \E\big[\calW_{\alpha}(\mu|_P,\mu^{N}|_P)\vert N_P\big]\Big\vert \geq \epsilon\,\big\vert\, N_P\Big)  \leq 2e^{-\frac{2N_P\,\epsilon^2}{\diam(P)^{2\alpha}}}.
\end{equation*}
Synchronizing our notation with that of Lemma~\ref{lem:subgaussian2} we set $C_P=2$, $\sigma_P^2 = \diam(P)^{2\alpha}/4N_P$, $\alpha_P = L\mu^{N}(P)$ and
\begin{equation*}
    X_P = \Big\vert\calW_{\alpha}(\mu|_P,\mu^{N}|_P) - 
        \E\big[\calW_{\alpha}(\mu|_P,\mu^{N}|_P)\vert N_P\big]\Big\vert.
\end{equation*}
for $ P\in\bP$. Apply Lemma~\ref{lem:subgaussian2} while conditioning on $N_P$ we have that for every $\epsilon > 0$ and each $N\in\nn$
\begin{equation*}
    \sP\Big[\big\vert\sum_{P \in \bP}\alpha_P X_P \big\vert \geq \epsilon \,\big\vert\, N_P \Big] \leq 2 e^{-\frac{\epsilon^2}{8\tilde{\sigma}^2}},
\end{equation*}
where $\tilde{\sigma}^2$ is can be bounded above as follows
\begin{align*}
    \tilde{\sigma}^2 =& \sum_{P \in \bP}C_P^2\alpha_P^2 \sigma_P^2
    \\
    = & \sum_{P \in \bP}4\mu^{N}(P)^2 L^2\frac{\diam(P)^{2\alpha}}{4N_P}
    \\
    =&
    \sum_{P \in \bP}  \frac{N_P}{N^2}L^2\diam(P)^{2\alpha} 
    \\
    \leq  & 
    \frac{L^2}{N}\max_{P \in \bP} \diam(P)^{2\alpha}.
\end{align*}
Therefore, we deduce the following concentration inequality
\begin{align*} \label{eq:thm:concentration_main_full_version:3}
    \sP\Big(\big\vert\sum_{P \in \bP}\alpha_P X_P \big\vert \geq \epsilon\Big) & = \sE\Big[\sP\Big(\big\vert\sum_{P \in \bP}\alpha_P X_P \big\vert \geq \epsilon \,\big\vert\, N_P\Big)\Big]\\ 
    &\leq \sE\Big[2 \exp\Big\{-\frac{N\epsilon^2}{L^2\max_{P \in \bP} \diam(P)^{2\alpha}}\Big\}\Big]\\
    &= 2 \exp\Big\{-\frac{N\epsilon^2}{L^2\max_{P \in \bP} \diam(P)^{2\alpha}}\Big\}
\end{align*}
Fix our ``threshold probability'' $0<\delta_1 \le 1$.  With probability $1-\delta_1$ it holds that
\begin{equation*}
\label{eq:thm:concentration_main_full_version:4}
\resizebox{0.95\hsize}{!}{$
        L\sum_{P\in \bP}
            \mu^{N}(P)
        \Big\vert\calW_{\alpha}(\mu|_P,\mu^{N}|_P) - 
        \E\big[\calW_{\alpha}(\mu|_P,\mu^{N}|_P)\vert N_P\big]\Big\vert
        \leq L\max_{P \in \bP} \diam(P)^{\alpha}\left(\frac{\ln (2/\delta_1)}{N} \right)^{1/2}.
$}
\end{equation*}
Combining it with \eqref{eq:thm:concentration_main_full_version:1}, we conclude that the following holds with probability $1-\delta_1$
\begin{equation}
\label{eq:thm:concentration_main_full_version:5}
\begin{split}
    \calB^N \leq & \sum_{P\in \bP}
            \mu^{N}(P)
        \Lip_{\alpha}(g^N|P)\,
        C_{d_{\calZ},\alpha}\,\diam(P)\, \mathrm{rate}_{d_{\calZ},\alpha}(N_P) \\
        & +  L\max_{P \in \bP} \diam(P)^{\alpha}\left(\frac{\ln (2/\delta_1)}{N} \right)^{1/2}.
\end{split}
\end{equation}
\textbf{Step 3: }(Global concentration)\textbf{.} It remains to estimate the term $\calR^N$. Let $\delta_2 > 0$, by Lemma~\ref{lem:con_global_mismatch}, we know that the following holds with probability $1-\delta_2$ 
\begin{equation}\label{eq:thm:concentration_main_full_version:6}
    \mathcal{R}_N \leq \frac{\Vert g^N \Vert_{\infty}}{N^{1/2}}\max\{\sqrt{2\ln(2/\delta_2)}, \sqrt{k}\}.
\end{equation} 
Combining \eqref{eq:thm:concentration_main_full_version:0}, \eqref{eq:thm:concentration_main_full_version:5} and \eqref{eq:thm:concentration_main_full_version:6}, we have with probability greater than $(1-\delta_1)(1-\delta_2)$
 \begin{equation*}
    \sE\big[g^N(Z)\big] -   \frac{1}{N}\sum_{n=1}^N g^N(Z_n) \leq  \sum_{P\in \bP}
            \mu^{N}(P)
        \Lip_{\alpha}(g^N|P)\,
        C_{d_{\calZ},\alpha}\,\diam(P)\, \mathrm{rate}_{d_{\calZ},\alpha}(N_P) + \epsilon,
\end{equation*}
where the term $\epsilon$ is given by
\begin{align*}   
\epsilon \eqdef L\max_{P \in \bP} \diam(P)^{\alpha}\left(\frac{\ln (2/\delta_1)}{N} \right)^{1/2} + \frac{\Vert g^N \Vert_{\infty} }{N^{1/2}}\max\{\sqrt{2\ln(2/\delta_2)}, \sqrt{k}\}.
\end{align*}
Let $\delta \in (0,1]$.  Set $\delta_1 \eqdef \delta_2 \eqdef \delta/2$.  We now have with probability greater than $1-\delta$ that 
\begin{equation*}
    \epsilon \eqdef L\max_{P \in \bP} \diam(P)\left(\frac{\ln (4/\delta)}{N} \right)^{1/2} + \frac{\Vert g^N \Vert_{\infty} }{N^{1/2}}\max\{\sqrt{2\ln(4/\delta)}, \sqrt{k}\}.
\end{equation*}
We now turn our attention to the proof of the smooth case; which is similar modulo some a few changes at key points in its proof. \\
\textbf{Step 1 }(Change of measure)\textbf{.}  By Applying Lemma~\ref{lem:com_smooth_loc} we have 
\begin{equation*}
\label{eq:COM_smoothcase_MainTheorem}
\begin{split}
     \int_{z\in \calZ}\, g^N(z)\,\mu(dz)
\leq \int_{z\in \calZ}\, g^N(z)\,\nu(dz) &+ 
    \sum_{P\in \bP}
            \nu(P)
        \Vert g^N \Vert_{s:P} 
        \calW_{\calC^{s}}(\mu|_P,\nu|_P)\\
    &+
    \sum_{P\in \bP}
        \bigg(1-\frac{
            \nu(P)
        }{
            \mu(P)
        }\bigg)
        \int_{z \in P}\, g^N(z)\,\mu(dz)
.
\end{split}
\end{equation*}
Setting $\nu \eqdef \mu^{N}$ in the above equation, we estimate the mean of $g^N$ with respect to $\mu$
\begin{equation}
\label{eq:integralgNmu_smoothcase}
\begin{split}
     \int_{z\in \calZ}\, g^N(z)\,\mu(dz)
\leq \int_{z\in \calZ}\, g^N(z)\,\mu^{N}(dz) &+ 
    \sum_{P\in \bP}
            \mu^{N}(P)
        \Vert g^N \Vert_{s:P} 
        \calW_{\calC^{s}}(\mu|_P,\mu^{N}|_P)\\
    &+
    \sum_{P\in \bP}
        \bigg(1-\frac{
            \mu^{N}(P)
        }{
            \mu(P)
        }\bigg)
        \int_{z \in P}\, g^N(z)\,\mu(dz)
.
\end{split}
\end{equation}
As before, we simplify our notation.  For each for all $P \in \bP$ we abbreviate  
\begin{align*}
    \calD_P &\eqdef \calW_{\calC^{s}}(\mu|_P,\mu^{N}|_P)\\
    \calB^N &\eqdef \sum_{P\in \bP}
            \mu^{N}(P)
        \Vert g^N \Vert_{s:P} 
        \calW_{\calC^{s}}(\mu|_P,\mu^{N}|_P)\\
    \calR^N &\eqdef \sum_{P\in \bP}
        \bigg(1-\frac{
            \mu^{N}(P)
        }{
            \mu(P)
        }\bigg)
        \int_{z \in P}\, g^N(z)\,\mu(dz)
.
\end{align*}
Therefore,~\eqref{eq:integralgNmu_smoothcase} can be succinctly written as
\begin{equation}
\label{eq:thm:concentration_main_full_version:0:smooth}
\begin{split}
     \int_{z\in \calZ}\, g^N(z)\,\mu(dz)
\leq \int_{z\in \calZ}\, g^N(z)\,\mu^{N}(dz) + \calB^N + \calR^N.
\end{split}
\end{equation}
As in the non-smooth case, we need only bound the terms $\calB^N$ and $\calR^N$ in order to control the left-hand side of~\eqref{eq:thm:concentration_main_full_version:0:smooth}.  \\
\textbf{Step 2 }(Integral probability metric concentration)\textbf{.} 
We again first control the term $\calB^N$.  Observe that
\begin{align*}
    \calB^N &= \sum_{P\in \bP}
            \mu^{N}(P)
        \Vert g^N \Vert_{s:P} 
        \calW_{\calC^{s}}(\mu|_P,\mu^{N}|_P)\\
        &= \sum_{P\in \bP}
            \mu^{N}(P)
        \Vert g^N \Vert_{s:P}\,
        \E\big[\calW_{\calC^{s}}(\mu|_P,\mu^{N}|_P)\vert N_P\big]\\
        &~ + \sum_{P\in \bP}
            \mu^{N}(P)
        \Vert g^N \Vert_{s:P}\,\Big(\calW_{\calC^{s}}(\mu|_P,\mu^{N}|_P) - 
        \E\big[\calW_{\calC^{s}}(\mu|_P,\mu^{N}|_P)\vert N_P\big]\Big)\\
        &\leq \sum_{P\in \bP}
            \mu^{N}(P)
        \Vert g^N \Vert_{s:P}\,
        \E\big[\calW_{\calC^{s}}(\mu|_P,\mu^{N}|_P)\vert N_P\big]\\
        &~ + L\sum_{P\in \bP}
            \mu^{N}(P)
        \Big\vert\calW_{\calC^{s}}(\mu|_P,\mu^{N}|_P) - 
        \E\big[\calW_{\calC^{s}}(\mu|_P,\mu^{N}|_P)\vert N_P\big]\Big\vert
\end{align*}
Applying Lemma~\ref{lem:con_smooth_wass} we have both that
\begin{equation}\label{eq:thm:concentration_main_full_version:1:smooth}
\begin{split}
        &\sum_{P\in \bP}
            \mu^{N}(P)
        \Vert g^N \Vert_{s:P}\,
        \E\big[\calW_{\calC^{s}}(\mu|_P,\mu^{N}|_P)\vert N_P\big]\\
        \leq~& \sum_{P\in \bP}
            \mu^{N}(P)
        \Vert g^N \Vert_{s:P}\,
        C_{d_{\calZ},s}\,\diam(P)\, \mathrm{rate}_{d_{\calZ},s}(N_P).
\end{split}
\end{equation}
and that
\begin{equation*}
    \sP \Big(\Big\vert\calW_{\calC^{s}}(\mu|_P,\mu^{N}|_P) - 
        \E\big[\calW_{\calC^{s}}(\mu|_P,\mu^{N}|_P)\vert N_P\big]\Big\vert \geq \epsilon\,\big\vert\, N_P\Big)  \leq 2e^{-\frac{2N_P\,\epsilon^2}{\diam(P)^{2}}}.
\end{equation*}
Synchronizing notation with Lemma~\ref{lem:subgaussian2} we denote $C_P=1$, $\sigma_P^2 = \diam(P)^{2}/4N_P$ and $\alpha_P = L\mu^{N}(P)$,
\begin{equation*}
    X_P = \Big\vert\calW_{\calC^{s}}(\mu|_P,\mu^{N}|_P) - 
        \E\big[\calW_{\calC^{s}}(\mu|_P,\mu^{N}|_P)\vert N_P\big]\Big\vert
\end{equation*}
for $ P\in\bP$. Applying Lemma~\ref{lem:subgaussian2} while conditioning on $N_P$, we have for all $\epsilon > 0$ and $N\in\nn$,
\begin{equation*}
    \sP\Big[\big\vert\sum_{P \in \bP}\alpha_P X_P \big\vert \geq \epsilon \,\big\vert\, N_P \Big] \leq 2 e^{-\frac{\epsilon^2}{8\tilde{\sigma}^2}},
\end{equation*}
where, similarly to the smooth case, $\tilde{\sigma}^2$ is bounded above by
\begin{align*}
    \tilde{\sigma}^2 =& \sum_{P \in \bP}C_P^2 \alpha_P^2 \sigma_P^2
    \\
    = & \sum_{P \in \bP}4\mu^{N}(P)^2 L^2\frac{\diam(P)^{2}}{4N_P}
    \\
    =&
    \sum_{P \in \bP}  \frac{N_P}{N^2}L^2\diam(P)^{2} 
    \\
    \leq  & 
    \frac{L^2}{N}\max_{P \in \bP} \diam(P)^{2}.
\end{align*}
Therefore, we arive at the fact that
\begin{align*} \label{eq:thm:concentration_main_full_version:3:smooth}
    \sP\Big(\big\vert\sum_{P \in \bP}\alpha_P X_P \big\vert \geq \epsilon\Big) & = \sE\Big[\sP\Big(\big\vert\sum_{P \in \bP}\alpha_P X_P \big\vert \geq \epsilon \,\big\vert\, N_P\Big)\Big]\\ 
    &\leq \sE\Big[2 \exp\Big\{-\frac{N\epsilon^2}{L^2\max_{P \in \bP} \diam(P)^{2}}\Big\}\Big]\\
    &= 2 \exp\Big\{-\frac{N\epsilon^2}{L^2\max_{P \in \bP} \diam(P)^{2}}\Big\}.
\end{align*}
Let $0<\delta_1 \le 1$ be our ``threshold probability''.  Thus, with probability $1-\delta_1$ it holds that
\begin{equation*}
\label{eq:thm:concentration_main_full_version:4:smooth}
\begin{split}
        L\sum_{P\in \bP}
            \mu^{N}(P)
        \Big\vert\calW_{\calC^s}(\mu|_P,\mu^{N}|_P) - 
        \E\big[\calW_{\calC^s}(\mu|_P,\mu^{N}|_P)\vert N_P\big]\Big\vert
        \leq L\max_{P \in \bP} \diam(P)\left(\frac{\ln (2/\delta_1)}{N} \right)^{1/2}.
\end{split}
\end{equation*}
Combine this with \eqref{eq:thm:concentration_main_full_version:1:smooth} we conclude that with probability $1-\delta_1$
\begin{equation}
\label{eq:thm:concentration_main_full_version:5:smooth}
\calB^N \leq \sum_{P\in \bP}
            \mu^{N}(P)
        \Vert g^N \Vert_{s:P}\,
        C_{d_{\calZ},s}\,\diam(P)\, \mathrm{rate}_{d_{\calZ},s}(N_P) +  L\max_{P \in \bP} \diam(P)\left(\frac{\ln (2/\delta_1)}{N} \right)^{1/2}.
\end{equation}
\textbf{Step 3: }(Global concentration)\textbf{.} As with the smooth case, it only now remains to control the term $\calR^N$.  Set $0<\delta_2 \le 1$.  Then, by Lemma~\ref{lem:con_global_mismatch}, we have that the following holds with probability at-least $1-\delta_2$
\begin{equation}\label{eq:thm:concentration_main_full_version:6:smooth}
    \mathcal{R}_N \leq \frac{\Vert g^N \Vert_{\infty}}{N^{1/2}}\max\{\sqrt{2\ln(2/\delta_2)}, \sqrt{k}\}.
\end{equation} 
Combining \eqref{eq:thm:concentration_main_full_version:0:smooth}, \eqref{eq:thm:concentration_main_full_version:5:smooth} and \eqref{eq:thm:concentration_main_full_version:6:smooth}, we have with probability greater than $(1-\delta_1)(1-\delta_2)$
 \begin{equation*}
    \sE\big[g^N(Z)\big] -   \frac{1}{N}\sum_{n=1}^N g^N(Z_n) \leq  \sum_{P\in \bP}
            \mu^{N}(P)
        \Vert g^N \Vert_{s:P}\,
        C_{d_{\calZ},s}\,\diam(P)\, \mathrm{rate}_{d_{\calZ},s}(N_P) + \epsilon,
\end{equation*}
where the ``error term'' $\epsilon$ is given by
\begin{align*}   
\epsilon \eqdef L\max_{P \in \bP} \diam(P)\left(\frac{\ln (2/\delta_1)}{N} \right)^{1/2} + \frac{\Vert g^N \Vert_{\infty} }{N^{1/2}}\max\{\sqrt{2\ln(2/\delta_2)}, \sqrt{k}\}.
\end{align*}
Fix $\delta \in (0,1]$.  Set $\delta_1 \eqdef \delta_2 \eqdef \delta/2$.  Thus, from our analysis we arrive at the conclusion that
\begin{equation*}
    \epsilon = L\max_{P \in \bP} \diam(P)\left(\frac{\ln (4/\delta)}{N} \right)^{1/2} + \frac{\Vert g^N \Vert_{\infty} }{N^{1/2}}\max\{\sqrt{2\ln(4/\delta)}, \sqrt{k}\},
\end{equation*}
holds with probability at-least $1-\delta$.  This concludes our proof.  
\end{proof}

\section{Helper Lemmas}
\label{app:sec:helper_lemmas}
\subsection{Change of Measure Helper Inequalities}\label{app:subsec:com}
\begin{lemma}[Change of Measure of H\"{o}lder Function]
   \label{lem:com_holder}
Let $\calZ$ be a complete and separable metric space and $\mu, \nu \in \calP_1(\calZ)$. Let $g:\calZ \rightarrow \sR$ be an $\alpha$-H\"{o}lder function with $\alpha \in (0,1]$. Then
\begin{equation*}
    \int_{z \in \calZ} g(z)\,\mu(dz)
    \leq 
    \Lip_{\alpha}(g) 
    \mathcal{W}_{\alpha}(\mu,\nu)
        + 
    \int_{z \in \calZ} g(z)\,\nu(dz).
\end{equation*}
\end{lemma}
\begin{proof}\textbf{of Lemma~\ref{lem:com_holder}.}
\edit{Since $\calZ$ is complete and separable then Kantorovich Duality}{}
\citep[Theorem 5.10]{villani2009optimal} \edit{with the transport-cost function $c(x_1,x_2) =  \norm{x_1-x_2}$ implies that}{By the definition of $\mathcal{W}_{\alpha}(\mu,\nu)$ as an integral probability distance,}

\begin{equation*}
    \mathcal{W}_{\alpha}(\mu,\nu)
        =  
    \sup_{f\in \calC(\calZ),\,  \Lip_{\alpha}(f)\leq 1}\, \int f(z)\,\mu(dz) - \int f(z)\,\nu(dz)
    .
\end{equation*}
If $g$ is constant, then Lemma~\ref{lem:com_holder} holds trivially. Otherwise, $0 < \Lip_{\alpha}(g) < \infty$ and we let $\tilde{g} \eqdef \Lip_{\alpha}(g)^{-1}g$. Then $\Lip_{\alpha}(\tilde{g})\leq 1$ and we have
\begin{equation*}
    \begin{aligned}
    \operatorname{Lip}_{\alpha}(g)^{-1}\int g\,\mu(dz)
    = &
    \int \tilde{g}(z)\,\mu(dz) \\
    \leq & \mathcal{W}_{\alpha}(\mu,\nu)
    + \int \tilde{g}(z)\,\nu(dz)\\
    = & \mathcal{W}_{\alpha}(\mu,\nu)
    + \Lip_{\alpha}(g)^{-1}\int g(z)\,\nu(dz).
    \end{aligned}
\label{eq:lem:com_holder}
\end{equation*}
Multiplying across by $\Lip_{\alpha}(g)>0$ yields the desired result.
\end{proof}
\begin{lemma}[Local Change of Measure of H\"{o}lder Function]
\label{lem:com_holder_loc}
Let $\calZ$ a subset of $\R^{d_{\calZ}}$ and $\mu, \nu \in \calP_1(\calZ)$. Let $g:\calZ \rightarrow \sR$ be locally $\alpha$-H\"{o}lder with $\alpha \in (0,1]$. Then
\begin{equation*}
\begin{split}
     \int_{z\in \calZ}\, g(z)\,\mu(dz)
\leq \int_{z\in \calZ}\, g(z)\,\nu(dz) &+ 
    \sum_{P\in \bP}
            \nu(P)
        \Lip_{\alpha}(g|P) 
        \calW_{\alpha}(\mu|_P,\nu|_P)\\
    &+
    \sum_{P\in \bP}
        \bigg(1-\frac{
            \nu(P)
        }{
            \mu(P)
        }\bigg)
        \int_{z \in P}\, g(z)\,\mu(dz)
.
\end{split}
\end{equation*}
\end{lemma}
\begin{proof}\textbf{of Lemma~\ref{lem:com_holder_loc}.}
Let for all $P \in \bP$ that
\begin{equation*}
    \mu_P(\cdot) = \mu(\cdot \cap P),\quad \nu_P(\cdot) = \nu(\cdot \cap P),\quad 
    \tilde{\mu}_{P} = \frac{\nu(P)}{\mu(P)}\mu_P.
\end{equation*}
Then we apply Lemma~\ref{lem:com_holder} to $\tilde{\mu}_{P}$ and $\nu_{P}$ for all $P \in \bP$ and have
\begin{equation*}
    \begin{aligned}
    \int_{z\in\calZ} -g(z)\nu(dz) 
    &= \sum_{P\in \bP}\int_{z\in\calZ} -g(z)\nu_{P}(dz)\\
    &\leq \sum_{P\in \bP} \Big(
    \mathrm{Lip}_{\alpha}(g\vert P) \mathcal{W}_{\alpha}\big(\nu_P, \tilde{\mu}_P\big) +  \int_{z\in\calZ} -g(z)\tilde{\mu}_{P}(dz)\Big).
\end{aligned}
\end{equation*}
By adding $\int_{z\in\calZ} g(z)\mu(dz)$ on both sides and rearranging terms we have that 
\begin{equation*}
    \begin{aligned}
    \int_{z\in\calZ} g(z)\mu(dz) &\leq \int_{z\in\calZ} g(z)\nu(dz) + \sum_{P\in \bP}\mathrm{Lip}_{\alpha}(g\vert P) \mathcal{W}_{\alpha}\big(\nu_P, \tilde{\mu}_P\big)\\
    &\qquad + \int_{z\in\calZ} g(z)\mu_P(dz) - \int_{z\in\calZ} g(z)\frac{\nu(P)}{\mu(P)}\mu_P(dz)\\
    &\leq \int_{z\in\calZ} g(z)\nu(dz) + \sum_{P\in \bP}\mathrm{Lip}_{\alpha}(g\vert P) \nu(P) \mathcal{W}_{\alpha}\Big(\frac{\nu_P}{\nu(P)},\frac{\mu_P}{\mu(P)}\Big)\\
    &\qquad + \int_{z\in\calZ} g(z)\mu_P(dz) - \int_{z\in\calZ} g(z)\frac{\nu(P)}{\mu(P)}\mu_P(dz)\\
    &\leq \int_{z\in\calZ} g(z)\nu(dz) + \sum_{P\in \bP}\mathrm{Lip}_{\alpha}(g\vert P) \nu(P) \mathcal{W}_{\alpha}\Big(\nu\vert_P,\mu\vert_P\Big)\\
    &\qquad + \Big(1 - \frac{\nu(P)}{\mu(P)}\Big)\int_{z\in\calZ} g(z)\mu_P(dz)\\
    &\leq \int_{z\in\calZ} g(z)\mu^{N}(dz) + \sum_{P\in \bP}\mathrm{Lip}_{\alpha}(g\vert P)\mu^{N}(P) \mathcal{W}_{\alpha}\Big((\mu\vert_P)^{N_P},\mu\vert_P\Big)\\
    &\qquad + \Big(1 - \frac{\mu^{N}(P)}{\mu(P)}\Big)\int_{z\in\calZ} g(z)\mu_P(dz).
\end{aligned}
\end{equation*}
\end{proof}

\begin{lemma}[Change of Measure of Smooth Function]
   \label{lem:com_smooth}
Let $\calZ = \R^{d_{\calZ}}$, $d_{\calZ}\in\sN$ and $\mu, \nu \in \calP_1(\calZ)$. Let $g\in \calC^{s}(\calZ)$ with $s \in \sN$. Then
\begin{equation*}
    \int_{z \in \calZ} g(z)\,\mu(dz)
    \leq 
    \Vert g \Vert_{s}  \calW_{\calC^{s}}(\mu, \nu)
        + 
    \int_{z \in \calZ} g(z)\,\nu(dz).
\end{equation*}
\end{lemma}
\begin{proof}\textbf{of Lemma~\ref{lem:com_smooth}.}
The proof is similar with the proof of Lemma~\ref{lem:com_holder}. Recall the definition of $\calW_{\calC^{s}}$ that
\begin{equation*}
    \calW_{\calC^{s}}(\mu,\nu)
        \eqdef 
    \sup_{f\in \calC(\calZ),\,  \Vert f \Vert_{s}\leq 1}\, \int f(z)\,\mu(dz) - \int f(z)\,\nu(dz)
    .
\end{equation*}
If $g$ is constant, then Lemma~\ref{lem:com_smooth} holds trivially. Otherwise, $0 < \Vert g \Vert_{s} < \infty$ and we Let $\tilde{g} \eqdef \Vert g \Vert_{s}^{-1}g$. Then $\Vert g \Vert_{s}\leq 1$ and we have
\begin{equation*}
    \begin{aligned}
    \Vert g \Vert_{s}^{-1}\int g\,\mu(dz)
    = &
    \int \tilde{g}(z)\,\mu(dz) \\
    \leq & \mathcal{W}_{\alpha}(\mu,\nu)
    + \int \tilde{g}(z)\,\nu(dz)\\
    = & \mathcal{W}_{\alpha}(\mu,\nu)
    + \Vert g \Vert_{s}^{-1}\int g(z)\,\nu(dz).
    \end{aligned}
\label{eq:lem:com_smooth}
\end{equation*}
Multiplying across by $\Vert g \Vert_{s}>0$ yields the desired result.
\end{proof}

\begin{lemma}[Local Change of Measure of Smooth Function]
\label{lem:com_smooth_loc}
Let $\calZ = \R^{d_{\calZ}}$, $d_{\calZ}\in\sN$ and $\mu, \nu \in \calP_1(\calZ)$. Let $g\in \calC^{s}(\calZ)$ with $s \in \sN$. Then
\begin{equation*}
\begin{split}
     \int_{z\in \calZ}\, g(z)\,\mu(dz)
\leq \int_{z\in \calZ}\, g(z)\,\nu(dz) &+ 
    \sum_{P\in \bP}
            \mu(P)
        \Vert g \Vert_{s:P}
        \calW_{\calC^{s}}(\mu|_P,\nu|_P)\\
    &+
    \sum_{P\in \bP}
        \bigg(1-\frac{
            \nu(P)
        }{
            \mu(P)
        }\bigg)
        \int_{z \in P}\, g(z)\,\mu(dz)
.
\end{split}
\end{equation*}
\end{lemma}
\begin{proof}\textbf{of Lemma~\ref{lem:com_smooth_loc}.}
The proof is similar with the proof of Lemma~\ref{lem:com_holder_loc}. Let for all $P \in \bP$ that
\begin{equation*}
    \mu_P(\cdot) = \mu(\cdot \cap P),\quad \nu_P(\cdot) = \nu(\cdot \cap P),\quad 
    \tilde{\mu}_{P} = \frac{\nu(P)}{\mu(P)}\mu_P.
\end{equation*}
Then we apply Lemma~\ref{lem:com_holder} to $\tilde{\mu}_{P}$ and $\nu_{P}$ for all $P \in \bP$ and have
\begin{equation*}
    \begin{aligned}
    \int_{z\in\calZ} -g(z)\nu(dz) 
    &= \sum_{P\in \bP}\int_{z\in\calZ} -g(z)\nu_{P}(dz)\\
    &\leq \sum_{P\in \bP} \Big(
    \Vert g \Vert_{s:P} \calW_{\calC^{s}}\big(\nu_P, \tilde{\mu}_P\big) +  \int_{z\in\calZ} -g(z)\tilde{\mu}_{P}(dz)\Big).
\end{aligned}
\end{equation*}
By adding $\int_{z\in\calZ} g(z)\mu(dz)$ on both sides and rearranging terms we have that 
\begin{equation*}
    \begin{aligned}
    \int_{z\in\calZ} g(z)\mu(dz) &\leq \int_{z\in\calZ} g(z)\nu(dz) + \sum_{P\in \bP}\Vert g \Vert_{s:P} \calW_{\calC^{s}}\big(\nu_P, \tilde{\mu}_P\big)\\
    &\qquad + \int_{z\in\calZ} g(z)\mu_P(dz) - \int_{z\in\calZ} g(z)\frac{\nu(P)}{\mu(P)}\mu_P(dz)\\
    &\leq \int_{z\in\calZ} g(z)\nu(dz) + \sum_{P\in \bP}\nu(P) \Vert g \Vert_{s:P}   \calW_{\calC^{s}}\Big(\frac{\nu_P}{\nu(P)},\frac{\mu_P}{\mu(P)}\Big)\\
    &\qquad + \int_{z\in\calZ} g(z)\mu_P(dz) - \int_{z\in\calZ} g(z)\frac{\nu(P)}{\mu(P)}\mu_P(dz)\\
    &\leq \int_{z\in\calZ} g(z)\nu(dz) + \sum_{P\in \bP}\nu(P)\Vert g \Vert_{s:P}   \calW_{\calC^{s}} \Big(\nu\vert_P,\mu\vert_P\Big)\\
    &\qquad + \Big(1 - \frac{\nu(P)}{\mu(P)}\Big)\int_{z\in\calZ} g(z)\mu_P(dz)\\
    &\leq \int_{z\in\calZ} g(z)\mu^{N}(dz) + \sum_{P\in \bP}\mu^{N}(P)\Vert g \Vert_{s:P}   \calW_{\calC^{s}}\Big((\mu\vert_P)^{N_P},\mu\vert_P\Big)\\
    &\qquad + \Big(1 - \frac{\mu^{N}(P)}{\mu(P)}\Big)\int_{z\in\calZ} g(z)\mu_P(dz).
\end{aligned}
\end{equation*}
\end{proof}

\subsection{Helper Wasserstein Concentration Inequalities}\label{app:subsec:concentration}

\begin{lemma}[Concentration of H\"{o}lder Wasserstein Metric]
   \label{lem:con_holder_wass}
Let $\calZ$ a compact subset of $\R^{d_{\calZ}}$ and $\mu \in \calP_1(\calZ)$.  Then for all $\epsilon > 0$, $N\in\nn$ 
\begin{equation*}
    \sP \Bigg(\bigg\vert \mathcal{W}_{\alpha}(\mu,\mu^N) - \sE\Big[\mathcal{W}_{\alpha}(\mu,\mu^N)\Big] \bigg\vert \geq \epsilon\Bigg)  \leq 2e^{-\frac{2N\epsilon^2}{\diam(\calZ)^{2\alpha}}}
\end{equation*}
where $C_{d_{\calZ},\alpha}$ is given in \cref{tab:concentration_main_table_version__FULL} and 
\begin{equation*}
    \sE\Big[\mathcal{W}_{\alpha}(\mu,\mu^N)\Big] \leq  C_{d_{\calZ},\alpha}\,\diam(\calZ)\, \mathrm{rate}_{d_{\calZ},\alpha}(N)
\end{equation*}
with $ \mathrm{rate}_{d_{\calZ},\alpha}(N)$ also given in \cref{tab:concentration_main_table_version__FULL}.
\end{lemma}
\begin{proof}\textbf{of Lemma~\ref{lem:con_holder_wass}.}
In the proof of Lemma~\ref{lem:con_holder_wass}, we consider two different norms on the cube $[0,1]^{d_{\calZ}}$ in order to apply \citep[Theorem 2.1]{MR4153634}. The first is the \textit{Euclidean norm} $\|u\|_2^2:=\sum_{i=1}^{d_{\calZ}} \,u_i^2$ and the second is the \textit{$\infty$-norm} defined by $\|u\|_{\infty}:=\max_{i=1,\dots,d_{\calZ}}\, |u_i|$.  When needed from the context, we emphasize implicitly used when defining the Wasserstein distance by $\mathcal{W}_{\alpha:2}$ and $\mathcal{W}_{\alpha:\infty}$ for the Euclidean and $\infty$ norms, respectively. 
By \citep[Theorem 2.1]{MR4153634}, we have for $\calZ = [0,1]^{d_{\calZ}}$ and $N \in \sN$ 
\begin{equation*}
    \sE \Big[\mathcal{W}_{\alpha:\infty}(\mu,\mu^N)\Big] \leq d_{\calZ}^{-\alpha/2}C_{d_{\calZ},\alpha}\, \mathrm{rate}_{d_{\calZ},\alpha}(N).
\end{equation*}
By a simple fact that $\mathcal{W}_{\alpha:2} \leq d_{\calZ}^{\alpha/2}\mathcal{W}_{\alpha:\infty}$, we have 
\begin{equation*}
    \sE \Big[\mathcal{W}_{\alpha}(\mu,\mu^N)\Big] = \sE \Big[\mathcal{W}_{\alpha:2}(\mu,\mu^N)\Big] \leq C_{d_{\calZ},\alpha}\, \mathrm{rate}_{d_{\calZ},\alpha}(N).
\end{equation*}
We scale $[0,1]^{d_{\calZ}}$ with $\diam(\calZ)$ to conclude that for general $\calZ \subset \R^{d_{\calZ}}$
\begin{equation*}
    \sE \Big[\mathcal{W}_{\alpha}(\mu,\mu^N)\Big] \leq C_{d_{\calZ},\alpha}\,\diam(\calZ)\, \mathrm{rate}_{d_{\calZ},\alpha}(N).
\end{equation*}
Now we define $f\colon \calZ^{N} \to \R$ s.t.
\begin{equation*}
    f_{N}(z_1, \dots, z_N) \eqdef \calW_{\alpha}\Big(\frac{1}{N}\sum_{n=1}^{N}\delta_{z_n},\mu\Big).
\end{equation*}
For every $i = 1,\dots, N$ and every $(z_1, \dots, z_N)$, $(z_1^{\prime},\dots, z_N^{\prime}) \in \calZ^{N}$ that differs only in the $i$-th coordinate, we have
\begin{equation*}
    \vert f(z_1, \dots, z_N) - f(z_1^{\prime},\dots, z_N^{\prime}) \vert \leq \calW_{\alpha}\Big(\frac{1}{N}\sum_{n=1}^{N}\delta_{z_n},\frac{1}{N}\sum_{n=1}^{N}\delta_{z_n^{\prime}}\Big)\leq \frac{\diam(\calZ)^{\alpha}}{N}.
\end{equation*}
Therefore, with $c = \frac{\diam(\calZ)^{\alpha}}{N}$, $f$ has $(c,\dots,c)$-bounded differences property i.e. Lipschitz w.r.t. Hamming distance. Applying Lemma~\ref{lem:mcdiarmid} (the McDiarmid's inequality) with $f$ proves that for all $\epsilon > 0$ 
\begin{equation*}
        \sP \Bigg(\bigg\vert \mathcal{W}_{\alpha}(\mu,\mu^N) - \sE\Big[\mathcal{W}_{\alpha}(\mu,\mu^N)\Big] \bigg\vert \geq \epsilon\Bigg)  \leq 2e^{-\frac{2N\epsilon^2}{\diam(\calZ)^{2\alpha}}}.
\end{equation*} 
\end{proof}

\begin{lemma}[Concentration of Smooth Wasserstein Metric]
   \label{lem:con_smooth_wass}
Let $\calZ = \R^{d_{\calZ}}$, $d_{\calZ}\in\sN$ and $\mu, \nu \in \calP_1(\calZ)$. Let $g\in \calC^{s}(\calZ)$ with $s \in \sN$. Then there exist constant $C_{d_{\calZ},s}> 0$ s.t. for all $\epsilon > 0$, $N\in\nn$ 
    \begin{equation*}
        \sP \Bigg(\bigg\vert \calW_{\calC^{s}}(\mu,\mu^N) - \sE\Big[\calW_{\calC^{s}}(\mu,\mu^N)\Big] \bigg\vert \geq \epsilon\Bigg)   \leq 2e^{-\frac{2N\epsilon^2}{\diam(\calZ)^{2}}},
    \end{equation*}
and 
\begin{equation*}
    \sE \Big[\calW_{\calC^{s}}(\mu,\mu^N)\Big] \leq C_{d_{\calZ},s}\diam(\calZ) \mathrm{rate}_{d_{\calZ},s}(N).
\end{equation*}
\end{lemma}
\begin{proof}\textbf{of Lemma~\ref{lem:con_smooth_wass}.} The proof is similar to the proof of Lemma~\ref{lem:con_holder_wass}. By \citep[Theorem 1.4]{MR4153634}, we have for $\calZ = [0,1]^{d_{\calZ}}$ and $N \in \sN$
\begin{equation*}
    \sE \Big[\calW_{\calC^{s}}(\mu,\mu^N)\Big] \leq C_{d_{\calZ},s}\, \mathrm{rate}_{d_{\calZ},s}(N).
\end{equation*}
Next, we scale $[0,1]^{d_{\calZ}}$ with $\diam(\calZ)$ to conclude that for general $\calZ \subset \R^{d_{\calZ}}$
\begin{equation*}
    \sE \Big[\calW_{\calC^{s}}(\mu,\mu^N)\Big] \leq C_{d_{\calZ},s}\,\diam(\calZ)\, \mathrm{rate}_{d_{\calZ},s}(N).
\end{equation*}
Now we define $f\colon \calZ^{N} \to \R$ s.t.
\begin{equation*}
    f_{N}(z_1, \dots, z_N) \eqdef \calW_{\calC^{s}}\Big(\frac{1}{N}\sum_{n=1}^{N}\delta_{z_n},\mu\Big).
\end{equation*}
For every $i = 1,\dots, N$ and every $(z_1, \dots, z_N)$, $(z_1^{\prime},\dots, z_N^{\prime}) \in \calZ^{N}$ that differs only in the $i$-th coordinate, we have
\begin{equation*}
    \vert f(z_1, \dots, z_N) - f(z_1^{\prime},\dots, z_N^{\prime}) \vert \leq \calW_{\calC^{s}}\Big(\frac{1}{N}\sum_{n=1}^{N}\delta_{z_n},\frac{1}{N}\sum_{n=1}^{N}\delta_{z_n^{\prime}}\Big)\leq \frac{\diam(\calZ)}{N}.
\end{equation*}
Therefore, with $c = \frac{\diam(\calZ)}{N}$, $f$ has $(c,\dots,c)$-bounded differences property i.e. Lipschitz w.r.t. Hamming distance. Applying Lemma~\ref{lem:mcdiarmid} (the McDiarmid's inequality) with $f$ proves that for all $\epsilon > 0$ 
\begin{equation*}
        \sP \Bigg(\bigg\vert \calW_{\calC^{s}}(\mu,\mu^N) - \sE\Big[\calW_{\calC^{s}}(\mu,\mu^N)\Big] \bigg\vert \geq \epsilon\Bigg)  \leq 2e^{-\frac{2N\epsilon^2}{\diam(\calZ)^{2}}}.
    \end{equation*}
\end{proof}

\begin{lemma}[Concentration of Wasserstein Metric on a Manifold]
\label{lem:con_holder_wass_manifold}
Let $\calZ$ be a $d_{\calZ}$-dimensional compact class $C^1$ Riemannian manifold. Let $\mu$ be a Borel probability measure on $\calZ$, and let $\mu^N$ denote the corresponding empirical distribution based on a sample of size $N$. 
Then exist for every $\epsilon > 0 $ and $N \in \sN$,
\begin{equation}
    \label{eq:lem:con_holder_wass_manifold:1}
\sP\left(
    \bigg\vert\WD\left(\mu^N,\mu\right) - \mathbb{E}\left[\WD\left(\mu^N,\mu\right)\right] \bigg\vert \geq
   \epsilon
\right)
    \leq 
2e^{\frac{-2N\epsilon^2}{\diam(\calZ)^2}},
\end{equation}
and there exists constant $C_{\calZ} > 0$ such that
\begin{equation}
    \label{eq:lem:con_holder_wass_manifold:2}
\mathbb{E}\left[
    \WD\left(
        \mu^N
            ,
        \mu
    \right)
\right]
        \leq 
        C_{\calZ}\cdot\diam(\calZ)N^{-1/d_{\calZ}}.
\end{equation}
\end{lemma}
\begin{proof}\textbf{of Lemma~\ref{lem:con_holder_wass_manifold}.}
We recall that a $d_{\calZ}$-dimensional class $C^1$-Riemannian manifold is $d_{\calZ}$-dimensional topological manifold which is locally $C^1$-diffeomorphic to an open subset of $\mathbb{R}^{d_{\calZ}}$.
We first show that $\calZ$ has Assouad dimension $d_{\calZ}$ see \citep[Definitions 9.1 and 9.5]{Robinson_2011_DimensionEmbeddingsAttractors}.  Then, we deduce the desired concentration inequality for metric spaces of Assouad dimension $d_{\calZ}$.  The for compact Riemannian $\calZ$ then follows.

\paragraph{Step 1 - Computing $\calZ$'s Metric (Assouad) Dimension}
Since $\calZ$ is a $d_{\calZ}$-dimensional manifold then, there exists smooth charts $\{(U_k,\phi_k)\}_{k=1}^{K}$ where $\calZ=\cup_{k\le K}\, U_k$, $K\in \nn \cup\{\infty\}$, and for $k=1,\dots,K$, $\phi_k:U_k\rightarrow B_{\mathbb{R}^{d_{\calZ}}}(0,1)$ is a (class $C^{1}$) diffeomorphism and each $U_k$ is an open and bounded subset of $\calZ$ and such that 
\begin{equation*}
    \calZ = \bigcup_{k=1}^K\, \phi^{-1}_k\Big[B_{\mathbb{R}^{d_{\calZ}}}(0,1/2)\Big].
\end{equation*}  
Since $\calZ$ is compact and $\{U_k\}_{k\le K}$ is an open cover thereof then, we may without loss of generality assume that $K$ is finite.%

Applying \citep[Lemma 9.6 (iii)]{Robinson_2011_DimensionEmbeddingsAttractors} we deduce that both $B_{\mathbb{R}^{d_{\calZ}}}(0,1/2)$ and $B_{\mathbb{R}^{d_{\calZ}}}(0,1)$ have Assouad dimension $d_{\calZ}$.  By \citep[Lemma 9.6 (i)]{Robinson_2011_DimensionEmbeddingsAttractors} we deduce that the closed Euclidean ball $\overline{B_{\mathbb{R}^{d_{\calZ}}}(0,1/2)}=\{u\in \mathbb{R}^{d_{\calZ}}:\,\|u\|\le 1/2\}$ must have Assouad dimension $d_{\calZ}$.  
Since each $\phi_k$ is a diffeomorphism onto its image then $\phi^{-1}_k:B_{\mathbb{R}^{d_{\calZ}}}(0,1)\rightarrow U_k$  and $\phi_k$ are both locally Lipschitz.  Thus, each $\phi_k$ is bi-Lipschitz when restricted to the compact set $\overline{B_{\mathbb{R}^{d_{\calZ}}}(0,1/2)}$.  Consequentially, \citep[Lemma 9.6 (v)]{Robinson_2011_DimensionEmbeddingsAttractors} implies that each $\phi^{-1}_k[\overline{B_{\mathbb{R}^{d_{\calZ}}}(0,1/2)}]$ has Assouad dimension $d_{\calZ}$.  Since $K$ is finite and $U_1,\dots,U_K$ all have Assouad dimension $d_{\calZ}$ and since $\{\phi^{-1}_k[B_{\mathbb{R}^{d_{\calZ}}}(0,1/2)]\}_{k=1}^K$ is a cover of $\calZ$ (since $\{\phi^{-1}_k[\overline{B_{\mathbb{R}^{d_{\calZ}}}(0,1/2)}]\}_{k=1}^K$ is a cover of $\calZ$) then \citep[Lemma 9.6 (ii)]{Robinson_2011_DimensionEmbeddingsAttractors} implies that $\calZ$ has Assouad dimension $d_{\calZ}$.  

\paragraph{Step 2 - The Concentration Inequality}
The assumption that $\calZ$ has finite Assouad dimension $d_{\calZ}$ is equivalent to the existence of a constant $K_{\calZ}$ satisfying: for every $r>0$
\begin{equation}
\mathcal{N}^{cov}_{\calZ}(r)
        \leq 
K_{\calZ}\Big(
     \frac{\diam(\calZ)}{r}
    \Big)^{d_{\calZ}}
\label{eq_ass_bounded_dimension___covering_number_bound}
\end{equation}
Therefore, $\calZ$ satisfies the Assumption made in \citep[Equation (2)]{BoissardLeGouic2014}; hence, we may apply \citep[Corollary 1.2]{BoissardLeGouic2014} to conclude that:
\begin{equation}
\mathbb{E}\left[
    \WD\left(
        \mu^N
            ,
        \mu
    \right)
\right]
        \leq 
    c\cdot K_{\calZ}^{1/d_{\calZ}}
    \left(\frac{2}{d_{\calZ}-2}\right)^{2/d_{\calZ}}
    \diam(\calZ)N^{-1/d_{\calZ}}
;
    \label{eq_PROOF___lem_concentration_inequality__Mean_bound}
\end{equation}
for some constant $0\leq c \leq \frac{2^6}{3}$. Let $C_{\calZ} \eqdef c \cdot K_{\calZ}$ and we prove \eqref{eq:lem:con_holder_wass_manifold:2}. Next, since $\diam(\calZ)<\infty$ and $\mu$ is a Borel measure on the polish space $\calZ$ then, \citep[Proposition 20]{WeedBach2019Bernoulli} applies; hence for every $\epsilon>0$ we have the estimate
\begin{equation*}
\sP\left(
    \bigg\vert\WD\left(\mu^N,\mu\right) - \mathbb{E}\left[\WD\left(\mu^N,\mu\right)\right] \bigg\vert \geq
   \epsilon
\right)
    \leq 
2e^{\frac{-2N\epsilon^2}{\diam(Z)^2}}.
\end{equation*}
\end{proof}

\begin{remark}[{Acceleration of Rates in Lemma~\ref{lem:con_holder_wass_manifold} Under Additional Regularity}]
\label{rem:furtherimprovedrates}
If $\mathcal{Z}$ is a submanifold of Euclidean space with finite-reach\footnote{The \textit{reach} of a submanifold $\mathcal{Z}$ of a Euclidean space is the largest radius $r\ge 0$ for which each point in the Euclidean space whose Euclidean distance from $\mathcal{Z}$ is at-most $r$ has a unique projection onto $\mathcal{Z}$; see \cite{MR2985939} for further details.} and if $\mu$ has a density with respect to the volume measure on $\mathcal{Z}$ then a variant of Lemma~\ref{lem:con_holder_wass_manifold} with a faster concentration rate can be derived using the results of \cite{block2021intrinsic} instead of \cite{BoissardLeGouic2014}. 
\end{remark}
\subsection{Helper Sub-Gaussian Concentration Inequalities}\label{app:subsec:gaussian}

\begin{definition}[Sub-Gaussian distribution]
    A centered random variable $X$ is called sub-Gaussian if there exists $C > 0$ and $\sigma > 0$ s.t. for all $x > 0$ that 
    \begin{equation*}
        \sP[|X| \geq x] \leq Ce^{-\frac{x^2}{2\sigma^2}},
    \end{equation*}
    denoted by $X \sim \mathrm{subG}(C,\sigma^2)$.
\end{definition}

\begin{lemma}\label{lem:subgaussian}
    Let $C,\sigma > 0$, $\tilde{\sigma} = \sigma\max\{C,1\}$, and $X$ be a centered random variable. Then each statement bellow implies the next:
    \begin{enumerate}
        \item $X \sim \mathrm{subG}(C,\sigma^2)$.
        \item $\sP[|X| \geq x] \leq Ce^{-\frac{x^2}{2\sigma^2}}$ for all $x > 0$.
        \item $\sE[|X|^{k}] \leq (2\tilde{\sigma}^2)^{\frac{k}{2}}\big(\frac{k}{2}\big)\Gamma\big(\frac{k}{2}\big)$ for all $k \in \sN_{\geq 2}$.
        \item $\sE[\exp(tX)] \leq e^{4\tilde{\sigma}^2 t^2}$ for all $t \in \R$.
        \item $X \sim \mathrm{subG}(2,4\tilde{\sigma}^2)$.
    \end{enumerate}
\end{lemma}
\begin{proof}\textbf{of Lemma~\ref{lem:subgaussian}.}
    $(i) \Rightarrow (ii)$ by definition.  $(ii) \Rightarrow (iii)$ is true by the following estimate:
    \begin{align*}
        \E[|X|^{k}] &= \int_{0}^{\infty} \sP[|X|^{k} \geq t]dt 
        = \int_{0}^{\infty} \sP[|X| \geq t^{\frac{1}{k}}]dt \\
        &\leq \int_{0}^{\infty} Ce^{-\frac{t^{2/k}}{2\sigma^2}}dt \\
        &\leq \frac{Ck(2\sigma^2)^{\frac{k}{2}}}{2}\int_{0}^{\infty} e^{-u}u^{\frac{k}{2} -1}du \\
        &\leq C\Big(\frac{k}{2}\Big)(2\sigma^2)^{\frac{k}{2}}\Gamma\Big(\frac{k}{2}\Big) \\
        &\leq \max\{1,C^2\}(2\sigma^2)^{\frac{k}{2}}\Big(\frac{k}{2}\Big)\Gamma\Big(\frac{k}{2}\Big) \\
        &\leq (2\max\{1,C^2\}\sigma^2)^{\frac{k}{2}}\Big(\frac{k}{2}\Big)\Gamma\Big(\frac{k}{2}\Big) \\
        &\leq (2\tilde{\sigma}^2)^{\frac{k}{2}}\Big(\frac{k}{2}\Big)\Gamma\Big(\frac{k}{2}\Big).
    \end{align*}
    $(iii) \Rightarrow (iv)$ is true because for all $t\in\R$
    \begin{align*}
        \E[e^{tX}] &\leq 1 + \sum_{k=2}^{\infty}\frac{t^{k}\E[|X|^{k}]}{k!}\\
        &\leq  1 + \sum_{k=1}^{\infty}\frac{(2\tilde{\sigma}^2 t^2)^{k}2k\Gamma(k)}{(2k)!} + \sum_{k=1}^{\infty}\frac{(2\tilde{\sigma}^2 t^2)^{k+1/2}(2k+1)\Gamma(k+1/2)}{(2k+1)!}\\
        &\leq 1 + (2 + \sqrt{2\tilde{\sigma}^2 t^2})\sum_{k=1}^{\infty}\frac{(2\tilde{\sigma}^2 t^2)^{k}k!}{(2k)!} \\
        &\leq 1 + (1 + \sqrt{\frac{\tilde{\sigma}^2 t^2}{2}})\sum_{k=1}^{\infty}\frac{(2\tilde{\sigma}^2 t^2)^{k}}{k!} \\
        &\leq e^{2\tilde{\sigma}^2 t^2} + \sqrt{\frac{\tilde{\sigma}^2 t^2}{2}}(e^{2\tilde{\sigma}^2 t^2} - 1)
        \leq e^{4\tilde{\sigma}^2 t^2}.
    \end{align*}
    $(iv)\Rightarrow (v)$: for all $x > 0$ and $t > 0$
    \begin{align*}
        \sP(X > x) &= \sP(e^{tX} > e^{tx}) \leq \frac{\E[e^{tX}]}{e^{tx}} \leq e^{4\tilde{\sigma}^2 t^2 - tx}.
    \end{align*}
    Therefore we have that 
    \begin{equation*}
        \sP(X \geq x) \leq e^{-\frac{x^2}{8\tilde{\sigma}^2}} ~, \text{and} ~~
        \sP(X \leq -x) \leq e^{-\frac{x^2}{8\tilde{\sigma}^2}}.
    \end{equation*}
    Therefore we conclude that 
    \begin{equation*}
        \sP(|X| \geq x) \leq 2e^{-\frac{x^2}{8\tilde{\sigma}^2}},
    \end{equation*}
    that is $X \sim \mathrm{subG}(2,4\tilde{\sigma}^2)$.

\end{proof}
\begin{lemma}\label{lem:subgaussian2}
    Let $X_{1},\dots,X_{n}$ be independent with $X_i \sim \mathrm{subG}(C,\sigma_i^2)$, and let $\alpha_i \geq 0$, $\tilde{\sigma}_i = \sigma_i \max\{C,1\}$, for all $ i = 1,\dots,n$. Then we have 
    \begin{equation*}
        \sum_{i=1}^{n}\alpha_i X_i \sim \mathrm{subG}(2,4\sum_{i=1}^{n}\alpha_i^2 \tilde{\sigma_i}^2),
    \end{equation*}
    that is, for all $x > 0$,
    \begin{equation*}
        \sP\Big[\big\vert\sum_{i=1}^{n}\alpha_i X_i\big\vert \geq x\Big] \leq 2e^{-\frac{x^2}{8\tilde{\sigma}^2}},
    \end{equation*}
    where $\tilde{\sigma}^2 = \sum_{i=1}^{n}\alpha_i^2 \tilde{\sigma_i}^2$.
\end{lemma}
\begin{proof}\textbf{of Lemma~\ref{lem:subgaussian2}.}
    By Lemma \ref{lem:subgaussian}, for all $i = 1,\dots, n$ and $t\in\R$ we have  that 
    \begin{equation*}
        \E[\exp(tX_i)] \leq e^{4\tilde{\sigma}_i^2 t^2}.
    \end{equation*}
    Then, by independence, we obtain
    \begin{equation*}
        \E[\exp(t\sum_{i=1}^{n}\alpha_i X_i)] = \prod_{i=1}^{n}\E[\exp(t\alpha_i X_i)] \leq \exp\left(4\sum_{i=1}^{n}\alpha_i^2 \tilde{\sigma}_i^2 t^2\right).
    \end{equation*}
    Then, by Lemma \ref{lem:subgaussian} we conclude that 
    \begin{equation*}
        \sum_{i=1}^{n}\alpha_i X_i \sim \mathrm{subG}(2,4\sum_{i=1}^{n}\alpha_i^2 \tilde{\sigma_i}^2).
    \end{equation*}
\end{proof}
\begin{lemma}[McDiarmid's inequality]\label{lem:mcdiarmid}
    Let $X_1,\cdots,X_N$ be independent random variables, where $X_i$ has range $\mathcal{X}_i$. Let $f \colon \mathcal{X}_1 \times \dots \times \mathcal{X}_N \to \R$ be any function with the $(c_1, \dots, c_N)$-bounded differences property: for every $i = 1,\dots, N$ and every $(x_1, \dots, x_N)$, $(x_1^{\prime},\dots, x_N^{\prime}) \in \mathcal{X}_1 \times \dots \mathcal{X}_N$ that differs only in the $i$-th coordinate, we have
    \begin{equation*}
        \vert f(x_1, \dots, x_N) - f(x_1^\prime, \dots, x_N^\prime)\vert \leq c_i.
    \end{equation*}
    Then for all $t > 0$ we have that
    \begin{equation*}
        \sP\Big[ f(X_1,\dots,X_N) - \sE[f(X_1,\dots,X_N)] \geq t\Big] \leq e^{\frac{-2t^2}{\sum_{i=1}^{N}c_i^{2}}},
    \end{equation*}
    and
    \begin{equation*}
        \sP\Big[ \big\vert f(X_1,\dots,X_N) - \sE[f(X_1,\dots,X_N)] \big\vert \geq t\Big] \leq 2e^{\frac{-2t^2}{\sum_{i=1}^{N}c_i^{2}}}.
    \end{equation*}
\end{lemma}
\begin{proof}\textbf{of Lemma \ref{lem:mcdiarmid}.}
See \cite{mcdiarmid1989method}.
\end{proof}

\begin{lemma}[Concentration of Global Mismatch]
   \label{lem:con_global_mismatch}
Let $\calZ$ a compact metric space and $\mu \in \calP_1(\calZ)$. Let $g^N\colon \calZ \to \R$ continuous and $\bP$ a finite partition of $\calZ$. Then for all $\delta > 0$, with probability $1-\delta$
    \begin{equation*}
        \sum_{P\in \bP}\bigg(1-
        \frac{\mu^{N}(P)}{\mu(P)}
        \bigg)\int_{z \in P}\, g(z)\,\mu(dz) \leq \frac{\Vert g^N \Vert_{\infty}}{N^{1/2}}\max\{\sqrt{2\ln(2/\delta)}, \sqrt{k}\}.
    \end{equation*}
\end{lemma}
\begin{proof}\textbf{of Lemma~\ref{lem:con_global_mismatch}.}
We notice that 
\begin{align*}
    \calR^N &\eqdef \sum_{P\in \bP}
        \bigg(1-\frac{
            \mu^{N}(P)
        }{
            \mu(P)
        }\bigg)
        \int_{z \in P}\, g^N(z)\,\mu(dz)\\
        &\leq \Vert g^N \Vert_{\infty}\sum_{P\in \bP}
        \big\vert \mu^{N}(P) - \mu(P) \big\vert.
\end{align*}
Let $\tilde{\mu}$ be a discrete distribution on $\bP$ s.t. $\tilde{\mu}(P) = \mu(P)$ and $\nu^{N}$ the empirical measure of $\nu$ with $N$ samples. Then we have
\begin{equation*}
    \sum_{P\in \bP}\big\vert \mu^{N}(P) - \mu(P) \big\vert = \mathrm{TV}(\nu,\nu^{N}),
\end{equation*}
where $\mathrm{TV}(\cdot,\cdot)$ denote the total variation distance. Therefore, by the empirical estimation under total variation distance \citep[Theorem 1]{canonne2020short}, for all $\epsilon>0$, $N \geq \max\{\frac{|\bP|}{\epsilon^2},\frac{2}{\epsilon^2}\log(2/\delta)\}$ 
\begin{equation*}
    \mathrm{TV}(\nu,\nu^{N}) \leq \epsilon.
\end{equation*}
Thus, we have with probability $1-\delta$
\begin{equation*}
    \mathcal{R}_N \leq \frac{\Vert g^N \Vert_{\infty}}{N^{1/2}}\max\{\sqrt{2\ln(2/\delta)}, \sqrt{k}\}.
\end{equation*} 
\end{proof}

\section{Uniform Rademacher Generalization Bound}\label{app:rademacher}
In this section we present the Rademacher generalization bound of Equation \eqref{eq:rademacher_bound} with more rigor.  
Consider $\calF_L$ the class of Lipschitz functions mapping $\calX$ to $\calY$, with Lipschitz constant of at most $L$, and let $\tilde \calF_L = \{ \Ls \circ f\, :\, f \in \calF_L \}$. 
Under assumptions~\ref{assum:domain}~and~\ref{assum:loss}, for any random sample $\calD^N$ of size $N$, and $0<\delta<1$ \citet[Theorem 8,][]{JMLR:bartlett2002rademacher} states that with probability greater than $1-\delta$
\begin{equation}
    \label{eq:rademacher_bound_2}
    \sup_{f \in \calF_L} \left\{ \mathfrak{R}(f ; \mu) - \hat{\mathfrak{R}}(f) \right\}\leq 2\hat\calR_N\left(\tilde \calF_L\right)+ \norm{\Ls}_\infty \sqrt{\frac{8\log 2/\delta}{N}}
\end{equation}
where $\hat\calR_N\left(\tilde \calF_L\right)$ is the empirical Rademacher complexity of $\tilde \calF_L$ which is defined via
\[
\hat \calR_N (\calH) = \sE_{\epsilon} \sup_{h \in  \calH }\frac{1}{N} \sum_{i=1}^N \epsilon_i h(X_i, Y_i)
\]
with $\epsilon = (\epsilon_1, \dots, \epsilon_N)$ being an i.i.d. vector of Rademacher random variables.
By contraction of Rademacher complexity \citep[Theorem 12,][]{JMLR:bartlett2002rademacher}, since the Loss is $L_\Ls$-Lipschitz we get
\begin{equation}
    \label{eq:contraction}
    \hat\calR_N\left(\tilde \calF_L\right) \leq 2 L_\Ls \hat\calR_N\left( \calF_L\right).
\end{equation}
In the next lemma, we bound the Rademacher complexity. 
\begin{lemma}\label{lem:rademacher}
\added{The Rademacher complexity of the class of $L$-lipschitz functions, defined on a $d$-dimensional domain is bounded as
\[
\hat\calR_N(\calF_L) \leq \left( \frac{8(d+1)^2D^2(16BL)^d}{N}\right)^{1/(d+3)} + 4 \sqrt{2} D \left( \frac{1}{N}\frac{(16BL)^d}{(8(d+1)D)^{d+1}}\right)^{1/(d+3)}
\]}
where $D \coloneqq \sup_{f \in \calF_L} \norm{f}_\infty$ and $\diam(\calX) \leq B$.
\end{lemma}
\added{By this lemma, and due to Equations~\eqref{eq:rademacher_bound_2}~and~\eqref{eq:contraction},
\begin{equation}\label{eq:exact_rademacher_bound}
\begin{split}
        \mathfrak{R}(f ; \mu) - \hat{\mathfrak{R}}(f) \leq 4L_\ell &  \left( \tfrac{8(d+1)^2D^2(16BL)^d}{N}\right)^{\tfrac{1}{(d+3)}} + 16 L_\ell \sqrt{2} D \left( \frac{1}{N}\tfrac{(16BL)^d}{(8(d+1)D)^{d+1}}\right)^{\tfrac{1}{(d+3)}} \\
    & + \norm{\Ls}_\infty \sqrt{\frac{8\log 2/\delta}{N}}
\end{split}
\end{equation}
}
Therefore, there exists $C$ for which with probability greater than $1-\delta$,
\[
 \mathfrak{R}(f ; \mu) - \hat{\mathfrak{R}}(f) \leq C L_\Ls \left( \frac{(dD)^2(BL)^d}{N}\right)^{1/(d+3)} + \norm{\Ls}_\infty \sqrt{\frac{8\log 2/\delta}{N}}
\]
implying that the generalization error vanishes at a $\mathcal{O}(N^{-1/(d+3)})$ rate. This concludes the derivation of Equation~\eqref{eq:rademacher_bound}.

\begin{proof}\textbf{of Lemma~\ref{lem:rademacher}.}
We start by bounding the Metric Entropy of the function class, and then applying a discretization bound.
Without a loss of generality, we may assume that $\bm{0}$ is included in $\calX \times \calY$. Since $\calX$ and $\calY$ are compact, then there exists $B$ such that $\calX \subset [0,B]^{d}$. \added{By \citet[Lemma 6]{gottlieb2017efficient}, the metric entropy of $\calF_L$ is bounded as
\[
\log \calN(\delta, \calF_L, \norm{\cdot}_\infty) \leq \left(\frac{16BL}{\delta}\right)^d\ln(8/\delta) \leq \left(\frac{16BL}{\delta}\right)^d\frac{8}{\delta}.
\]
}
Moreover, the 1-step discretization bound \citep[Proposition 5.17]{wainwright2019high} implies that
\[
\calR_N(\calF_L) \leq \frac{1}{\sqrt{N}}\inf_{\delta>0}\left( \delta \sqrt{N}  + 2 \sqrt{D^2 \log \calN(\delta, \calF_L, \norm{\cdot}_\infty)}\right)
\]
where $D^2$ is used to upper bound the $N$-norm $\sup_{f \in \calF_L} \sum_{i=1}^N f^2(X_i)/N$. 
By solving for $\delta$ and plugging in the optimal value we get that there exists constants $C_1$ and $C_2$ for which
\added{\[
\calR_N(\calF_L) \leq \left( \frac{8(d+1)^2D^2(16BL)^d}{N}\right)^{1/(d+3)} + 4 \sqrt{2} D \left( \frac{1}{N}\frac{(16BL)^d}{(8(d+1)D)^{d+1}}\right)^{1/(d+3)}.
\]}
\end{proof}

\section{Experiment Details}\label{app:experiments}

We include the details of the experiments in \cref{sec:props}.
For visualizing the bounds in all experiments, we have used $\Lip(\Ls \circ \fhat^N | P)$ since splitting the constant as $L_\Ls \Lip( \fhat^N | P)$ may loosen the bound, in particular for the classification experiments. 
All experiments are repeated for multiple random seeds, in each run the following are randomized: the learning problem (i.e. the training and test sets), the network initialization, the training algorithm.

\subsection{Task Descriptions} \label{appx:problem_setup}
\begin{figure}
     \centering
     \begin{subfigure}[b]{0.48\textwidth}
         \centering
         \includegraphics[width=\textwidth]{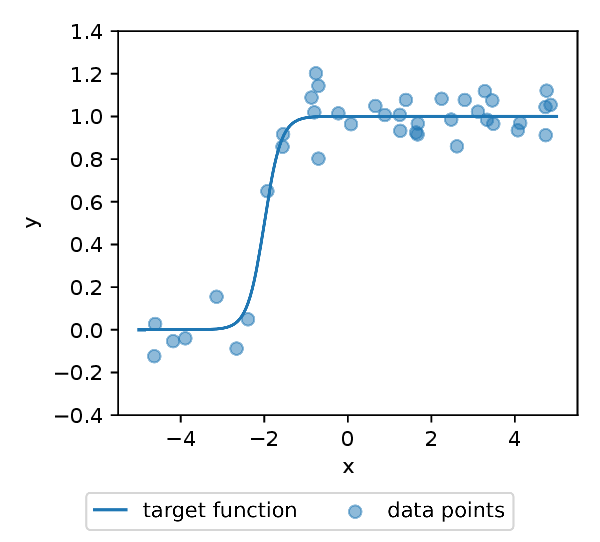}
         \caption{Regression}
         \label{fig:reg_problem}
     \end{subfigure}
     \hfill
     \begin{subfigure}[b]{0.48\textwidth}
         \centering
         \includegraphics[width=\textwidth]{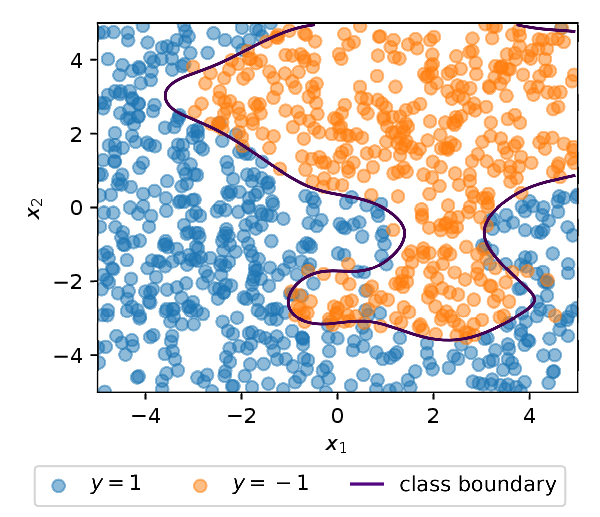}
         \caption{Classification}
         \label{fig:class_problem}
     \end{subfigure}
        \caption{One instance of the random datasets generated for neural network experiments}
        \label{fig:reg_class_problem}
\end{figure}

We generate random datasets for two toy regression and classification tasks. 
\paragraph{Regression problem.} \looseness -1 For our empirical evaluations of neural network regression, we use the simplistic problem of regressing on noisy observation of a modified logistic function. Formally, our target function $f^\star: \mathcal{X} \mapsto \mathcal{Y}$ with $\mathcal{X} = [-5, 5] \subset \R$ and  $\mathcal{Y} = [-1, 2] \subset \R$ is defined as
\begin{equation}
    f^\star(x) = \frac{1}{1 + \exp(5 (x+2))} \;.
\end{equation}
The inputs $x \in \mathcal{X}$ are sampled i.i.d from a uniform distribution $\mathcal{U}(-5, 5)$ and the corresponding regression labels follow as $y = f^\star(x) + \epsilon$, where $\epsilon \sim \mathcal{N}(0, 0.1^2)$ is i.i.d. Gaussian observation noise. Figure \ref{fig:reg_problem} illustrates $f^\star$ together with 20 noisy observations.
The loss function we use for regression is the Huber loss,
\begin{equation}
\label{eq:huber_loss}
    \Ls(y, \hat{f}(x)) := \left\{
\begin{array}{ll}
 (y - \hat{f}(x))^2 & \, \textrm{if} ~ |y - \hat{f}(x)| < 1\\
|y - \hat{f}(x)| & \, \textrm{otherwise} \\
\end{array}
\right. 
\end{equation}
which behaves like the mean squared error (MSE) for small and like the mean absolute error for large regression residuals. Training with the Huber loss is equivalent to training with the MSE plus gradient clipping and thus a common choice of practitioners to prevent large regression residuals from destabilizing the neural network training.

\paragraph{Classification problem.}
We also consider a binary classification with $\mathcal{X} = [-5, 5]^2 \subset \R^2$ and $\calY = \{-1, 1\}$. The input features $x \in \mathcal{X}$ are sampled i.i.d. from a uniform distribution over $\mathcal{X}$. The labels are sampled i.i.d from the Bernoulli distribution $\mathcal{B}(\sigma(f^{\text{logit}}(x_1, x_2)))$ where $\sigma(z) = 1 / (1 + \exp (-z))$ is the logistic function and 
\begin{equation}
    f^{\text{logit}}(x_1, x_2) = 10 \sqrt{(x_1 - 2)^2 + (x_2 - 2)^2} - \frac{1}{4} \sin (2 x_1) + \frac{3}{2} \cos (x_2) \;.
\end{equation}
\looseness -1 This binary classification problem is illustrated in Figure \ref{fig:class_problem}. 
During training, we use the negative cross-entropy error, 
\begin{equation}
    \Ls(x_1, x_2, y) = (1-y) f^{\text{logit}}(x_1, x_2) - \log(\sigma(f^{\text{logit}}(x_1, x_2)))
\end{equation}
which is commonly used for training neural network classifiers.
We visualize the bound of Corollary~\ref{cor:classification_bound}. To calculate the bound we consider the ramp lost $\Ls_\gamma$ (see \cref{sec:props}) with $\gamma = 5$.

\subsection{Details on the Neural Network Training}\label{app:exp_training}

In our empirical evaluations in \cref{sec:props}, we use fully-connected neural networks with leaky ReLU activation functions,
$$
\rho(z) = \left\{
\begin{array}{ll}
z & z \geq 0, \\
z / 10 & \, \textrm{otherwise}. \\
\end{array}
\right. \;
$$
We train the neural network by stochastic gradient descent with the AdamW \citep{loshchilov2017decoupled} optimizer which combines the adaptive learning rate method Adam with weight decay. Unless stated otherwise, we set the weight decay parameter to 0 (i.e., no weight decay), use an initial learning rate of 0.05 and decay the learning rate every 1000 iterations by 0.85. By default, we train for 20000 iterations with a mini-batch size of 8 in case of regression and 16 in case of classification. In the experiments where we do not vary the neural network size, we use $l = 3$ hidden layers with $w = 64$ neurons each.

\subsection{ Details of the Training Techniques Experiment (Fig.~\ref{fig:regularization})} \label{app:exp_gen}

The experiment is repeated for 10 random seeds and the error bounds show the standard error. Across the regression bounds, we use a partition of size $25$, i.e. a square mesh with $5$ intervals along the $x$ and $y$ dimension. As for the classification plots, we use a partition of size $1800$. 
We construct is as the union of two $30\times 30$ meshes in $\calX$, one located at $y=1$ and the other $y=-1$.
For a more formal definition, see \cref{app:proof_cor_classification_bound}. We visualize the plots for different sizes of training set $N$, the legend in \cref{fig:regularization} shows the values.

For adversarial training, we use perturbed samples $x^{\text{adv}} = x + \epsilon \nabla_x l(x, y)$ during stochastic gradient descent. How strongly the adversary perturbs the training inputs is controlled by $\epsilon$, i.e., the higher $\epsilon$, the higher the regularization effect. The $x$-axis of \cref{fig:class_avd_training} and Fig.~\ref{fig:reg_avd_training} corresponds to this parameter.
For training with weight-decay we effectively use the loss function
$
\Ls_{\mathrm{new}}(\bm{W}) = \Ls_{\mathrm{original}}(\bm{W}) + \lambda \norm{\bm{W}}_{\mathrm{F}}^2
$ where $\bm{W}$ denotes the network weights and $\lambda$ is the weight-decay constant which is down on the $x$-axis of  \cref{fig:class_weight_decay} and \cref{fig:reg_weight_decay}. Lastly, \cref{fig:class_early_stopping} and \cref{fig:reg_early_stopping} show the effect of early stopping, where the $x$-axis corresponds to the number of gradient descent iterations.

\subsection{Details of the Network Size Experiment (Fig.~\ref{fig:bounds_num_nn_params})}\label{app:exp_nnsize}
For this experiment, we pick the depth of the network as $l \in \{1, 2, 3, 4\}$ and the width of the network as $w \in \{ 32, 64, 128, 256 \}$.
The experiment is repeated for 10 random seeds and the error bounds show the standard error. 
For the regression plot, we use a partition of size $50$, i.e. a $10\times 5$ mesh with $10$ intervals along the $\calX$ and $5$ along the $\calY$ dimension. As for the classification plots, we use a partition of size $5000$. 
A partition is constructed as union of two $50 \times 50$ meshes in $\calX$, one located at $y=1$ and the other $y=-1$.
We visualize the plots for different sizes of training set $N$, the legend in \cref{fig:regularization} shows the values.

\subsection{Details of the Partitioning Experiment (Fig.~\ref{fig:num_parts})}\label{app:exp_part}
The experiment is repeated for 10 random seeds and the error bounds show the standard error. 
For the regression bounds, we consider partitions that divide the space into a uniform $M_\calX \times M_\calY$ mesh. 
The legend of \cref{fig:num_parts_reg} shows $M_\calY$, i.e. the number of parts made in $\calY$, and the 
horizontal axis shows $M_\calX$ the number of parts along $\calX$. The regression curves are all for a dataset size of $N=2560$.

\looseness -1 For the classification plot, we consider partitions of size $2M^2$, where $M$ is shown on the horizontal axis of the plot. A partition is constructed as union of two $M \times M$ meshes in $\calX$, one located at $y=1$ and the other $y=-1$. For a more formal definition, see \cref{app:proof_cor_classification_bound}. We visualize the plots for different sizes of training set $N$, the legend in \cref{fig:num_parts_class} shows the values.

\bibliography{refs}

\end{document}